\documentclass{article} 
\usepackage[dvipsnames]{xcolor}
\usepackage{iclr2022_conference,times}
\usepackage{dutchcal}
\usepackage{floatrow}
\usepackage{caption}

\usepackage[toc,page,header]{appendix}
\usepackage{minitoc}

\usepackage[colorlinks, citecolor=OliveGreen]{hyperref}
\usepackage{url}

\usepackage[utf8]{inputenc}
\usepackage{amsmath,bm, amssymb, amsfonts} 

\usepackage{dsfont}
\usepackage{multirow}
\usepackage{stmaryrd}
\usepackage{booktabs}
\usepackage{multirow}
\usepackage{graphicx}
\usepackage{subcaption}
\usepackage{nicefrac}
\usepackage{relsize}
\usepackage{mathabx}
\usepackage{xspace}
\usepackage{tikz}
\usepackage{array}

\usepackage{amsmath,amsfonts,bm}

\def\eqref#1{equation~\ref{#1}}

\def\1{\bm{1}}

\def\vn{{\bm{n}}}

\def\vq{{\bm{q}}}

\def\vu{{\bm{u}}}

\def\vx{{\bm{x}}}
\def\vy{{\bm{y}}}

\DeclareMathAlphabet{\mathsfit}{\encodingdefault}{\sfdefault}{m}{sl}
\SetMathAlphabet{\mathsfit}{bold}{\encodingdefault}{\sfdefault}{bx}{n}

\def\gI{{\mathcal{I}}}

\def\gL{{\mathcal{L}}}

\def\gN{{\mathcal{N}}}

\def\gP{{\mathcal{P}}}
\def\gQ{{\mathcal{Q}}}

\def\gS{{\mathcal{S}}}

\def\gU{{\mathcal{U}}}
\def\gV{{\mathcal{V}}}

\makeatletter
\DeclareRobustCommand\onedot{\futurelet\@let@token\@onedot}
\def\@onedot{\ifx\@let@token.\else.\null\fi\xspace}

\def\eg{\emph{e.g}\onedot} 
\def\ie{\emph{i.e}\onedot}

\makeatother

\newcommand{\eltwonorm}{$l_2$ norm\xspace}
\newcommand{\ltnorm}[1]{\left\|#1\right\|_2}

\newcommand{\mbm}[1]{\bm{#1}}

\newcommand{\mx}{\mbm{x}}

\newcommand{\rulesep}{\unskip\ \vrule\ }

\newcommand{\aboveEqVspace}{\vspace*{-0.1cm}}

\newcommand{\sfeatures}{Super-features\xspace}
\newcommand{\sfeature}{Super-feature\xspace}
\newcommand{\sfeaturesshort}{\color{Maroon}{FI}\color{BrickRed}{Re}\xspace} 
\newcommand{\sfeatureid}{\sfeature ID\xspace} 

\newcommand{\fimlong}{Local feature Integration Transformer\xspace}
\newcommand{\fim}{LIT\xspace}
\newcommand{\litlong}{\fimlong}
\newcommand{\lit}{\fim}

\newcommand{\mname}{\sfeatures}

\newcommand{\mnameshort}{\sfeaturesshort} 
\newcommand{\fire}{{\sfeaturesshort}\xspace}
\newcommand{\firelong}{{\color{Maroon}{F}}eature {\color{Maroon}{I}}ntegration-based {\color{BrickRed}{Re}}trieval\xspace}

\newcommand{\inputdim}{D}
\newcommand{\attdim}{d}
\newcommand{\nsfeat}{N}

\newcommand{\sfeatfunc}{\Phi} 
\newcommand{\sfeatset}{\gS} 

\newcommand{\sfeat}{\bm{s}} 
\newcommand{\sfeatfunccore}{\phi} 
\newcommand{\attfunc}{\psi} 

\newcommand{\sfeatureidfunc}{\mathcal{i}} 

\newcommand{\Kfunc}{K}
\newcommand{\Vfunc}{V}
\newcommand{\Qfunc}{Q}

\newcommand{\ROxford}{$\mathcal{R}$Oxford\xspace}
\newcommand{\RParis}{$\mathcal{R}$Paris\xspace}
\newcommand{\Rdistr}{$\mathcal{R}$1M\xspace}
\newcommand{\Rdis}{\Rdistr}

\newcommand{\val}{SfM-120k val.\xspace}

\newcommand{\bI}[1]{\textbf{\color{OliveGreen}{($\uparrow$ #1)}}}

\newcommand{\bB}[1]{\textbf{#1}}
\newcommand{\mpm}[1]{{\footnotesize ($\pm #1$)}}

\newcolumntype{H}{>{\setbox0=\hbox\bgroup}c<{\egroup}@{}}

\newcommand{\myparagraph}[1]{\vspace*{-0.1cm}\paragraph{#1}}
\newcommand{\mypartight}[1]{\vspace{0cm}{\noindent\textbf{#1}}}

\newfloatcommand{capbfigbox}{figure}[][\FBwidth]
\newfloatcommand{hcapbfigbox}{figure}[][1\linewidth]

\newcommand{\myparref}[2]{Section~\ref{#1}\hyperref[#1]{§#2}}
\newcommand{\myparlabel}[1]{\phantomsection 
\label{#1}}

\usepackage{pgfplots, pgfplotstable}
\pgfplotsset{compat=newest}
\usepgfplotslibrary{fillbetween}

\newcommand{\oxf}{MidnightBlue}

\newcommand{\paris}{RedViolet}

\newcommand{\valc}{PineGreen}

\newcommand{\ourscolor}{Maroon}
\newcommand{\howcolor}{Black}

\newcommand{\oxfmark}{o}
\newcommand{\parmark}{triangle}
\newcommand{\valmark}{square}

\newcommand{\leg}[1]{\addlegendentry{#1}}

\tikzset{every mark/.append style={solid}}
\pgfplotsset{
	grid=both, width=\linewidth, try min ticks=5,
    legend cell align=left, 
    legend style={fill opacity=0.8},
	ylabel near ticks,
    xlabel near ticks,
    every tick label/.append style={font=\footnotesize},
}

\pgfplotsset{
oxmed/.style={thick, color=\oxf, mark=o},
    oxhard/.style={thick, dashed, color=\oxf, mark=o},
    pamed/.style={thick, color=\paris, mark=star}, 
    pahard/.style={thick, dashed, color=\paris, mark=star},
    val/.style={thick, color=\valc, mark=x},
    oursoxf/.style={thick, color=\ourscolor, mark=\oxfmark},
    howoxf/.style={thick, dashed, color=\howcolor, mark=\oxfmark},
    ourspar/.style={thick, color=\ourscolor, mark=\parmark},
    howpar/.style={thick, dashed, color=\howcolor, mark=\parmark},
    oursval/.style={thick, color=\ourscolor, mark=\valmark},
    howval/.style={thick, dashed, color=\howcolor, mark=\valmark},
    numean/.style={thick, color=\ourscolor, mark=none},
    numin/.style={thick, color=gray, mark=none},
    wid/.style={thick, color=\ourscolor, mark=o, mark size=0.5pt},
    woid/.style={thick, densely dashdotted, color=RedOrange, mark=o, mark size=0.5pt},
}

\title{Learning \mname for Image Retrieval}

\author{Philippe Weinzaepfel, Thomas Lucas, Diane Larlus, and Yannis Kalantidis 
\\
NAVER LABS Europe, Grenoble, France
}

\iclrfinalcopy 
\begin{document}
\doparttoc 
\faketableofcontents 

\maketitle

\begin{abstract}
Methods that combine local and global features have recently shown excellent performance on multiple challenging deep image retrieval benchmarks, but their use of local features raises at least two issues. First, these local features simply boil down to the localized map activations of a neural network, and hence can be extremely redundant. 
Second, they are typically trained with a global loss that only acts on top of an aggregation of local features; by contrast, testing is based on local feature matching, which creates a discrepancy between training and testing. 
In this paper, we propose a novel architecture 
for deep image retrieval, based solely on mid-level features that we call \emph{\sfeatures}. 
These \sfeatures are constructed by an iterative attention module and constitute an \emph{ordered set} in which each element focuses on a localized \textit{and} discriminant image pattern. 
For training, they require only image labels.
A contrastive loss operates directly at the level of \sfeatures and focuses on those that match across images. 
A second complementary loss encourages diversity. 
Experiments on common landmark retrieval benchmarks validate that \sfeatures substantially outperform state-of-the-art methods when using the same number of features, and only require a significantly smaller memory footprint to match their performance.

Code and models are available at: \url{https://github.com/naver/FIRe}.

\end{abstract}

\section{Introduction}
\label{sec:intro}

Image retrieval is a task that models exemplar-based recognition, \ie a class-agnostic, fine-grained understanding task which requires to retrieve all images matching a query image over an (often very large) image collection. 
It requires learning features that are discriminative enough for a highly detailed visual understanding but also robust enough to extreme viewpoint/pose or illumination changes. A popular image retrieval task is landmark retrieval, 
whose goal is to single out pictures of the exact same landmark out of millions of images, possibly containing a different landmark from the exact same fine-grained class (\eg `gothic-era churches with twin bell towers'). 

While early approaches relied on handcrafted local descriptors, 
recent methods use image-level (global) or local Convolutional Neural Networks (CNN) features, see~\cite{csurka2018handcrafted} for a review. 
The current state of the art performs matching or re-ranking using CNN-based 
local features~\citep{delf,delg, how} and only learns with global (\ie image-level) annotations and losses. This is done by aggregating all local features into a global representation on 
which the loss is applied, creating a discrepancy between training and inference.

Attention maps from modules like the ones proposed by~\citet{vaswani2017attention} are able to capture intermediate scene-level information, which makes them fundamentally similar to mid-level features~\citep{xiao2015application,chen2019graph}. Unlike individual neurons in CNNs which are highly localized, attention maps may span the full input tensor and focus on more global or semantic patterns. 
Yet, the applicability of mid-level features for instance-level recognition and image retrieval is currently underwhelming; we argue that this is due to the following reasons: generic attention maps are not localized and may fire on multiple unrelated locations; at the same time, object-centric attentions such as the one proposed by~\citet{slotattention} produce too few attentional features and there is no mechanism to supervise them individually. In both cases, methods apply supervision at the global level, and the produced attentional features are simply not discriminative enough.

\begin{figure}[t]
    \centering
    \newlength{\sfeatvizwidth}
    \setlength{\sfeatvizwidth}{0.16\columnwidth}
    \includegraphics[width=\sfeatvizwidth]{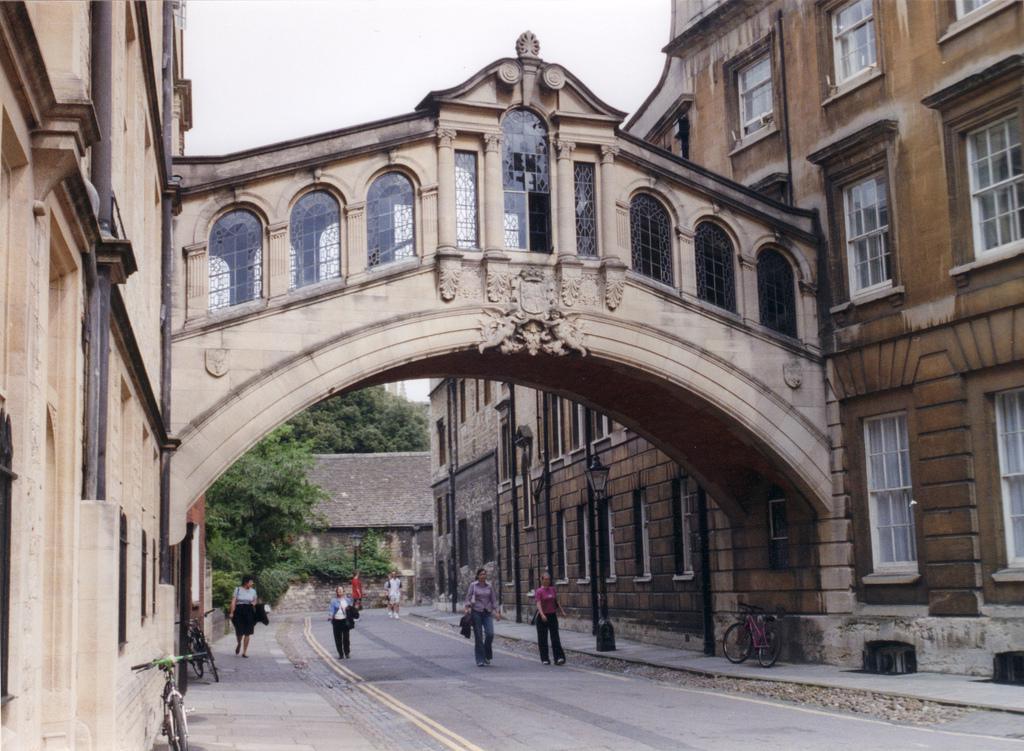} \hfill
    \includegraphics[width=\sfeatvizwidth]{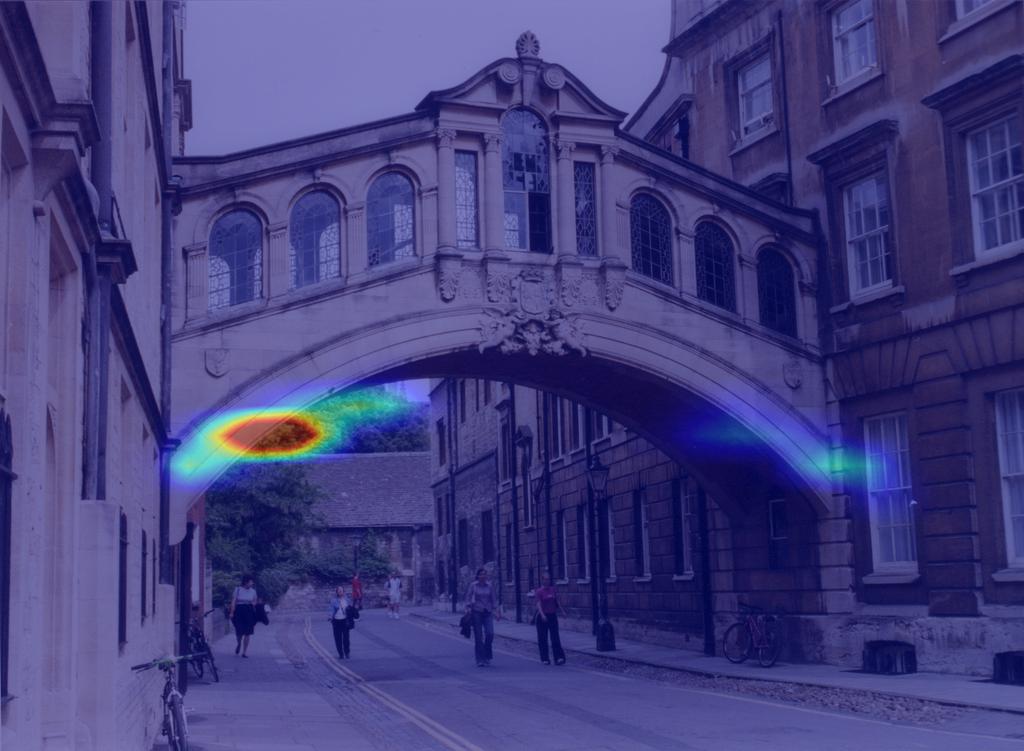} \hfill
    \includegraphics[width=\sfeatvizwidth]{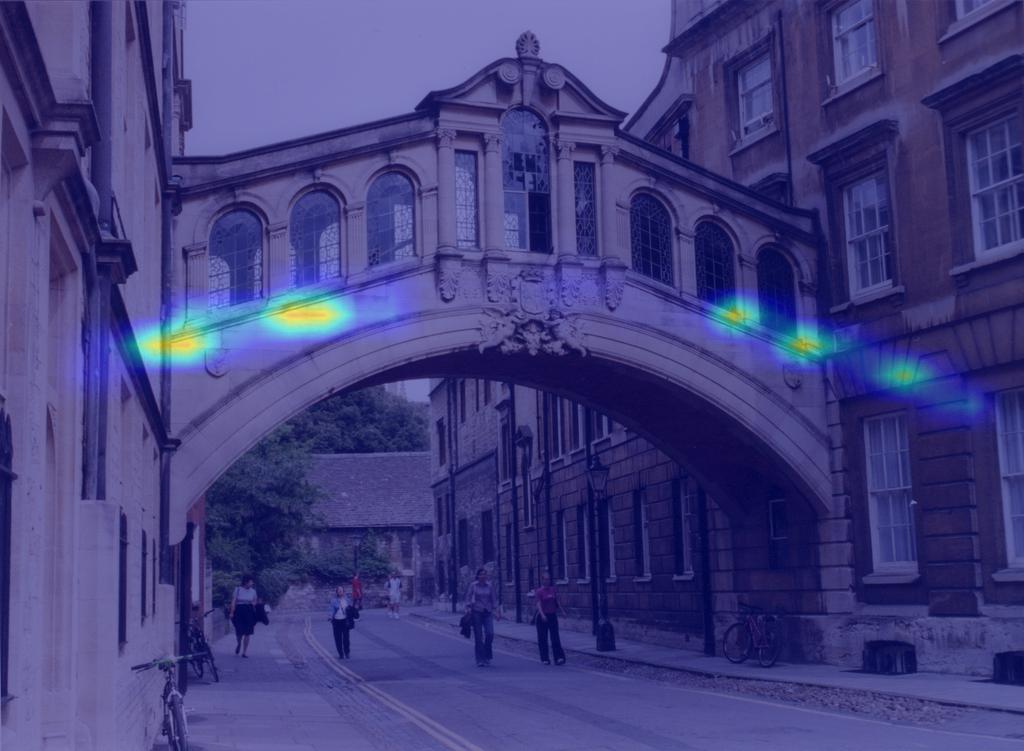} \hfill
    \includegraphics[width=\sfeatvizwidth]{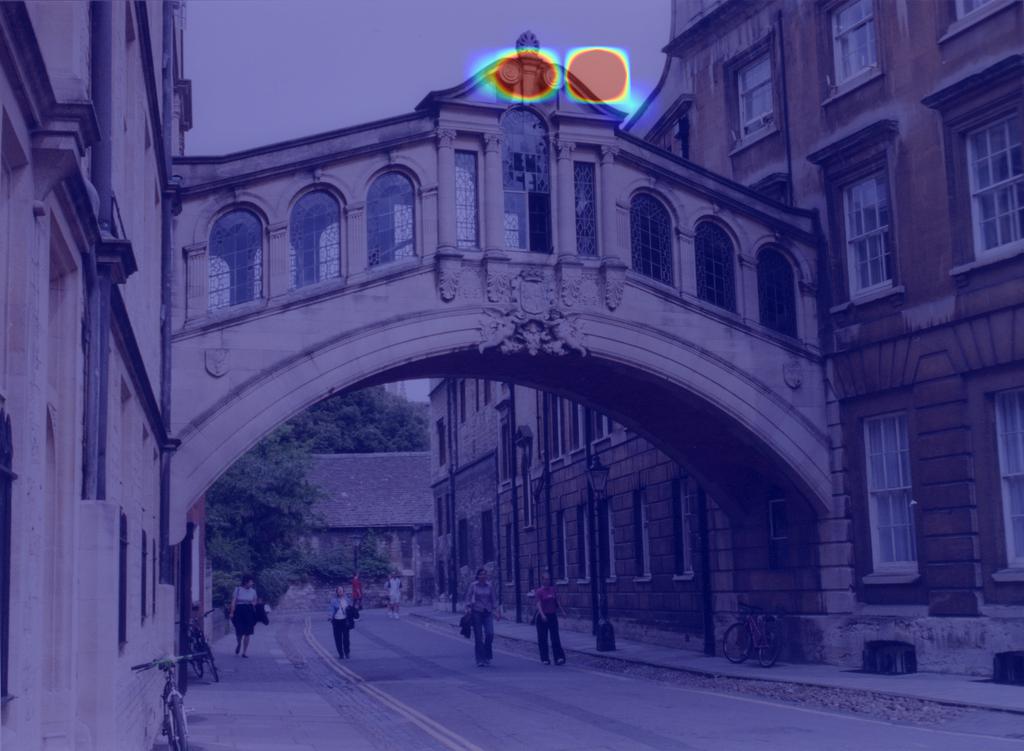} \hfill
    \includegraphics[width=\sfeatvizwidth]{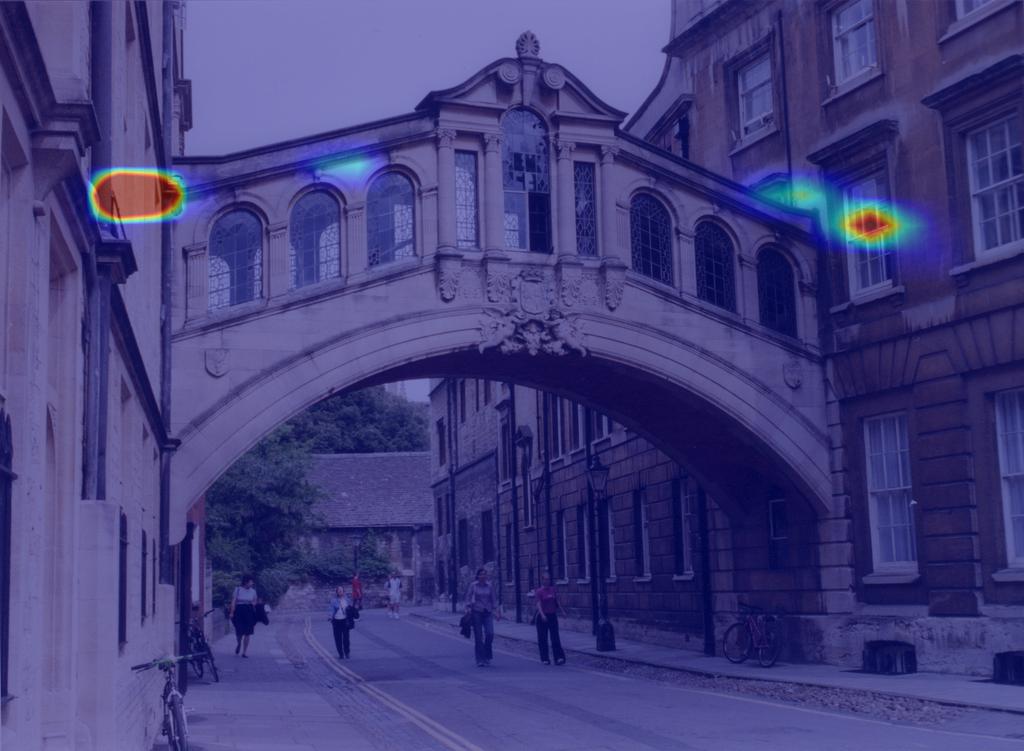} \hfill
    \includegraphics[width=\sfeatvizwidth]{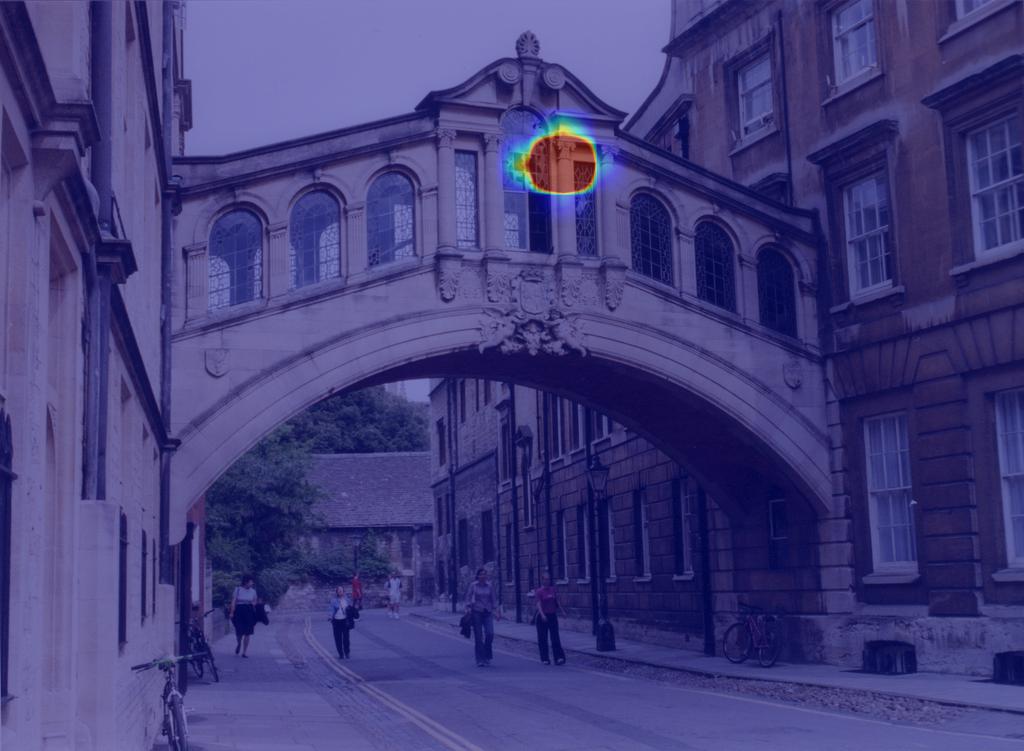} \\
    \includegraphics[width=\sfeatvizwidth]{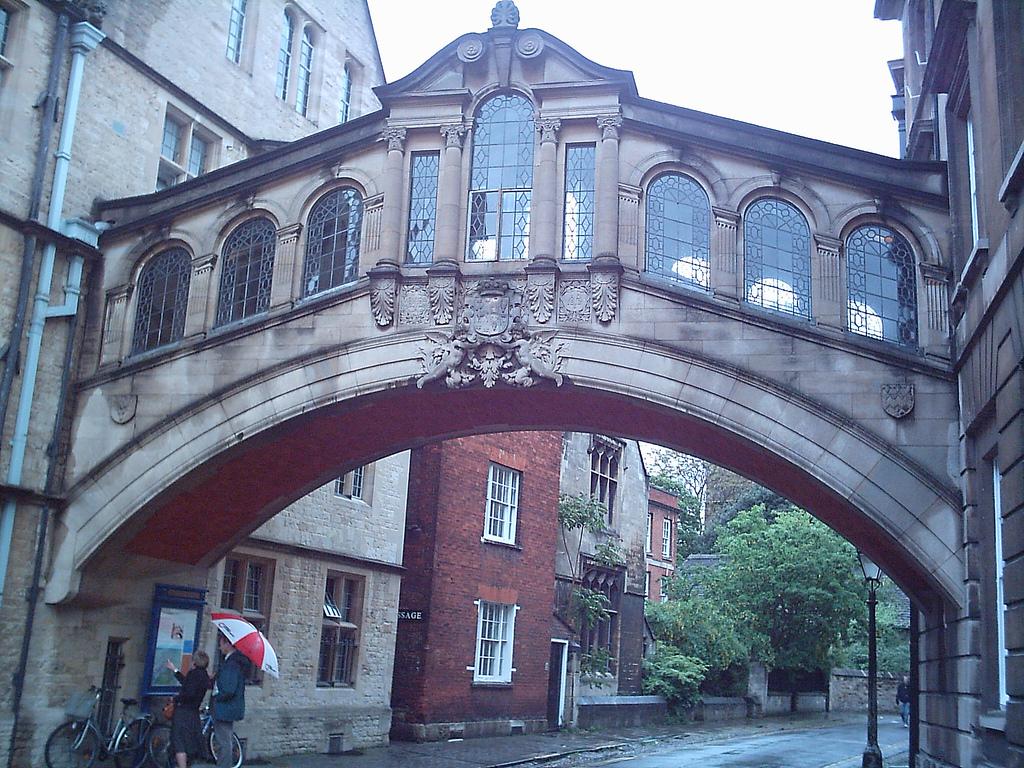} \hfill
    \includegraphics[width=\sfeatvizwidth]{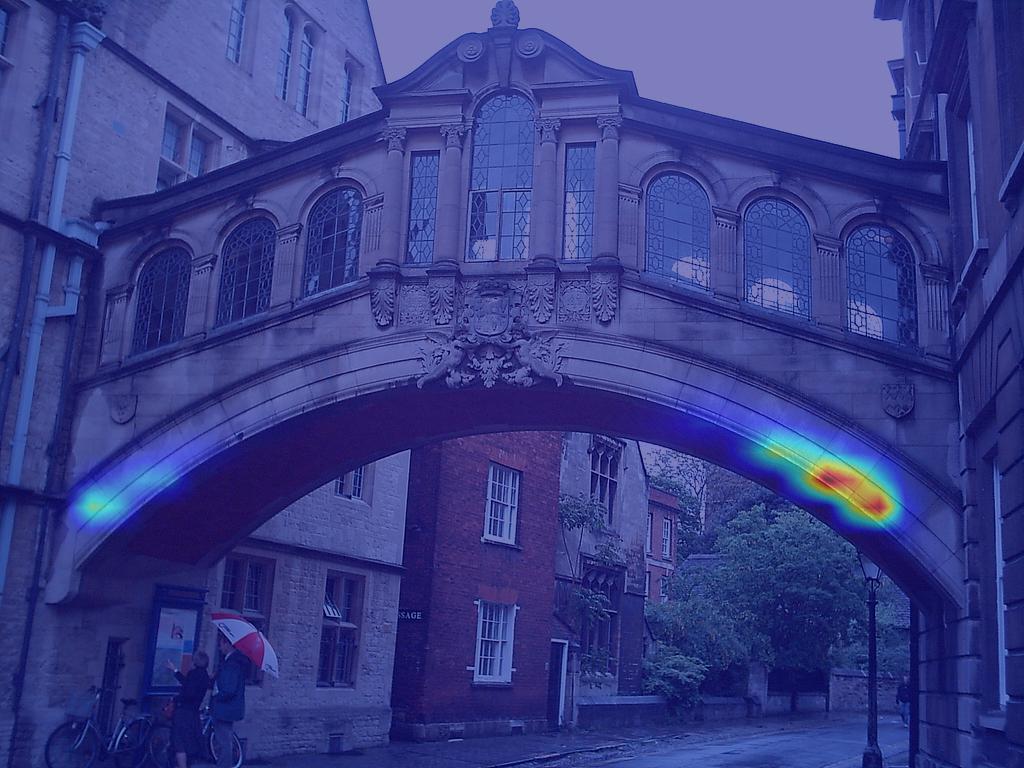} \hfill
    \includegraphics[width=\sfeatvizwidth]{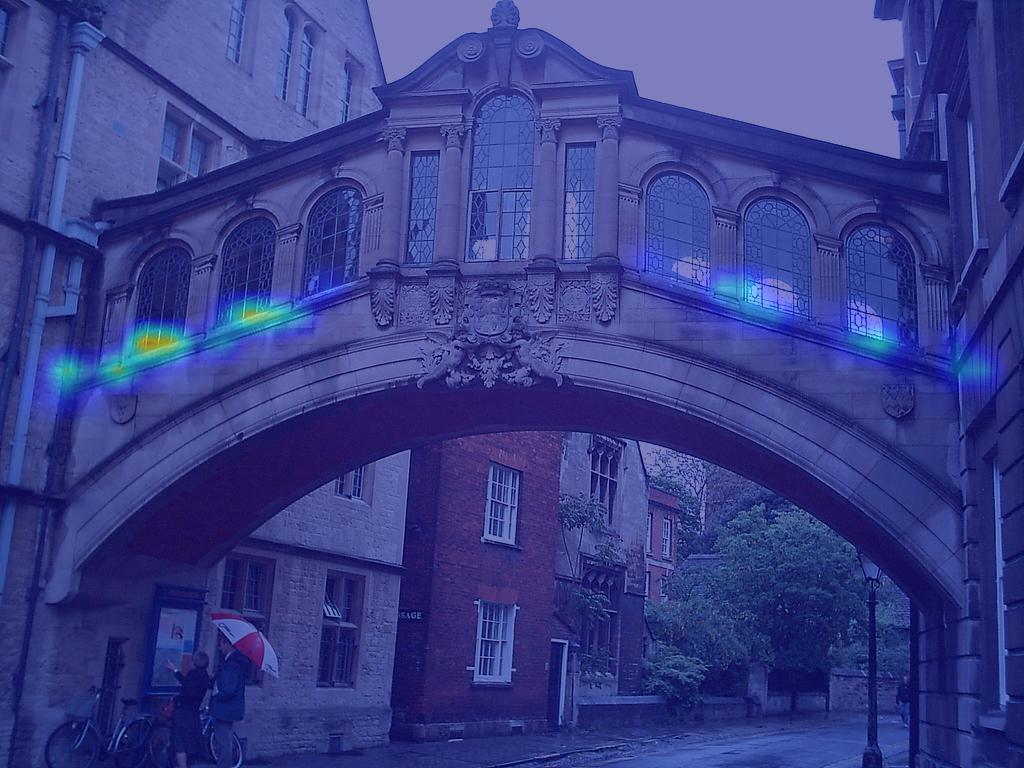} \hfill
    \includegraphics[width=\sfeatvizwidth]{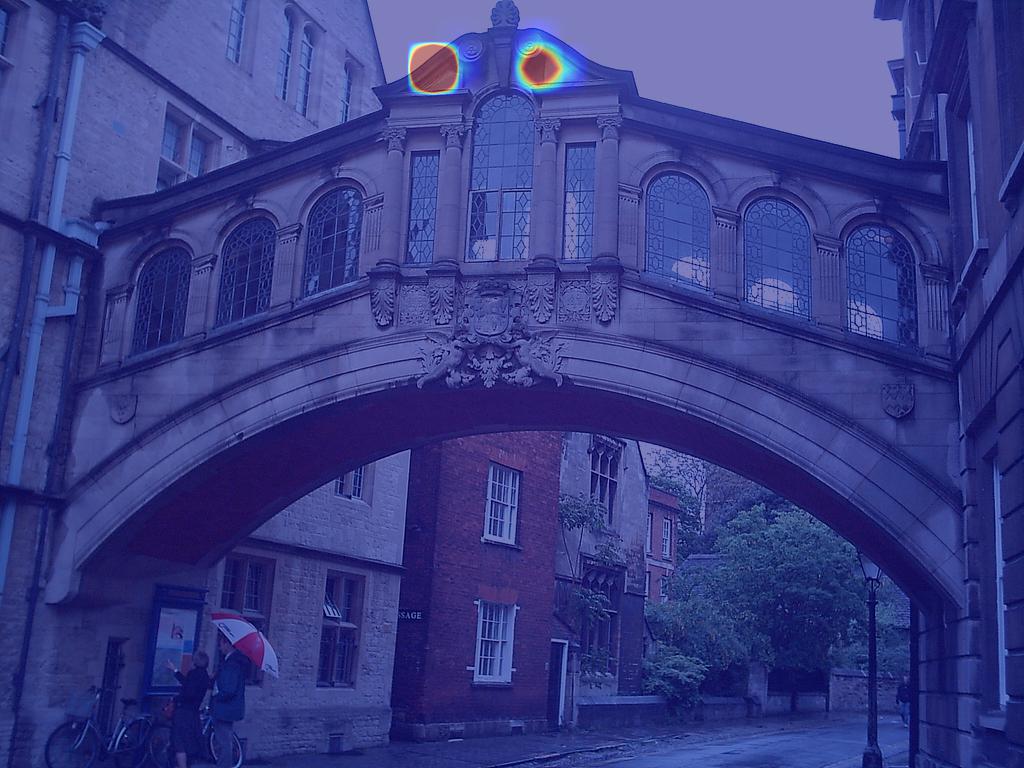} \hfill
    \includegraphics[width=\sfeatvizwidth]{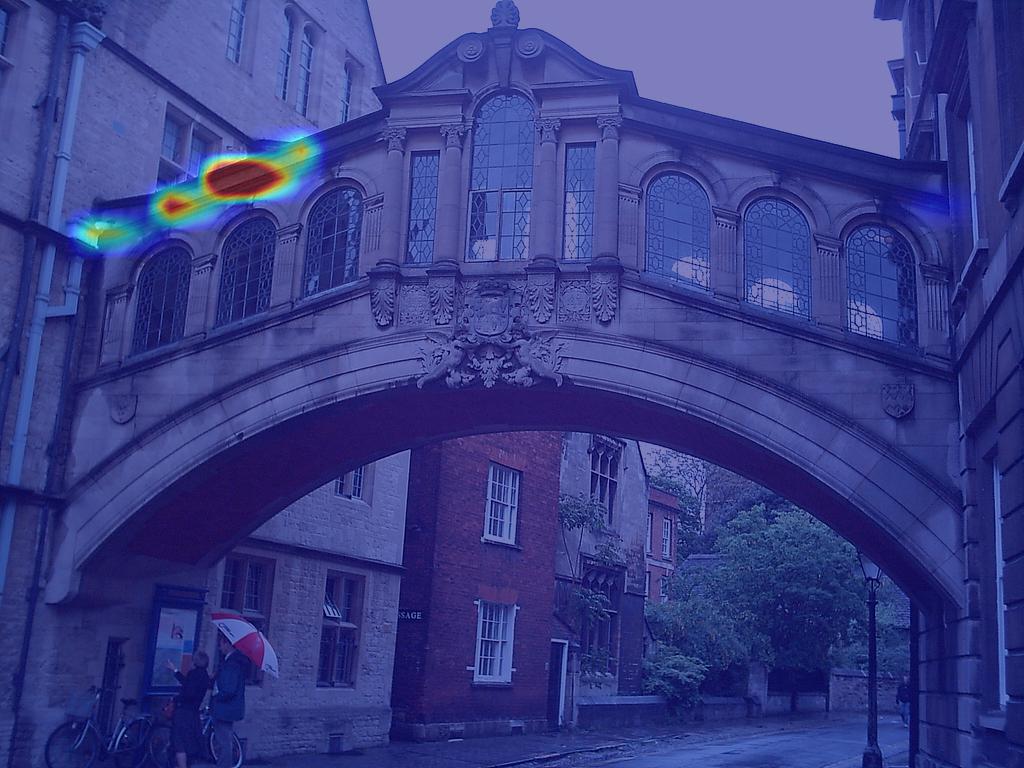} \hfill
    \includegraphics[width=\sfeatvizwidth]{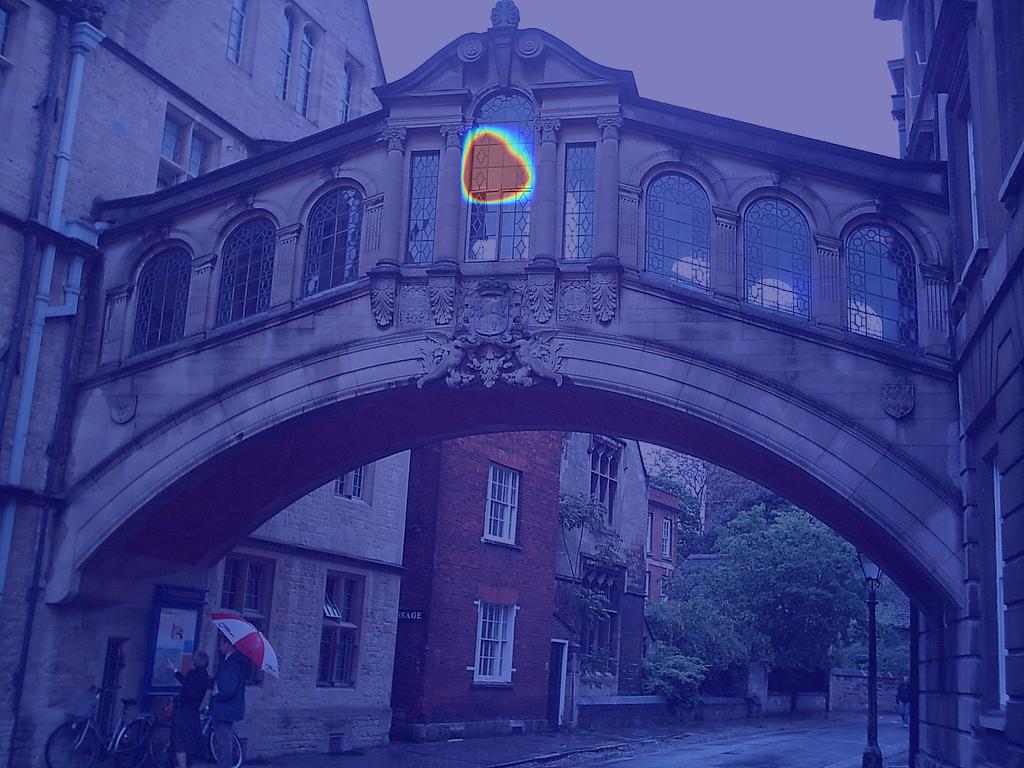} \\
    \includegraphics[width=\sfeatvizwidth]{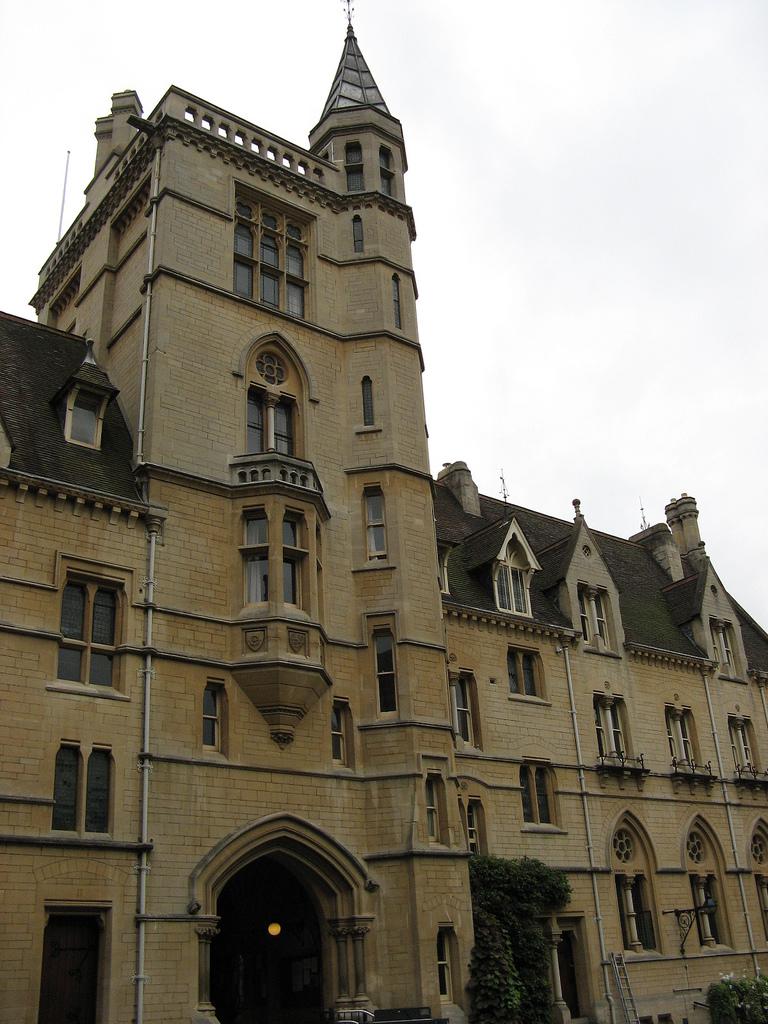} \hfill
    \includegraphics[width=\sfeatvizwidth]{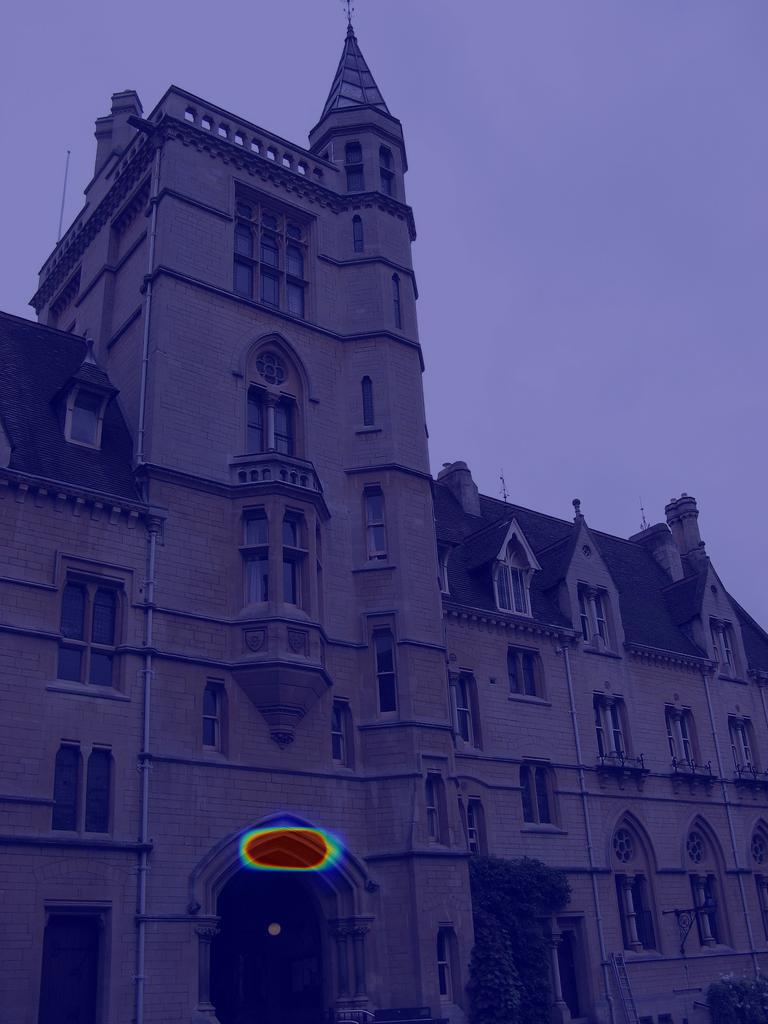} \hfill
    \includegraphics[width=\sfeatvizwidth]{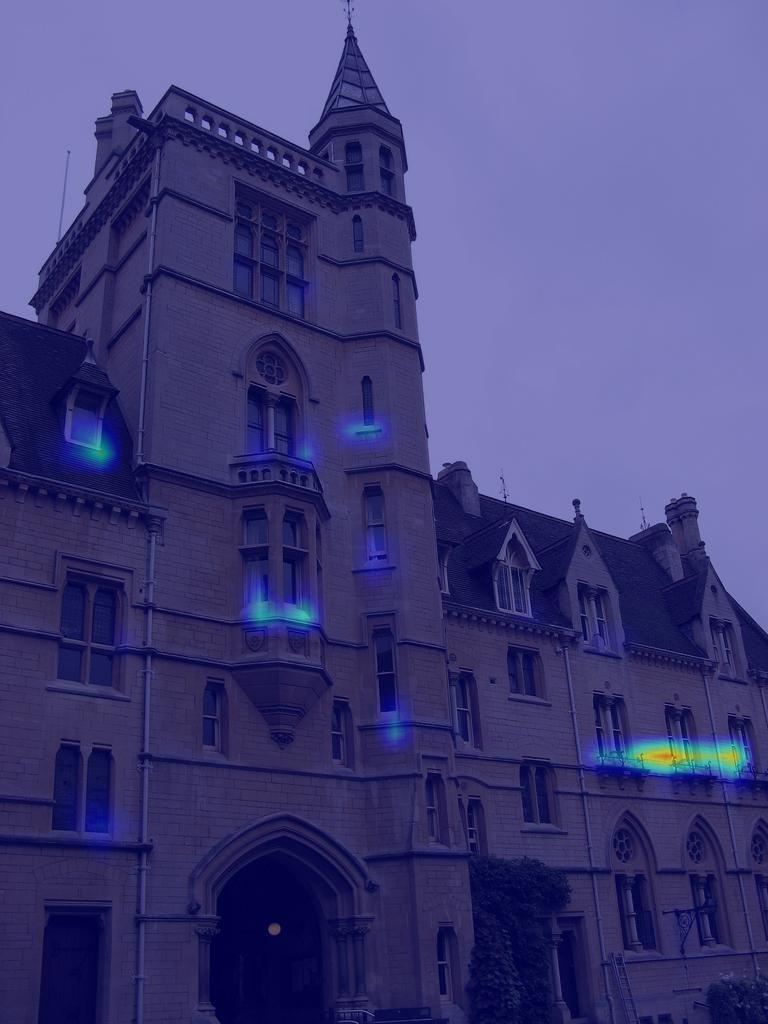} \hfill
    \includegraphics[width=\sfeatvizwidth]{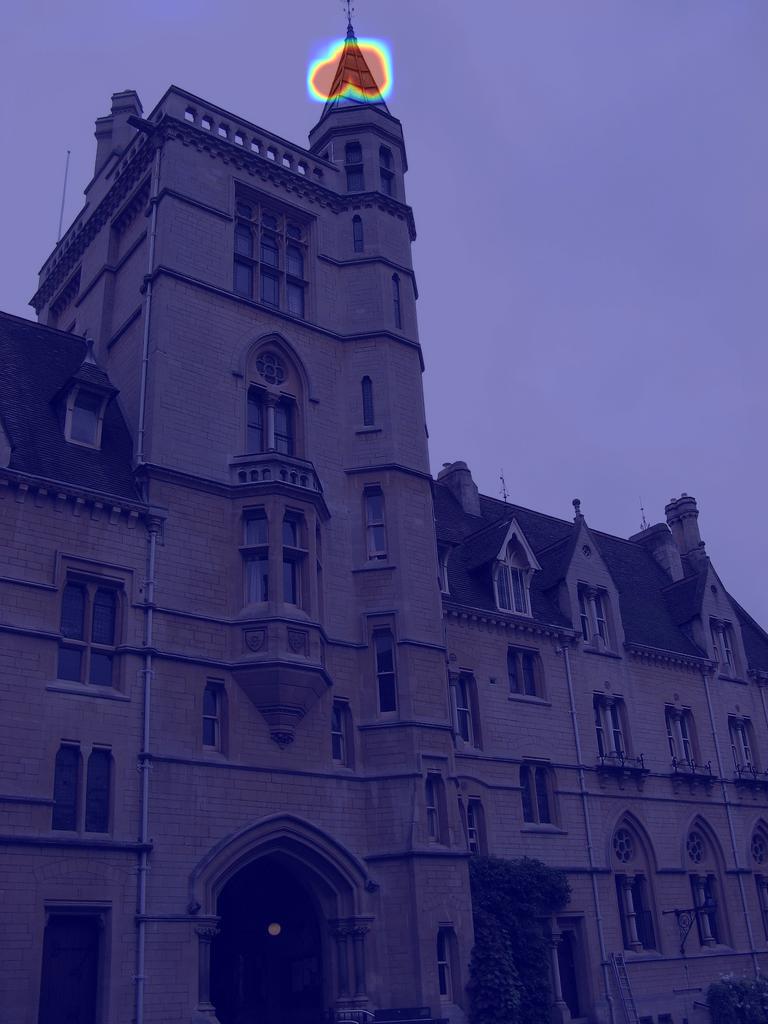} \hfill
    \includegraphics[width=\sfeatvizwidth]{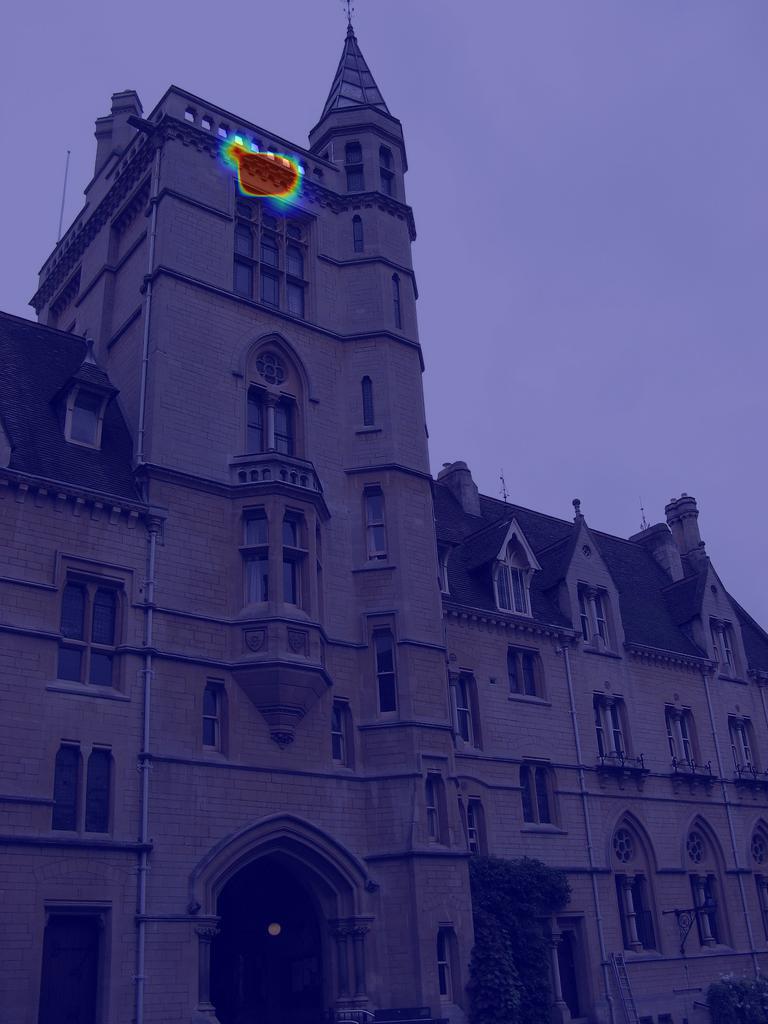} \hfill
    \includegraphics[width=\sfeatvizwidth]{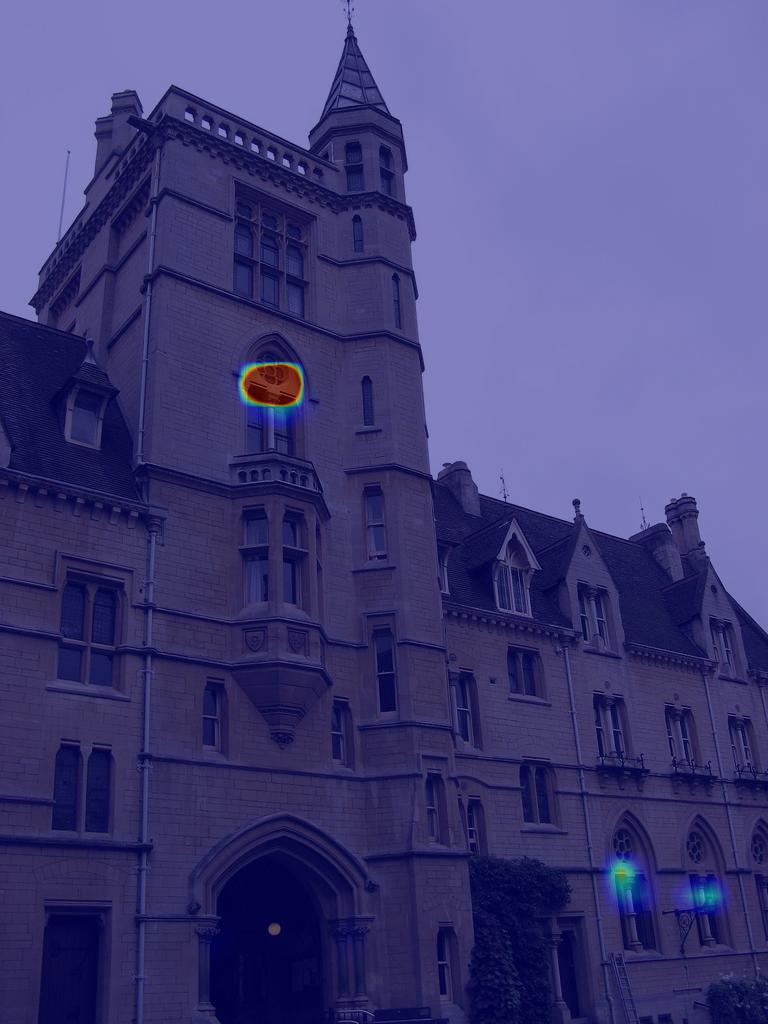} \\[-0.25cm]
    \caption{\textbf{\mname attention maps} produced by
    our iterative attention module (\fim)
    for three images (left), with the first two that match, for five \sfeatures. They tend to consistently fire on some semantic patterns, \eg circular shapes, windows, building tops (second to fourth columns). \vspace*{-0.4cm}
    }
    \label{fig:visualizing_sfeatures}
\end{figure}  

In this paper, we present a novel image representation and training framework based solely on attentional 
features we call \emph{\sfeatures}. 
We introduce an iterative \litlong (\lit), which tailors existing attention modules to the task of image retrieval.
Compared to the slot attention~\citep{slotattention} for example, it is able to output
an ordered and much larger 
set of features, as it is based on learned templates, and has a simplified recurrence mechanism. For learning, we devise a loss that is applied directly to \sfeatures, yet it only requires image-level annotations. 
It pairs a contrastive loss on a set of 
matching \sfeatures across matching images, with a decorrelation loss on the attention maps 
of each image, 
to encourage \sfeature diversity.
In the end, our network extracts for each image a fixed-size set of \sfeatures that are semantically ordered, \ie, 
each firing on 
different types of patterns; see Figure~\ref{fig:visualizing_sfeatures} for some examples.

At test time, we follow the protocol of the best performing recent retrieval methods and use ASMK~\citep{asmk},
except that we aggregate and match \sfeatures instead of local features. Our experiments show that the proposed method significantly outperforms the state of the art on common benchmarks like \ROxford and \RParis~\citep{revisited}, while requiring less memory. We further show that performance gains persist in the larger scale, \ie after adding 1M distractor images. Exhaustive ablations 
suggest
that \sfeatures are less redundant and more discriminative than local features.

\mypartight{Contributions.}
Our contribution is threefold:
(a) an image representation based on \textit{\sfeatures} and an
iterative module to extract them; (b) a framework to learn such representations, based on a loss applied directly on \sfeatures yet only requiring image-level labels; 
(c) extensive evaluations that show \textit{significant} performance gains over the state of the art for landmark image retrieval. We call our method \emph{\firelong} or \fire for short. 

\section{Background: Learning local features with a global loss}
\label{sec:background}

Let function $f : \gI \rightarrow \mathbb{R}^{W \times H \times \inputdim}$ denote a convolutional neural network (CNN) backbone that encodes an input image $\vx \in \gI$ into a $(W \times H \times \inputdim)$-sized tensor of $\inputdim$-dimensional local activations over a $(W \times H)$ spatial grid. After flattening the spatial dimensions, the output of $f$ can also be seen as set of $L = W \cdot H$ feature vectors denoted 
by $\gU = \{ \vu_l \in \mathbb{R}^\inputdim : l \in 1~..~L\}$; 
note that the size of this set varies with the resolution of the input image.
These local features are then typically whitened and their dimension reduced, a 
process that we represent by function $o(\cdot)$ in this paper. 
Global representations, \ie image-level feature vectors, are commonly produced by averaging all local features, \eg via global average or max pooling~\citep{spoc, rmac, dir}. 

\mypartight{A global contrastive loss for training.} 
\citet{how} argue that optimizing global representations is a good surrogate for learning local features to be used together with efficient match kernels for image retrieval. 
When building their global representation $g(\gU)$, they weight the contribution of each local feature to the aggregated vector using its \eltwonorm:
\begin{equation}
\aboveEqVspace
g(\gU) = \frac{\hat{g}(\gU)}{\ltnorm{\hat{g}(\gU)}}, \quad\quad \hat{g}(\gU) = \sum_{l=1}^L \ltnorm{\vu_l} \cdot o(\vu_l)  ,
\label{eq:how_pooling}
\end{equation}
where $\ltnorm{\cdot}$ denotes the \eltwonorm. 
Given a database where each image pair is annotated as matching with each other or not, 
they minimize a contrastive loss over tuples of global representations. 
Intuitively, this loss encourages the global representations of matching images to be similar and those of non-matching images to be dissimilar. 
Let tuple $(\gU, \gU^+, \gV^-_1,  \ldots, \gV^-_n)$ represent the sets of local features of 
images $(\vx, \vx^+, \vy^-_1, \ldots, \vy^-_n)$, where $\vx$ and $\vx^+$ are matching images (\ie a positive pair) 
and none of the images $\vy^-_1, \ldots, \vy^-_n$ 
is matching with image $\vx$ (\ie they are negatives).
Let $[\cdot]^+$ denote the positive part
and $\mu$ a margin hyper-parameter. They define a contrastive loss over global representations as:
\begin{equation}
\aboveEqVspace
\mathcal{L}_{global} = \ltnorm{g(\gU) - g(\gU^+)}^2 + \sum_{j=1}^n \big[ \mu - \ltnorm{g(\gU) - g(\gV^-_j)}^2 \big]^+.
\label{eq:loss_global}
\end{equation}
In \emph{HOW}, \citet{how} employ the global contrastive loss of Eq.(\ref{eq:loss_global}) to learn a model whose local features are then used with match kernels such as ASMK~\citep{asmk} for image retrieval. ASMK is a matching process defined over selective matching kernels of local features; it is a much stricter and more precise matching function than comparing global representations, and is crucial for achieving good performance. 
By learning solely using a loss defined over global representations and directly using ASMK over local features $\gU$, HOW achieves excellent image retrieval performance.

\section{Learning with \sfeatures}
\label{sec:method}

\begin{figure}[t]
    \centering
    \begin{subfigure}[b]{0.56\linewidth}
    \includegraphics[width=\linewidth]{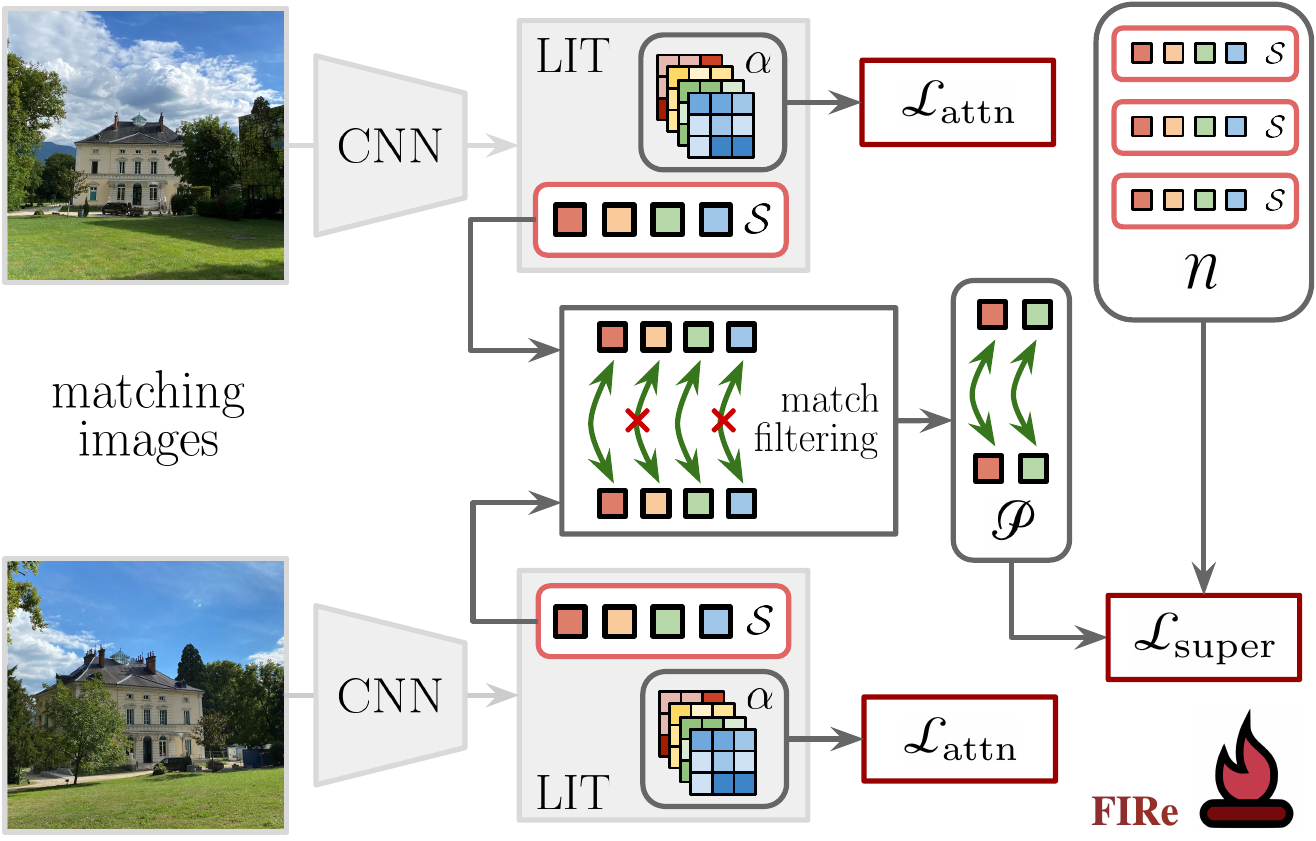}%
    \caption{The \fire training process.}
    \end{subfigure}%
    \rulesep
    \begin{subfigure}[b]{0.41\linewidth}
    \includegraphics[width=\linewidth]{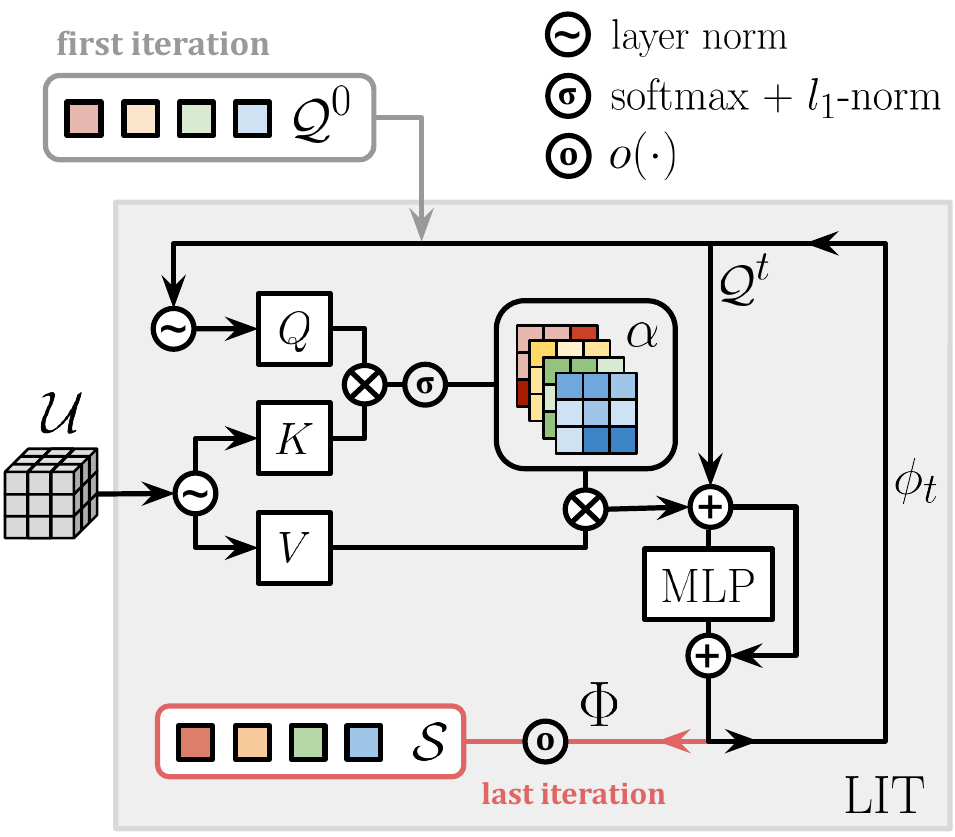}%
    \caption{The \lit module architecture.}
    \end{subfigure}
    
    \vspace{-7pt}
    
    \caption{\textbf{An overview of \fire.}
    Given a pair of matching images encoded by a CNN encoder, 
    the iterative attention module \lit (Section \ref{sub:iterative_attention_module}) outputs an ordered set of \sfeatures. 
    A filtering process keeps only reliable \sfeature pairs across matching images~(\myparref{par:pair_selection}{1}), which are fed into a \sfeature-level contrastive loss~(\myparref{par:contrastive_loss}{2}), while a decorrelation loss reduces the spatial redundancy of the \sfeatures attention maps for each image~(\myparref{par:spatial_correl}{3}). \vspace*{-0.4cm}    }
\label{fig:training_overview}
\end{figure}

Methods 
using a
global loss for training but local features for matching have a number of disadvantages. 
First, using local 
activations as local features leads to high redundancy, 
as they exhaustively cover highly overlapping patches of the input image. 
Second, using ASMK on local features from a model trained with a global loss introduces a mismatch between training and testing: the local features used for ASMK are only trained \textit{implicitly},
but are expected to individually match in the matching kernel. 
To obtain less redundant feature sets, we propose to learn \emph{\sfeatures} using a loss function that operates directly
on those features, and to also use the latter during retrieval;
the train/testing discrepancy of the pipeline presented in Section~\ref{sec:background} is thus eliminated.

In this section, we first introduce the \litlong (\emph{\lit}), an iterative attention module which produces an ordered set of \sfeatures (Section~\ref{sub:iterative_attention_module}). 
We then present a framework for effectively learning such features (Section~\ref{sub:learning_sfeatures}) that consists of two 
losses: A contrastive loss that matches individual \sfeatures across positive image pairs, and a decorrelation loss on the \sfeature attention maps that encourages them to be
diverse. 
An overview of the pipeline is depicted in Figure~\ref{fig:training_overview}.
We refer to our approach
as \emph{\firelong} or \fire for short, an homage to the 
feature integration theory of \citet{treisman1980feature}.

\subsection{\fimlong (\fim)}
\label{sub:iterative_attention_module}

Inspired by the recent success of attention mechanisms for encoding semantics from global context in sequences~\citep{vaswani2017attention} or images~\citep{caron2021emerging}, we rely on attention to design our \litlong (\emph{\lit}), a module that outputs an ordered set of \emph{\sfeatures}.

Let \lit be represented by function $\sfeatfunc(\gU): \mathbb{R}^{L \times \inputdim} \rightarrow \mathbb{R}^{\nsfeat \times \attdim}$ that takes as input the set of local features $\gU$ and outputs $\nsfeat$ \sfeatures. We define \lit as an \emph{iterative} module:
\begin{equation}
\aboveEqVspace
\sfeatfunc(\gU) = \gQ^T,
\quad\quad
\gQ^{t} = \sfeatfunccore(\gU ; \gQ^{t-1}),
\end{equation}
where $\sfeatfunccore$ denotes the core function of the module applied $T$ times, and $\gQ^0 \in \mathbb{R}^{\nsfeat \times \attdim}$ denotes a set of learnable \textit{templates}, \ie a matrix of learnable parameters. 
\sfeatures are progressively formed by iterative refinement of the templates, conditioned on the local features from the CNN.

The architecture of the core function $\sfeatfunccore$ is inspired by the Transformer architecture~\citep{vaswani2017attention} and is composed of a dot-product attention function $\attfunc$, followed by a multi-layer perceptron (MLP).
The dot-product attention function $\attfunc$ 
receives three inputs, the \textit{key}, the \textit{value} and the \textit{query}\footnote{\scriptsize The term `query' has a precise meaning for retrieval; yet, for this subsection only, we overload the term to refer to one of the inputs of the dot-product attention, consistently with the terminology from seminal works on attention by~\citet{vaswani2017attention}.} 
which are 
passed through layer normalization and 
fed to linear projection functions $\Kfunc$, $\Vfunc$ and $\Qfunc$ that project them to dimensions $\attdim_k$, $\attdim_v$ and $\attdim_q$, respectively. 
In practice, we set $\attdim_k{=}\attdim_v{=}\attdim_q{=}d{=}1024$.

The key and value inputs are set as the local features $\vu_l \in \gU$ across all iterations. The query input is 
the set of templates $\gQ^{t}= \{\vq^t_n \in \mathbb{R}^{\attdim}, n = 1~..~\nsfeat \}$.
It is initialized as the learnable templates  $\gQ^0$ for iteration $0$, 
and is set as the previous output of function $\sfeatfunccore$ for the following iterations.
After projecting with the corresponding linear projection functions, the key and the query are multiplied to construct a set of $\nsfeat$ \emph{attention maps} over the local features, \ie, the \emph{columns} of matrix $\mbm{\alpha} \in \mathbb{R}^{L \times N}$,
while the $L$ rows $\mbm{\alpha}_l$ of that matrix can be seen as the \textit{responsibility} that each of the $N$ templates has for each local feature $l$, and is given by%
\footnote{\scriptsize The attention maps presented in Eq.(\ref{eq:attention_weights}) are technically taken at iteration $t$, but we omit iteration superscripts for clarity. For the rest of the paper and visualizations, we use \emph{attention maps} to refer to the attention maps of Eq.(\ref{eq:attention_weights}) after the final ($T$-th) iteration of the iterative module.}:
\begin{equation}
\aboveEqVspace
 \mbm{\alpha} = \begin{bmatrix}
           \mbm{\alpha}_1 \\
           \vdots \\
           \mbm{\alpha}_L
         \end{bmatrix} \in \mathbb{R}^{L \times N}, ~~
\mbm{\alpha}_l = \frac{\mbm{\hat{\alpha}}_l}{\sum_{i=1}^L \mbm{\hat{\alpha}_i}}, ~~ 
\mbm{\hat{\alpha}_l} = \frac{\mbm{e^{{M}_l}}}{ \sum_{n=1}^N e^{M_{ln}}}, ~~
M_{ln} = \frac{\Kfunc (\vu_l)  \cdot~\Qfunc (\vq^t_n)}{\sqrt{\attdim}}.
\label{eq:attention_weights}
\end{equation}

The dot product between keys and templates can be interpreted as a tensor of compatibility scores between local features and templates. These scores are normalized across templates via a softmax function, and are further turned into attention maps by $l_1$ normalization across all $L$ spatial locations. 
This is a common way of approximating a joint normalization function across rows and columns,  also used by~\citet{slotattention}.
The input value is first projected with $\Vfunc$, re-weighted with the attention maps and then residually fed to a $\texttt{MLP}$\footnote{\scriptsize The $\texttt{MLP}$ function consists of a layer-norm, a fully-connected layer with half the dimensions of the features, a ReLU activation and a fully-connected layer that projects features back to their initial dimension.} to produce the output of function $\sfeatfunccore$: 
\begin{equation}
\aboveEqVspace
\gQ^t = \sfeatfunccore(\gU ; \gQ^{t-1}), 
\quad \sfeatfunccore(\gU ; \gQ) = \texttt{MLP}( \attfunc(\gU ; \gQ) ) + \attfunc(\gU ; \gQ), \quad \attfunc(\gU ; \gQ) = \Vfunc(\gU) \cdot \mbm{\alpha}  + \gQ.
\label{eq:phi}
\end{equation}
Following standard image retrieval practice,
we further whiten and $l_2$-normalize the output of $\sfeatfunc(\gU)$ to get the final set of \sfeatures. 
Specifically, let $\sfeatfunc(\gU)  = [ \hat{\sfeat}_1 ; \ldots ; \hat{\sfeat}_N ] \in \mathbb{R}^{\nsfeat \times \attdim}$ be the raw output of our iterative attention module, we define the ordered set of \sfeatures as:
\begin{equation}
\aboveEqVspace
\sfeatset = \Big\{ \sfeat_n: \sfeat_n=\frac{o(\hat{\sfeat}_n)}{\ltnorm{o(\hat{\sfeat}_n)}}, \quad n = 1, .., N \Big\}, 
\label{eq:sfeatures}
\end{equation}
where, as in Section~\ref{sec:background}, $o(\cdot)$ denotes dimensionality reduction and whitening. 
Figure~\ref{fig:training_overview} (right) illustrates the architecture of \lit. Note that all learnable parameters of $\phi$ are shared across all iterations.

\mypartight{What do \sfeatures attend to?}
\myparlabel{par:how_semantic}
Each \sfeature in $\mathcal{S}$ is a function of all local features in $\mathcal{U}$, invariant to permutation of its elements, and can thus attend arbitrary regions of the image. To visualize the patterns captured by \sfeatures,
Figure~\ref{fig:visualizing_sfeatures} shows the attention maps of the same five \sfeatures for three images, including two of the same landmark (first two rows). The type of patterns depends on the learned initialization templates $\mathcal{Q}^0$; this explains why the \mname form an ordered set, a property which allows to directly compare \sfeatures with the same ID. We observe the attention maps to be similar across images of the same landmark and to contain some mid-level patterns (such as a half-circle on the second column, or windows on the third one).

\subsection{Learning with \sfeatures}
\label{sub:learning_sfeatures}

We jointly fine-tune the parameters of the CNN and of the \lit module using contrastive learning. However, departing from recent approaches like HOW~\citep{how} or DELG~\citep{delg} that use a global contrastive loss similar to the one presented in Eq.~(\ref{eq:loss_global}), we introduce a contrastive loss that operates directly on \sfeatures, \emph{the representations we use at test time}, and yet only requires image-level labels. Given a positive pair (\ie matching images of the same landmark) and a set of negative images, we select promising \sfeature pairs 
by exploiting their ordered nature and without requiring any extra supervision signal.
We then minimize their pairwise distance, 
while simultaneously reducing the spatial redundancy of \sfeatures within an image.

\mypartight{Selecting matching \sfeatures.}
\myparlabel{par:pair_selection}
Since we are only provided with pairs of matching images, \ie image-level labels, defining correspondences at the \sfeature level is not trivial.
Instead of approximating sophisticated metrics \citep{liu2020self, mialon2020trainable}--see Section~\ref{sec:related} for a discussion--, we leverage the fact that \sfeatures are ordered and we select promising matches and filter out erroneous ones only relying on simple, nearest neighbor-based constraints.

For any $\sfeat \in \sfeatset$, let $\sfeatureidfunc(\sfeat)$ be the function that returns the position/order or \emph{\sfeatureid}, \ie $\sfeatureidfunc(\sfeat_i) = i,~\forall \sfeat_i \in \sfeatset$. Further let function $n(\sfeat, \sfeatset) =  \arg\min_{\sfeat_i \in \sfeatset}\ltnorm{\sfeat - \sfeat_i}$ be the function that returns the nearest neighbor of $\sfeat$ from set $\sfeatset$. Now, given a positive pair of images $\vx, \vx^+$, let  $\sfeat \in \sfeatset, \sfeat^\prime \in  \sfeatset^\prime$ be two \sfeatures from their \sfeature sets $\sfeatset, \sfeatset^\prime$, respectively. 
In order for \sfeature pair $(\sfeat, \sfeat^\prime)$  to be \emph{eligible}, all of the following criteria must be met: a) $\sfeat,\sfeat^\prime$ have to be reciprocal nearest neighbors, b) they need to pass Lowe's first-to-second neighbor ratio test~\citep{lowe2004distinctive} and c) they
need to have the same \sfeatureid. Let $\gP$ be the set of eligible pairs and $\tau$ the hyper-parameter that controls the ratio test; the conditions above can formally be written as:
\begin{equation}
\aboveEqVspace
    (\sfeat, \sfeat^\prime) \in \gP \iff 
    \begin{cases}
    s = n(\sfeat^\prime, \sfeatset) \\
    s^\prime = n(\sfeat, \sfeatset^\prime)
    
    \end{cases} \quad \text{and} \quad\quad 
    \begin{cases}
    \sfeatureidfunc(\sfeat) = \sfeatureidfunc(\sfeat^\prime) \\
    
    \ltnorm{\sfeat - \sfeat^\prime} ~ / ~ \ltnorm{\sfeat^\prime - n(\sfeat^\prime, \sfeatset \backslash \{ \sfeat \})} \ge \tau
    
    \end{cases}.
    \label{eq:pair_matching_criteria}
\end{equation}
We set $\tau = 0.9$. 
Our ablations (Section~\ref{sub:ablations}) show that all criteria are important.
Note that the pair selection step is non-differentiable and no gradients are computed for \sfeatures not in $\gP$. 
We further discuss this in Appendix~\ref{app:alltrained} and empirically show that all \sfeatures are trained.

\mypartight{A contrastive loss on \sfeatures.}
\myparlabel{par:contrastive_loss}
Once the set $\gP$ of all eligible \sfeature pairs has been constructed, we  define a contrastive margin loss on these matches.
Let pair $p = (\sfeat, \sfeat^+) \in \gP$ be a pair of \sfeatures{}, selected
from a pair of matching images $\vx, \vx^+$, 
and let  $\gN(j) = \{\vn^k_j: k = 1~..~n\}$ for $j = 1~..~\nsfeat$ be the set of \sfeatures with \sfeatureid $j$ extracted from negative images $(\vy^-_1, \ldots, \vy^-_n)$. The contrastive \sfeature loss can be written as:
\begin{equation}
\aboveEqVspace
\mathcal{L}_{super} = \sum_{ (\sfeat, \sfeat^+)\in \gP}  \Big[ \ltnorm{\sfeat - \sfeat^+}^2 + \sum_{\vn \in \gN({\sfeatureidfunc(\sfeat)})} [ \mu^\prime - \ltnorm{\sfeat - \vn}^2]^+ \Big],
\label{eq:loss_superfeatures}
\end{equation}
where $\mu^\prime$ is a margin hyper-parameter and the negatives for each $\sfeat$ are the \sfeatures from all $n$ negative images of the training tuple with \sfeatureid equal to $i(\sfeat)$. 

\mypartight{Reducing the spatial correlation between attention maps.}
\myparlabel{par:spatial_correl}
To obtain \sfeatures that are as complementary as possible, we encourage them to attend to different local features, \ie different locations of the image. To this end, we 
 minimize the cosine similarity between the attention maps of all \sfeatures of every image.

Specifically, let matrix $\mbm{\alpha} = [\tilde{\mbm{\alpha}}_1, \ldots, \tilde{\mbm{\alpha}}_N]$ now be seen as column vectors denoting the $N$ attention maps after the last iteration of \lit. 
The \emph{attention decorrelation loss} is given by:
\begin{equation}
\aboveEqVspace
\mathcal{L}_{attn}(\mx) = \frac{1}{\nsfeat(\nsfeat-1)}\:\: \mathlarger{\sum}\limits_{i \neq j}\:\:
\frac{\tilde{\mbm{\alpha}}_i^\intercal \cdot \tilde{\mbm{\alpha}}_j}{\ltnorm{\tilde{\mbm{\alpha}}_i}\ltnorm{\tilde{\mbm{\alpha}}_j}}, \quad i,j \in \{1,..,\nsfeat\}.
\label{eq:loss_attn}
\end{equation}
In other words, this loss minimizes the off-diagonal elements of the $\nsfeat \times \nsfeat$ self-correlation matrix 
of $\tilde{\mbm{\alpha}}$.
We ablate the benefit of this loss and others components presented in this section in Section~\ref{sub:ablations}. 

\mypartight{Image retrieval with \sfeatures.} 
Our full pipeline, \fire, is composed of a model that outputs \sfeatures, trained with the contrastive \sfeature loss of Eq.(\ref{eq:loss_superfeatures}) and the attention decorrelation loss of Eq.(\ref{eq:loss_attn}).
As our ablations show (see Section~\ref{sec:experiments}), the combination of these two losses is required for achieving state-of-the-art performance, while
adding a third loss on the aggregated global features as in Eq.(\ref{eq:loss_global}) does not bring any gain.
\section{Experiments}
\label{sec:experiments}

This section validates our proposed \fire approach on standard landmark retrieval tasks.
We use the SfM-120k dataset~\citep{cirtorch}
following the 551/162 3D model train/val split from~\cite{how}. For testing, we evaluate instance-level search on the \ROxford~\citep{oxford} and the \RParis~\citep{paris} datasets in their revisited version~\citep{revisited}, with and without the 1 million distractor set called \Rdistr. They both contain 70 queries, with 4993 and 6322 images respectively. We report mean average precision (mAP) on the Medium (med) and Hard (hard) setups, or the average over these two setups (avg).  

\myparagraph{Image search with \sfeatures.}
At test time, we follow the exact procedure described by \citet{how} and extract \sfeatures from each image at 7 resolutions/scales \{2.0, 1.414, 1.0, 0.707, 0.5, 0.353, 0.25\}. We then keep the top \sfeatures (the top 1000 unless otherwise stated) according to their L2 norm and use the binary version of ASMK with a codebook of 65536 clusters. Note that one can measure the memory footprint of a database image $\vx$ via the number of non-empty ASMK clusters, \ie clusters with at least one assignment, denoted as $|\mathcal{C}(\vx)|$. 

\mypartight{Implementation details.}
We build our codebase on HOW~\citep{how}\footnote{\url{https://github.com/gtolias/how}} and sample tuples composed of one query image, one positive image and 5 hard negatives. Each epoch is composed of 400 batches of 5 tuples each, while hard negatives are updated at each epoch using the global features of Eq.(\ref{eq:how_pooling}). 
We train our model for 200 epochs on SfM-120k 
using an initial learning rate of $3.10^{-5}$ and random flipping as data augmentation. We multiply the learning rate by a factor of $0.99$ at each epoch, and use an Adam optimizer with a weight decay of $10^{-4}$. We use a ResNet50~\citep{resnet} without the last convolutional block as backbone (R50$^-$). For \lit, we use $D=\attdim_k= \attdim_q = \attdim_v=\attdim=1024$. Following HOW, we reduce the dimensionality of features to 128 and initialize $o(\cdot)$ using PCA and whitening before training and keep it frozen. We pretrain the \lit module 
with the backbone on ImageNet-1K for image classification, see details in Appendix~\ref{app:pretrain}. 
We use $\mu'=1.1$ in Eq.(\ref{eq:loss_superfeatures}) and weight
$\gL_{super}$ and $\gL_{attn}$ 
with $0.02$ and $0.1$, respectively.

\subsection{Analysis and ablations}
\label{sub:ablations}

In this section, we analyze the proposed \fire framework and perform exhaustive ablations on the training losses and the matching constraints. The impact of the number of iterations ($T$) and templates ($N$) in \lit is studied in Appendix~\ref{app:ablate}. For the rest of the paper, we set $T=6$ and $N=256$.%

\begin{figure}[tb]
\begin{floatrow}
\ttabbox[1\linewidth]{%
  \vspace*{-2.8cm}
  \resizebox{1\linewidth}{!}{
            \raggedleft
            \centering
\begin{tabular}{lcc c cc cc}
\toprule
reci- & ratio & same & SfM-120k & \multicolumn{2}{c}{\ROxford} & \multicolumn{2}{c}{\RParis} \\

 proc.    & test  & ID  & val & med & hard & med & hard \\

\cmidrule(l){1-3} \cmidrule(l){4-4} \cmidrule(l){5-6} \cmidrule(l){7-8}

 &             &              & 68.3 & 64.3 & 39.8 & 74.1 & 52.4 \\
$\checkmark$    &     &              & 79.6 & 69.2 & 44.3 & 79.2 & 60.9 \\
$\checkmark$ &   $\checkmark$ &           & 80.8 & 70.7 & 45.1 & 80.3 & 61.9 \\
            & &  $\checkmark$ & 75.9 & 63.8 & 35.1 & 77.3 & 56.5 \\
$\checkmark$ &   $\checkmark$  & $\checkmark$ & \bf{89.7} & \bf{81.8} & \bf{61.2}	& \bf{85.3} & \bf{70.0} \\
\bottomrule
\end{tabular}       
        }
}{%
  \caption{\footnotesize \textbf{Ablation on matching constraints}: Impact of removing constraints on reciprocity, Lowe’s ratio test and the \sfeature ID.\vspace*{-1.45cm}}\label{tab:match}%
}
\hcapbfigbox{%
    \begin{tikzpicture}
\begin{axis}[%
  height=4cm, 
  xlabel={\scriptsize Epochs},
   x label style={at={(axis description cs:1.15,0.0)},anchor=north},
  ylabel={\footnotesize \hspace*{-0.6cm}~~~~~~Match ratio }, 
  minor tick num=1,
    xmin=0, xmax=200,
  legend style={at={(0.9,0.5)},anchor=east},
   ylabel near ticks, yticklabel pos=right,
  ]

\pgfplotstableread{
xxx id noid 
1 0.41997927059332335 0.2715170740516815
2 0.44504480857924933 0.2751341442890694
3 0.4609426352563114 0.27721018787388946
4 0.48491277802965943 0.28423492353548163
5 0.496776750330251 0.2869743077443483
6 0.4884136312438439 0.2819880614598237
7 0.4900718763333224 0.28212137081837807
10 0.51235126809633 0.2861563044741429
13 0.5321785232977742 0.29377187970287866
16 0.55156367795773 0.29684059894443265
19 0.5626758688377387 0.30027225939603197
22 0.5765442347820234 0.3031296184907918
25 0.5914481932882267 0.3052692019638845
28 0.5783934594126892 0.29934647966583705
31 0.5880259589896712 0.30100893551483376
34 0.6078176403262244 0.30981043309810435
37 0.6000301210489656 0.30535121561193157
40 0.6120156098530983 0.3062143363608738
43 0.6124085237883774 0.3066533258332628
46 0.6119163940430445 0.30497937762781036
49 0.6307199086242367 0.3122196559167831
52 0.6180602561973971 0.3100401180264122
55 0.6470437100555609 0.31738722532556396
58 0.6407312178602784 0.3168293336135793
61 0.6370916866437788 0.31296198067000547
64 0.6485509128210455 0.31843709981787655
67 0.642073048169257 0.3148615770737829
70 0.6336478998187172 0.3124367228210255
73 0.6447837876409305 0.3170280456341704
76 0.6493915877489415 0.3169418098899312
79 0.6665232927474855 0.32454266153054384
82 0.6635166943928653 0.32275479694722964
85 0.6544481382978723 0.3202721511874195
88 0.6734666586430773 0.32817079961197826
91 0.673673379391445 0.3251875692173307
94 0.6629870395355493 0.32423690666741223
97 0.6707207489603577 0.32624099603300905
100 0.6723263717590309 0.32633739720947563
103 0.674921725078275 0.3276724065341866
106 0.6800805096286061 0.3305531083114249
109 0.677154856803393 0.32927093072939106
112 0.6862312210774759 0.33452967891286295
115 0.6878794997188232 0.33261187038217077
118 0.6911004482085407 0.33511402960705927
121 0.6899211506920759 0.33498099403692383
124 0.7030867683041596 0.33917995037575194
127 0.6840252222863167 0.33159306908656283
130 0.7090959244190056 0.3413943276706018
133 0.7043927684820264 0.33743335591490403
136 0.7022499276380348 0.3373984987104215
139 0.7098026416299774 0.3417326700504682
142 0.709806986543428 0.3411853421189651
145 0.7004974280707704 0.33919612143767025
148 0.7075727642050198 0.3410444023785891
151 0.708421768707483 0.34041577800341316
154 0.7165174927424106 0.34293900762062335
157 0.7124069582671239 0.34221326307560906
160 0.7186848985640982 0.3447465072406974
163 0.7194276862872983 0.3463668642462567
166 0.7187262711747407 0.34594129412558366
169 0.7168176586474979 0.3457184201566224
172 0.7194665733663473 0.344482144007559
175 0.7259417275052006 0.3502503590070684
178 0.7227934257790232 0.34932250367815204
181 0.725886946898035 0.34897611078501545
184 0.7347561604289102 0.35304425369820586
187 0.7263010386327416 0.3478721341581641
190 0.7269049086054631 0.34878715170844826
193 0.7298588640275387 0.3536960605926123
196 0.7267479240124255 0.3517172827007631
199 0.734146849528622 0.35650422325807896
200 0.7334242640838836 0.35599541799901196

}{\map}

    \addplot[wid]      table[x=xxx,  y=id]   \map;     
        \leg{\scriptsize With same-ID constraint}
    \addplot[woid]      table[x=xxx,  y=noid]   \map; 
        \leg{\scriptsize Without same-ID constraint}
    
\end{axis}
\end{tikzpicture} \\[-0.3cm]
}{%
  \caption{\footnotesize \textbf{Evolution of the matching quality} measured as the ratio of matches coming from the positive pair over all pairs with the query in each training tuple.}
  \label{fig:trainingmatch}%
}
\end{floatrow}
\vspace*{-0.35cm}
\end{figure}

\mypartight{Matching constraints ablation.} Table~\ref{tab:match} reports the impact of the different matching constraints: nearest neighbor, reciprocity, Lowe ratio test and constraint on \mname ID.
Adding reciprocity and Lowe's ratio test significantly improves performance, which indicates that it reduces the number of incorrect matches.
Keeping all pairs with the same ID yields lower performance: the ID constraint alone is not enough.
One possible explanation is that two images from the same landmark may differ significantly, \eg when some parts are visible in only one of the two images.
In that case, the other constraints allow features attending non-overlapping regions of a positive image pair to be excluded from $\gP$. 
Finally, combining the 
selective ID constraint with the others yields the best performance.

In order to better understand the impact of the \sfeature ID constraint, a constraint only applicable for \sfeatures due to their ordered nature, we measure the quality of selected matches during training with and without it. 
Since there is no ground-truth for such localized matches on landmark retrieval datasets, we measure instead the ratio of matches coming from the positive pair, over all matches (from the positive and all negatives).
Ideally, a minimal number of matches should come from the negatives, hence this ratio should be close to 1.
In Figure~\ref{fig:trainingmatch} we plot this match ratio for all epochs; we observe that it is significantly higher when using the \mname ID constraint.

\begin{figure}[tb]
\begin{floatrow}
\ttabbox[1\linewidth]{%
  \vspace*{-2.35cm}
  \resizebox{1.\linewidth}{!}{
            \raggedleft
            \begin{tabular}{l  ll  c cc cc}
\toprule
\multirow{2}{*}{$\mathcal{L}_{global}$} & 
\multirow{2}{*}{$\mathcal{L}_{attn}$} &   
\multirow{2}{*}{$\mathcal{L}_{super}$} & 
SfM-120k & \multicolumn{2}{c}{\ROxford} & \multicolumn{2}{c}{\RParis} \\
        & &   & val & med & hard & med & hard \\
        
\cmidrule(l){1-3} \cmidrule(l){4-4} \cmidrule(l){5-6} \cmidrule(l){7-8}

$\checkmark$ & & & 79.0 & 64.3 & 38.0 & 75.4 & 51.7 \\ 
$\checkmark$ & $\checkmark$ & & 87.7 & 75.8 & 51.2 & 79.0 & 57.0 \\ 
 $\checkmark$ & $\checkmark$ & $\checkmark$ & 88.4 & 79.0 & 57.2 & 83.0 & 65.6 \\ 
 
\cmidrule(l){1-3} \cmidrule(l){4-4} \cmidrule(l){5-6} \cmidrule(l){7-8}

   & & $\checkmark$           &     61.7 &	59.3 &	32.8 & 69.9 & 47.3 \\
 & $\checkmark$ & $\checkmark$ &         \bf{89.7} & \bf{81.9} & \bf{61.5}	& \bf{85.3} & \bf{70.1} \\

\bottomrule
\end{tabular}
\label{tab:losses}
        }
}
{
\caption{\textbf{Ablation on loss components}: Impact of removing $\mathcal{L}_{\text{attn}}$, using either a global loss $\mathcal{L}_{\text{global}}$ or a loss directly on \mname $\mathcal{L}_{\text{super}}$ or a combination of both.\vspace*{-1.9cm}}%
 \label{tab:loss_ablation}
}
\hcapbfigbox{%
    \includegraphics[width=0.40\linewidth]{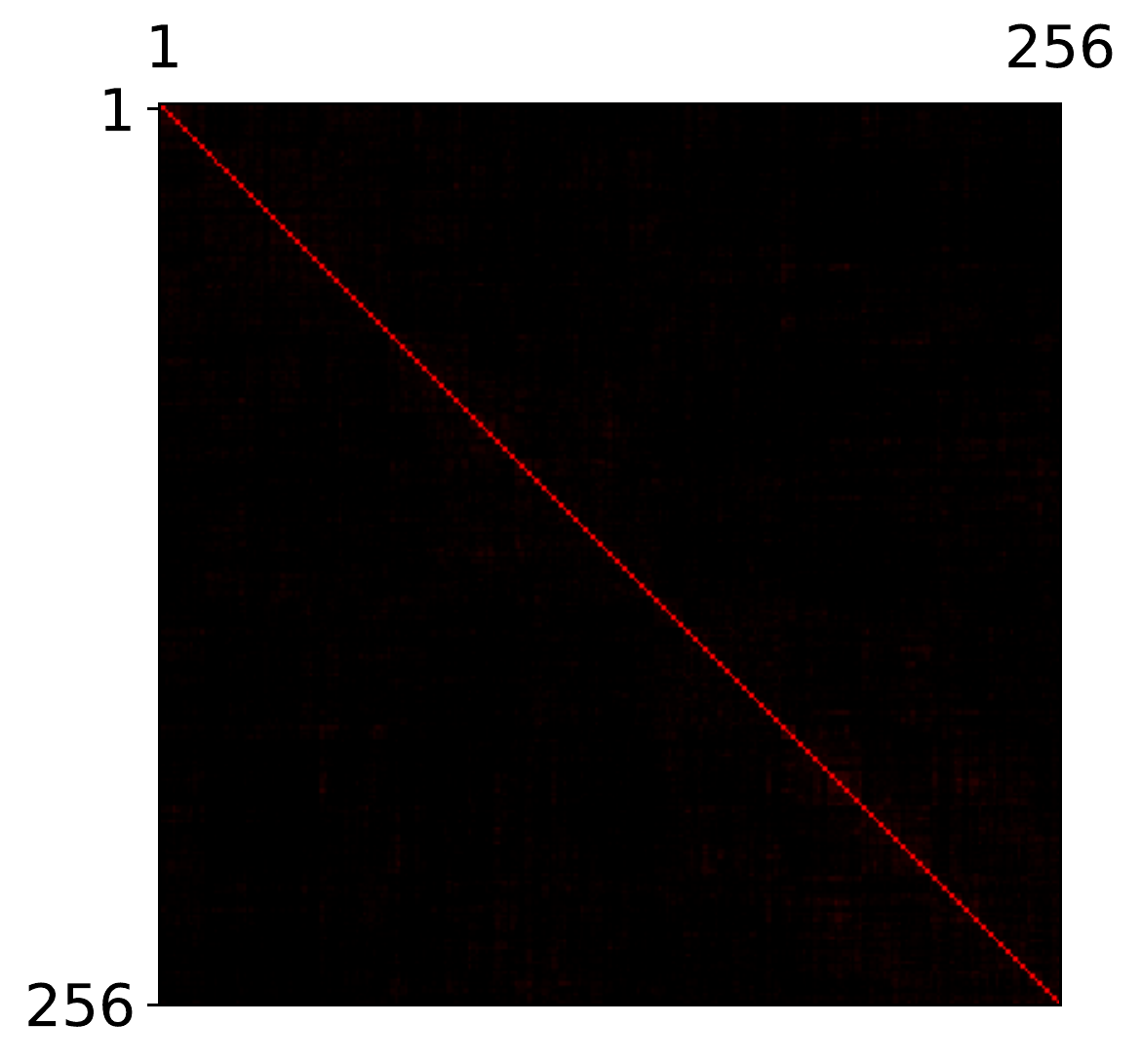}
    \includegraphics[width=0.56\linewidth]{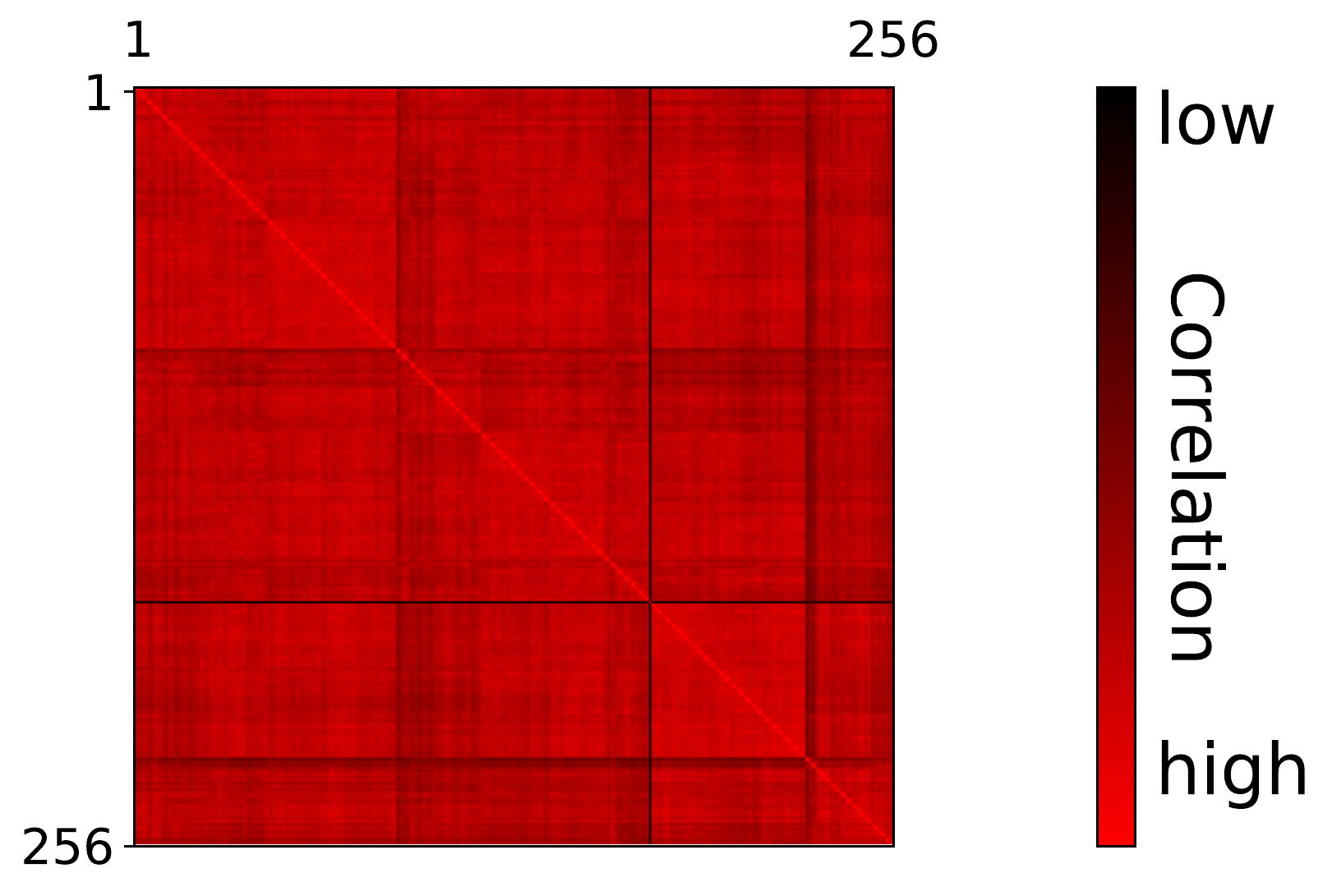}
}{%
\vspace{-0.3cm}
 \caption{\textbf{Impact of  $\mathcal{L}_{\text{attn}}$ on the correlation matrix between attention maps}  (the darker, the lower is the correlation) at the last iteration of \lit when training with (left) and without (right). This is averaged over the 70 queries of \ROxford.}\label{fig:correlation}%
}
\end{floatrow}
\vspace*{-0.35cm}
\end{figure}

\mypartight{Training losses ablation.} We study the impact of the different training losses in Table~\ref{tab:loss_ablation}. We start from a global loss similar to HOW (first row). We then add the decorrelation loss (second row) and observe a clear gain. It can be explained by the fact that without this loss, \sfeatures tend to be redundant (see the correlation matrices in Figure~\ref{fig:correlation}).
Adding the loss operating directly on the \mname further improves performance (third row).
Next, we remove the global loss and keep only the loss on \mname alone (fourth row) or with the decorrelation loss (last row). The latter performs best (fourth vs last row).
Figure~\ref{fig:correlation} displays the correlation matrix of \sfeatures attention maps with and without $\mathcal{L}_{\text{attn}}$. Without it, we observe that most \mname have correlated attentions.
In contrast, training with $\mathcal{L}_{\text{attn}}$ leads to uncorrelated attention maps.
This is illustrated in Figure~\ref{fig:visualizing_sfeatures} which shows several attention maps focusing on different areas.

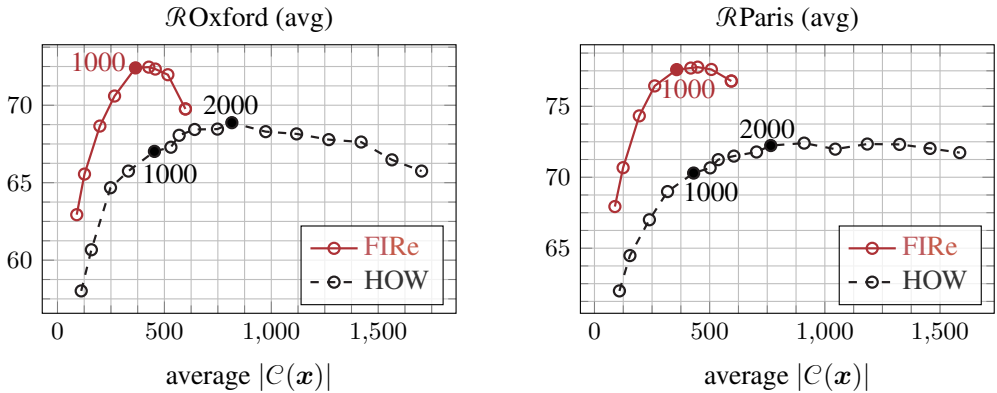
\begin{figure}[t]
\begin{center}
\resizebox{\linewidth}{!}{
    \hspace*{-0.5cm}
    \begin{subfigure}{.49\linewidth}
        \centering
        \begin{tikzpicture}
\begin{axis}[%
  height=5cm, 
   xlabel={average $|\mathcal{C}(\vx)|$},
  minor tick num=3,
  legend pos=south east,
    title={\ROxford (avg)},
            title style={yshift=-5pt},
  ]

\pgfplotstableread{
nfeats xbaseval xbaseox xbasepar ybaseval ybaseoxavg ybaseparavg  xoursval xoursox xourspar yoursval  yoursoxavg yoursparavg  
200 110.6890837	110.9841778	108.346093 73.05	58.015	61.99 85.84249055	90.04986982	88.67953179 81.06	62.935	67.935
400 158.0255008	158.4636491	153.8555837 77.47	60.67	64.48 120.6713056	126.2984178	124.3187283 83.45	65.555	70.675
600 245.7272947	247.9719607	238.4152167 82.23	64.675	67.005 189.0612984	198.2695774	194.618317 87.15	68.655	74.325
800 327.9332314	331.854997	317.198355 84.18	65.75	68.99 254.1856649	266.9405167	261.4708953 88.95	70.595	76.435
1000 445.2857373	453.4352093	430.4623537 85.78	67.02	70.295 347.1687716	364.8590026	356.8740905 90.31	72.41	77.585
1200 519.4181482	530.4356099	502.1787409 87.22	67.29	70.66 406.8280106	427.138594	418.0639038 90.38	72.46	77.695
1400 556.8782882	568.9857801	537.2674786 87.66	68.06	71.24 436.2373904	457.4324054	448.17795 90.47	72.335	77.76
1600 626.5781514	641.5930302	605.2753875 87.65	68.435	71.49 494.7869037	516.2467454	507.4163239 90.75	71.97	77.595
1800 727.8111978	748.189265	703.3933882 88.44	68.46	71.78 583.2631325	598.3306629	594.1403037 90.44	69.76	76.78
2000 793.1785858	815.4312037	765.219709 88.59	68.875	72.235 583.2631325	598.3306629	594.1403037 90.44	69.76	76.78
2500 944.8674282	973.6853595	910.1904461 89.4  68.295    72.405   nan nan nan   nan nan nan 
3000 1086.92575	    1118.15021	1045.214805 89.6  68.15     71.98   nan nan nan   nan nan nan 
3500 1234.912477	1268.622872	1184.211958 89.65 67.78     72.34   nan nan nan   nan nan nan 
4000 1385.286381	1419.973363	1324.301487 89.32 67.625    72.315  nan nan nan   nan nan nan 
4500 1530.464243	1564.169037	1456.49826  89.37 66.475    72.03  nan nan nan   nan nan nan 
5000 1674.861717	1702.774084	1585.542866 88.5  65.75     71.735   nan nan nan   nan nan nan 
}{\map}

    \addplot[oursoxf]      table[x=xoursox,  y=yoursoxavg]   \map; 
    \leg{\mnameshort}
    \addplot[howoxf]      table[x=xbaseox,  y=ybaseoxavg]   \map; 
    \leg{HOW}
    
    \node[] at (axis cs: 525,65.6) {1000};
    \node[] at (axis cs: 810,70) {2000};
    \node[\ourscolor] at (axis cs: 190,72.8) {1000};
    
    \draw[fill] (axis cs: 453.4352093,67.02) circle (2pt);
    \draw[fill] (axis cs: 815.4312037,68.875) circle (2pt);
    \draw[fill,\ourscolor] (axis cs: 364.8590026,72.41) circle (2pt);
    
\end{axis}
\end{tikzpicture}
        \label{fig:ablation_nfeat_all_scales_par}
    \end{subfigure}%
    \hfill
    \begin{subfigure}{.49\linewidth}
        \centering
        \begin{tikzpicture}
\begin{axis}[%
  height=5cm, 
   xlabel={average $|\mathcal{C}(\vx)|$},
  minor tick num=3,
  legend pos=south east,
    title={\RParis  (avg)},
            title style={yshift=-5pt},
  ]

\pgfplotstableread{
nfeats xbaseval xbaseox xbasepar ybaseval ybaseoxavg ybaseparavg  xoursval xoursox xourspar yoursval  yoursoxavg yoursparavg  
200 110.6890837	110.9841778	108.346093 73.05	58.015	61.99 85.84249055	90.04986982	88.67953179 81.06	62.935	67.935
400 158.0255008	158.4636491	153.8555837 77.47	60.67	64.48 120.6713056	126.2984178	124.3187283 83.45	65.555	70.675
600 245.7272947	247.9719607	238.4152167 82.23	64.675	67.005 189.0612984	198.2695774	194.618317 87.15	68.655	74.325
800 327.9332314	331.854997	317.198355 84.18	65.75	68.99 254.1856649	266.9405167	261.4708953 88.95	70.595	76.435
1000 445.2857373	453.4352093	430.4623537 85.78	67.02	70.295 347.1687716	364.8590026	356.8740905 90.31	72.41	77.585
1200 519.4181482	530.4356099	502.1787409 87.22	67.29	70.66 406.8280106	427.138594	418.0639038 90.38	72.46	77.695
1400 556.8782882	568.9857801	537.2674786 87.66	68.06	71.24 436.2373904	457.4324054	448.17795 90.47	72.335	77.76
1600 626.5781514	641.5930302	605.2753875 87.65	68.435	71.49 494.7869037	516.2467454	507.4163239 90.75	71.97	77.595
1800 727.8111978	748.189265	703.3933882 88.44	68.46	71.78 583.2631325	598.3306629	594.1403037 90.44	69.76	76.78
2000 793.1785858	815.4312037	765.219709 88.59	68.875	72.235 583.2631325	598.3306629	594.1403037 90.44	69.76	76.78
2500 944.8674282	973.6853595	910.1904461 89.4  68.295    72.405   nan nan nan   nan nan nan 
3000 1086.92575	    1118.15021	1045.214805 89.6  68.15     71.98   nan nan nan   nan nan nan 
3500 1234.912477	1268.622872	1184.211958 89.65 67.78     72.34   nan nan nan   nan nan nan 
4000 1385.286381	1419.973363	1324.301487 89.32 67.625    72.315  nan nan nan   nan nan nan 
4500 1530.464243	1564.169037	1456.49826  89.37 66.475    72.03  nan nan nan   nan nan nan 
5000 1674.861717	1702.774084	1585.542866 88.5  65.75     71.735   nan nan nan   nan nan nan 
}{\map}

    \addplot[oursoxf]      table[x=xourspar,  y=yoursparavg]   \map; 
    \leg{\mnameshort}
    \addplot[howoxf]      table[x=xbasepar,  y=ybaseparavg]   \map; 
    \leg{HOW}

    \node[] at (axis cs: 505,69) {1000};
    \node[] at (axis cs: 750,73.4) {2000};
    \node[\ourscolor] at (axis cs: 405,76.2) {1000};

    \draw[fill] (axis cs: 430.4623537,70.295) circle (2pt);
    \draw[fill] (axis cs: 765.219709,72.235) circle (2pt);
    \draw[fill,\ourscolor] (axis cs: 356.8740905,77.585) circle (2pt);

\end{axis}
\end{tikzpicture}
        \label{fig:ablation_nfeat_all_scale_ox}
    \end{subfigure}%
    }
\end{center}
\vspace{-0.5cm}
\caption{\textbf{Performance versus memory} for HOW and \fire. 
The x-axis shows the average number of clusters per image used in ASMK (proportional to memory usage). 
We vary the number of features extracted per image before aggregation in $\{200, \ldots, 2000, 2500, \ldots, 5000\}$; solid markers denote the commonly used settings ($1000$/$2000$). \fire has at most 1,792 features (256 per scale).
\vspace*{-0.7cm}}
\label{fig:ablation_nfeat}
\end{figure}

\mypartight{Varying the number of features at test time.} 
Figure~\ref{fig:ablation_nfeat} compares HOW~\citep{how} with our approach, as we vary the number of local features / \sfeatures.
The x-axis shows the average number of clusters used in ASMK for the database images, \ie, which is proportional to the average memory footprint of an image. We observe that our approach does not only significantly improve accuracy compared to HOW, but also requires overall less memory. Notably, \fire matches the best performance of HOW with a memory footprint reduced by a factor of $4$. For both methods, performance goes down once feature selection no longer discards background features. The gain in terms of performance is even larger  when considering a single scale at test time (see Appendix~\ref{app:ablation_nfeat_onescale}). 

\begin{figure}[t]
\centering
    \resizebox{\linewidth}{!}{
            \begin{subfigure}{.33\linewidth}
        \centering
        \begin{tikzpicture}
            \begin{axis}[
                ylabel={features (\%)},
                ybar=2pt,
                every x tick label/.append style={font=\tiny},
                title style={align=center, font=\scriptsize,
                yshift=-5pt}, 
                title={Scale of selected features (\%)},
                symbolic x coords={0.25, 0.35, 0.5, 0.7, 1.0, 1.4, 2.0},
                xtick=data,
                xtick align=inside,
                legend pos=north west,
                legend style={nodes={scale=0.8, transform shape}}]
            
                \addplot[ybar, fill=\howcolor, bar width=4pt]
                coordinates {
                    (2.0,  0.3518029522531492)
                    (1.4,  0.2385755306408092)
                    (1.0,  0.16225848808853874)
                    (0.7,  0.10816270579917175)
                    (0.5,  0.06894362437412521)
                    (0.35, 0.043879774324343826)
                    (0.25, 0.026376924519862056)
                };\addlegendentry{HOW};
                
                                \addplot[ybar, fill=\ourscolor, bar width=4pt]
                coordinates {
                    (2.0,  0.13437142857142856)
                    (1.4,  0.14255714285714285)
                    (1.0,  0.14724285714285715)
                    (0.7,  0.14597142857142859)
                    (0.5,  0.14477142857142858)
                    (0.35, 0.14082857142857144)
                    (0.25, 0.14425714285714286)
                };\addlegendentry{\fire};
            \end{axis}
        \end{tikzpicture}
        \label{fig:bars_1_1}
    \end{subfigure}%
    \begin{subfigure}{.33\linewidth}
        \centering
        \begin{tikzpicture}
            \begin{axis}[
                ylabel={features (\%)},
                every x tick label/.append style={font=\tiny},
                ybar=2pt,
                title style={align=center, font=\scriptsize,
                yshift=-5pt}, 
                title={Features selected per scale(\%)},
                symbolic x coords={0.25, 0.35, 0.5, 0.7, 1.0, 1.4, 2.0},
                xtick=data,
                xtick align=inside
                ]
            
                \addplot[ybar, fill=\howcolor, bar width=4pt]
                coordinates {
                    (2.0, 0.0743906048336654)
                    (1.4, 0.10035080904577512)
                    (1.0, 0.13549988552699754)
                    (0.7, 0.17889360889695002)
                    (0.5, 0.22406677921590695)
                    (0.35, 0.28356956359567326)
                    (0.25, 0.3315799020497007)
                };
                
                                \addplot[ybar, fill=\ourscolor, bar width=4pt]
                coordinates {
                    (2.0, 0.5248883928571428)
                    (1.4, 0.5568638392857143)
                    (1.0, 0.5751674107142857)
                    (0.7, 0.5702008928571428)
                    (0.5, 0.5655133928571429)
                    (0.35,0.5501116071428571)
                    (0.25,0.5635044642857143)
                };
            \end{axis}
        \end{tikzpicture}
        \label{fig:bars_1_3}
    \end{subfigure}%
     \begin{subfigure}{.33\linewidth}
        \centering
        \begin{tikzpicture}
            \begin{axis}[
                ylabel={\# features},
                every x tick label/.append style={font=\tiny},
                ymin=0,
                ybar=2pt,              
                ytick={1000,2000,3000,4000},
                yticklabels={1K, 2K, 3K, 4K},
                title style={align=center, font=\scriptsize,
                yshift=-5pt}, 
                title={Total nb. features per scale},
                symbolic x coords={0.25, 0.35, 0.5, 0.7, 1.0, 1.4, 2.0},
                xtick=data,
                xtick align=inside,
                ]
            
                \addplot[ybar, fill=\howcolor, bar width=4pt]
                coordinates {
                    (2.0, 4682.042857142857)
                    (1.4, 2353.7428571428572)
                    (1.0, 1185.557142857143)
                    (0.7, 598.6)
                    (0.5, 304.62857142857143)
                    (0.35, 153.2)
                    (0.25, 78.75714285714285)
                }; 
                
                                \addplot[ybar, fill=\ourscolor, bar width=4pt]
                coordinates {
                    (2.0,  256)
                    (1.4,  256)
                    (1.0,  256)
                    (0.7,  256)
                    (0.5,  256)
                    (0.35, 256)
                    (0.25, 256)
                };
                
            \end{axis}
        \end{tikzpicture}
        \label{fig:bars_1_2}
    \end{subfigure}%
    
    }
    
    \vspace{-0.3cm}
    
    \caption{ \textbf{Statistics on the feature selected across scales} for HOW and \fire, averaged over the 70 queries from \ROxford. \textit{Left:}  Among the 1000 selected features, we show the percentage coming from each scale.  \textit{Middle:} For each scale, we show the ratio of features that are selected.  \textit{Right:} Total number of features per scale; we extract $\nsfeat=256$ \sfeatures regardless of scale.    \vspace*{-0.35cm}
}
    \label{fig:scales_barplot}
\end{figure}
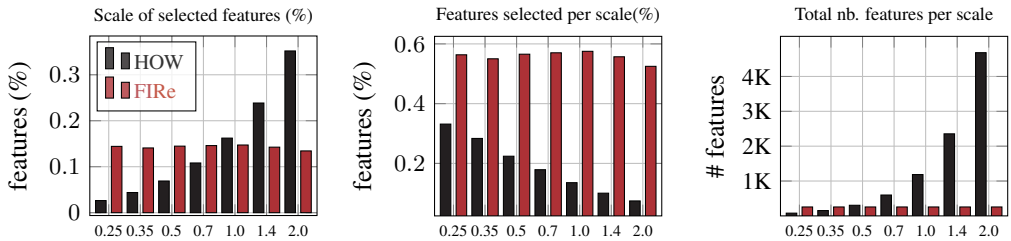

\mypartight{Statistics on the number of selected features per scale.} 
The left plot of Figure~\ref{fig:scales_barplot} shows the percentage of the 1000 selected features that comes from each scale, for HOW and \fire. For HOW most selected features come from the higher resolution inputs; by contrast, selected \sfeatures are almost uniformly distributed across scales. 
Interestingly, Figure~\ref{fig:scales_barplot} (middle) shows that HOW keeps a higher percentage of the features coming from coarser scales. Yet, the final feature selection for HOW is still dominated by features from higher scales, due to the fact that the number of local features significantly increases with the input resolution (see Figure~\ref{fig:scales_barplot} right).

\subsection{Comparison to the state of the art}
\label{sub:sota}

We compare our method to the state of the art in Table~\ref{tab:sota}. All reported methods are trained on SfM-120k 
for a fair comparison.\footnote{\scriptsize The GLDv2-clean dataset~\citep{gldv2_clean} is sometimes used for training. Yet, there is significant overlap between its training classes and the \ROxford and \RParis query landmarks. This departs from standard image retrieval evaluation protocols. See Appendix~\ref{app:gldv2} for details.}
First, we observe that methods based on global descriptors~\citep{rmac,apgem} compared with a $\text{L}2$ 
distance tend to be less robust than methods based on local features~\citep{delf,how}. DELG~\citep{delg} 
shows better performance, owing to a re-ranking step based on local features, at the cost of re-extracting local features for the top-ranked images of a given query, given that local features would take too much memory to store.
Our \fire method outperforms DELF~\citep{delf} as well as HOW~\citep{how} by a significant margin when extracting 1000 features per image. Importantly, our approach also requires less memory, as it uses fewer ASMK clusters  per image, as shown in Figure~\ref{fig:ablation_nfeat}: for the whole \Rdis set, HOW uses 469 clusters per image while \fire uses only 383, on average.
\section{Related Work}
\label{sec:related}

\mypartight{Image descriptors for retrieval.}
The first approaches for image retrieval were based on handcrafted local descriptors and bag-of-words representations borrowed from text retrieval~\citep{sivic2003video, csurka2004visual}, or other aggregation techniques like Fisher Vectors~\citep{FVretrieval}, VLAD~\citep{vlad} or ASMK~\citep{asmk}.
First deep learning techniques were extracting a global vector per image either directly or by aggregating local activations~\citep{dir, cirtorch} and have shown to highly outperform handcrafted local features, see~\citep{csurka2018handcrafted} for a review.
Methods that perform matching or re-ranking using CNN-based local features are currently the state of the art in the area~\citep{delf, teichmann2019detect, delg, how}. They are able to learn with global (\ie image-level) annotations. 
Most of them
use simple variants of attention mechanisms~\citep{delf, ng2020solar} or simply the feature norm~\citep{how} to weight local activations.

  \begin{table}
    \centering
    \resizebox{\linewidth}{!}{
    \begin{tabular}{llrH cc cc cc cc}
    \toprule
         \multirow{2}{*}{method} & \multirow{2}{*}{FCN} & Mem & \multirow{2}{*}{training data} & \multicolumn{2}{c}{\ROxford} & \multicolumn{2}{c}{\ROxford+\Rdis} & \multicolumn{2}{c}{\RParis} & \multicolumn{2}{c}{\RParis+\Rdis} \\
				 &&(GB)& & med & hard & med & hard & med & hard & med & hard\\
    
    \cmidrule(l){1-1}  \cmidrule(l){2-2}  \cmidrule(l){3-3}  \cmidrule(l){5-6}  \cmidrule(l){7-8}  \cmidrule(l){9-10}  \cmidrule(l){11-12} 
    
    \multicolumn{12}{l}{\textit{Global descriptors}} \vspace{2pt} \\
	 ~~RMAC~{\tiny\citep{rmac}}     & R101 & 7.6    & SfM-120k & 60.9 & 32.4 & 39.3  & 12.5 & 78.9 & 59.4 & 54.8  & 28.0 \\
	 ~~\hspace*{-0.2cm}\begin{tabular}{l}AP-GeM$^\ddagger$~{\tiny\citep{apgem}}\end{tabular}   & R101 & 7.6  & SfM-120k & 67.1 & 42.3 & 47.8 & 22.5 & 80.3 & 60.9 & 51.9 & 24.6 \\
	 
	 ~~\hspace*{-0.2cm}\begin{tabular}{l}GeM+SOLAR~{\tiny\citep{ng2020solar}}\end{tabular} & R101   & 7.6 & SfM-120k & 69.9 & 47.9 & 53.5  & 29.9 & 81.6 & 64.5 & 59.2  & 33.4 \\
	 
    \cmidrule(l){1-1}  \cmidrule(l){2-2}  \cmidrule(l){3-3}  \cmidrule(l){5-6}  \cmidrule(l){7-8}  \cmidrule(l){9-10}  \cmidrule(l){11-12}

    \multicolumn{12}{l}{\textit{Global descriptors + reranking with local features}} \vspace{2pt} \\                  
    
    ~~DELG~{\tiny\citep{delg}}
	& R50     & 7.6 & SfM-120k & 75.1 & 54.2 & 61.1  & 36.8 & 82.3 & 64.9 &  60.5 &  34.8 \\
    ~~DELG~{\tiny\citep{delg}}
	& R101    & 7.6 & SfM-120k & $\textit{78.5}$ & $\textit{59.3}$ & $62.7$  & $\textit{39.3}$ & $\textit{82.9}$ & $\textit{65.5}$ & $\textit{62.6}$ & $\textit{37.0}$  \\
    
    \cmidrule(l){1-1}  \cmidrule(l){2-2}  \cmidrule(l){3-3}  \cmidrule(l){5-6}  \cmidrule(l){7-8}  \cmidrule(l){9-10}  \cmidrule(l){11-12}

    \multicolumn{12}{l}{\textit{Local features + ASMK matching~(max. $1000$ features per image)} } \vspace{2pt} \\                                                                           
    ~~DELF~{\tiny\citep{delf}}
	& R50$^-$     & 9.2 & SfM-120k & 67.8 & 43.1 & 53.8  & 31.2 & 76.9 & 55.4 & 57.3  & 26.4  \\
	 ~~\hspace*{-0.2cm}\begin{tabular}{l}{\small DELF-R-ASMK}~{\tiny\citep{teichmann2019detect}}\end{tabular}                                 & R50$^-$  & 27.4  & SfM-120k & 76.0 & 52.4 & \textit{\underline{64.0}}  & \underline{38.1} & \underline{80.2} & 58.6 & \underline{59.7}  & 29.4  \\
    ~~HOW~{\tiny\citep{how}}  & R50$^-$ & 7.9 & SfM-120k & \underline{78.3} & \underline{55.8} & 63.6  & 36.8 & 80.1 & \underline{60.1} & 58.4  & \underline{30.7}  \\ 
~~\textbf{\fire} (ours) & R50$^-$ & \bB{6.4} & SfM-120k &    \bB{81.8} & \bB{61.2} & \bB{66.5} &  \bB{40.1} & \bB{85.3}& \bB{70.0}& \bB{67.6} & \bB{42.9}  \\
    \multicolumn{3}{l}{\quad~~~\small{\emph{(standard deviation over 5 runs)}}}  && \mpm{0.6} & \mpm{1.0} & \mpm{0.8} & \mpm{1.1} & \mpm{0.4} & \mpm{0.6} & \mpm{0.7} & \mpm{0.8} \\
    \multicolumn{3}{l}{\quad~~~\small{\emph{(mAP gains over HOW)}}} & &  \bI{3.5} & \bI{5.4} & \bI{2.9} & \bI{3.3} & \bI{5.2} & \bI{9.9} & \bI{9.2}  & \bI{12.2} \\
    \bottomrule
    \end{tabular}
    }
    
    \vspace{-0.3cm}
    
    \caption{ \textbf{Comparison to the state of the art.} All models are trained on SfM-120k. FCN denotes the fully-convolutional network backbone, with R50$^-$ denoting a ResNet-50 without the last block. Memory is reported for the image representation of the full \Rdis set (without counting local features for the global descriptors + reranking methods).
    $^\ddagger$ result from~\citep{how}.
    \textbf{Bold} denotes best performance, \underline{underlined} second best among methods using ASMK, \textit{italics} second best overall. \vspace{-0.7cm} }
    \label{tab:sota}
\end{table}

\mypartight{Low-level features aggregation.}
Aggregating local features into 
regional features has a long 
history. 
Crucial to the success of traditional 
approaches, popular methods include selecting discriminative patches in the input image \citep{singh2012unsupervised, gordo2015supervised},  regressing saliency map scores at different resolutions \citep{jiang2013salient}, aggregating SIFT descriptors with coding and pooling schemes \citep{boureau2010learning}, or mining frequent patterns in sets of SIFT features \citep{singh2012unsupervised}. More recently, \citet{patch-net-vlad} introduced a multi-scale fusion of patch features. 

\mypartight{Supervision for local features.}
Several works provide 
supervision at the level of local features in the context of contrastive learning.
\cite{xie2021propagate} and \cite{chen2021multisiam} obtain several views of an input image using data augmentations with known pixel displacements. 
Similarly, \cite{liu2020self} train a model to predict the probability for a pair of local features to match, evaluated using known displacements. 
\cite{wang2020learning} and \cite{zhou2021patch2pix} obtain local supervision by relying on epipolar coordinates and relative camera poses. 
Positive pairs in image retrieval depart from these setups, as pixel-level correspondences cannot be known.  
To build matches, 
\cite{wang2021dense} use a standard nearest neighbor algorithm to build pairs of features, similarly to our approach, but without the use of filtering which 
is critical to our final performance. Using the current model predictions to define targets is reminiscent of modern self-supervised learning approaches which learn without any label~\citep{caron2021emerging,BYOL}. The additional filtering step can be seen as a way to keep only the most reliable model predictions, similar to \textit{self-distillation} as in \eg \citet{fixmatch}.

\mypartight{Attention modules.} 
Our \lit module is an iterative variant of standard attention~\citep{attention, vaswani2017attention},
adapted to map a variable number of input features to an ordered set of $N$ output features, similar to~\citet{set-transformer}. The Perceiver model ~\citep{perceiver} has demonstrated the flexibility of such constructions by using it to scale attention-based deep networks to large inputs.
Our design was heavily inspired by \emph{slot-attention}~\citep{slotattention}, but has some key differences that enable us to achieve high performance in more complex visual environments: a) unlike the slot attention which initializes its slots with i.i.d sampling, we \textit{learn} the initial 
templates and therefore define an ordering on the output set, a crucial property for selecting promising matches; b) we replace the recurrent network gates with a 
residual connection across iterations. These modifications, together with the attention decorrelation loss 
enable our module to go from a handful of object-oriented slots to a much larger set of 
output
features.
For object detection,~\cite{DETR} rely on a set of learned \textit{object queries} to initialize a stack of transformer layers. Unlike ours, their module is not recurrent;  Appendix~\ref{app:ablate} experimentally shows substantial benefits from 
applying \lit $T$ times, with weight sharing, to increase model flexibility without extra parameters.
\vspace{-0.1cm}

\section{Conclusions}
\label{sec:conclusions}

\vspace{-0.1cm}

We present an approach 
that aggregates local features into \sfeatures for image retrieval, a task that has up to now been dominated by approaches that work at the local feature level. 
We design an attention mechanism that outputs an ordered set of such features that 
are more discriminative and expressive than local features.
Exploiting their ordered nature and without any extra supervision, we present a loss working directly on 
the proposed features.
Our method not only significantly improves performance, but also requires less memory, a crucial requirement for scalable image retrieval.

{\small
\bibliographystyle{iclr2022_conference}
\bibliography{biblio}}

\newpage
\appendix
\renewcommand \thepart{}
\renewcommand \partname{}

\part{Appendix}

\parttoc

In this appendix, we present additional ablations (Appendix~\ref{app:ablations}), a deeper analysis of our framework (Appendix~\ref{app:analysis}) as well as results of the application of our \fire framework for image retrieval to the task of visual localization (Appendix~\ref{app:localize}). We briefly summarize the findings in the following paragraphs.

\mypartight{Ablations.} We study the impact of some hyper-parameters of our model, namely the size $N$ of the set of \sfeatures, and the number of iterations $T$ in the \lit module, in Appendix~\ref{app:ablate}. We show that $256$ \sfeatures and $6$ iterations offer the best trade-off between performance and computational cost. We also show in Appendix~\ref{app:nnegative} that we obtain further performance gains of over 1\% by increasing the number of negatives to $10$ or $15$. We then study the impact of replacing the residual connection inside the \lit module by a recurrent network (Appendix~\ref{app:update}). In Appendix~\ref{app:matching_extended} we present an extended version of Table~\ref{tab:match} with further matching constraints and  we finally study the impact of the template initialization on performance in Appendix~\ref{app:template_init}.

\mypartight{Properties of \sfeatures.} 
We show single-scale results in Appendix~\ref{app:ablation_nfeat_onescale}.
In Appendix~\ref{app:consistency}, we display the attention maps of \sfeatures, at different scales for a fixed \sfeature ID.  We further study the amount of redundancy in \sfeatures, compared to local features, in Appendix~\ref{app:redundancy}. Next, we verify in Appendix~\ref{app:alltrained} that all \sfeatures receive training signal, as a sanity check. We discuss the case of applying a loss directly on local features in Appendix~\ref{app:local} and give details about the pretraining on ImageNet in Appendix~\ref{app:pretrain}. In Appendix~\ref{app:speed} we report the average extraction time for the proposed Super-features, while in Appendix~\ref{app:gldv2} we discuss the fact that there is an overlap between the queries from the common \ROxford and \RParis datasets and the Google Landmarks-v2-clean dataset that is commonly used as training set for retrieval on \ROxford and \RParis.

\mypartight{Application to visual localization.} We further evaluate our model using a visual localization setting on the Aachen Day-Night v1.1 dataset~\citep{aachen} in Appendix~\ref{app:localize}. To do so, we leverage a retrieval + local feature matching pipeline, and show that it is beneficial to use our method, especially in hard settings.
\newpage

\section{Additional ablations}
\label{app:ablations}

\subsection{Impact of the iteration and the number of templates hyper-parameters}
\label{app:ablate}
In this section, we perform ablations on the number of \mname $N$ and the number of iterations $T$ in the \fimlong.
We first study the impact of the number $N$ of \mname extracted for each scale of each image in Figure~\ref{fig:ablation_N}.
We observe that the best value is 256 on both the validation and test sets (\ROxford and \RParis).
We then study the impact of the number of iterations T in 
Figure~\ref{fig:ablation_T}.
While the performance decreases when doing 2 or 3 iterations compared to just 1, a better performance is reached for 6 iterations, after which the performance saturates while requiring more computations. We thus use $T=6$.

\begin{figure}[ht]
\begin{center}
\begin{center}
\resizebox{.35\linewidth}{!}
{
    \begin{tikzpicture}
    \begin{axis}[%
        hide axis,
        xmin=10, xmax=50,
        ymin=0,ymax=0.4,
        legend columns=-1,
        legend style={draw=white!15!black,legend cell align=left}
        ]
        \addlegendimage{val} \addlegendentry{\val}
        \addlegendimage{oxmed} \addlegendentry{\ROxford (avg)}
        \addlegendimage{pamed} \addlegendentry{\RParis (avg)}
    \end{axis}
    \end{tikzpicture}
    }
    \end{center}
\vspace{10pt}
    \begin{subfigure}{.5\linewidth}
        \centering
        \begin{tikzpicture}
\begin{axis}[%
  height=4.5cm, 
  xlabel={number of templates (N)},
  xtick = {64, 128, 256, 512, 1024},
  xticklabels = {64, 128, 256, 512, 1024},
  ylabel={mAP},   minor tick num=3,
    legend pos=outer north east,
    xmode=log,
  ]

\pgfplotstableread{
n val oxmed oxhard pamed pahard ox par
64 83.9 73.8 47.2 80.1 60.9 60.5 70.5
128 88.6 78.9 55.8 83.9 67.5 67.4 75.7
256 89.7 81.8 61.2 85.3	70.0 71.5 77.6
512 87.1 79.8 56.3 83.1 66.4 68.0 74.7
1024 82.5 74.1 48.8 78.0 58.3 61.5 68.1
}{\map}

    \addplot[val]      table[x=n,  y=val]   \map; 
    \addplot[oxmed]      table[x=n,  y=ox]   \map; 
    \addplot[pamed]      table[x=n,  y=par]   \map;

\end{axis}
\end{tikzpicture}
  
        \caption{Impact of the number of \sfeatures}
        \label{fig:ablation_N}
    \end{subfigure}%
    \begin{subfigure}{.5\linewidth}
        \centering
        \begin{tikzpicture}
\begin{axis}[%
  height=4.5cm, 
  xlabel={number of iterations (T)},
  xtick = {1,2,3,4,6,9,12},
  xticklabels = {1,2,3,4,6,9,12},
  ylabel={mAP},   minor tick num=3,
  ]

\pgfplotstableread{
n val oxmed oxhard pamed pahard ox par
1 85.4 77.7 54.4 80.6 61.6 66.0 71.1
2 78.5 71.8 45.3 80.5 62.5 58.5 71.5
3 79.9 71.3 43.9 78.7 59.4 57.6 69.0
4 83.1 75.8 52.1 82.7 66.0 64.0 74.3
6 89.7 81.8 61.2 85.3	70.0 71.5 77.6
9 89.4 81.7 59.7 85.6 70.1 70.7 77.9
12 88.9 80.5 58.6 83.9 67.6 69.5 75.7
}{\map}

    \addplot[val]      table[x=n,  y=val]   \map; 
    \addplot[oxmed]      table[x=n,  y=ox]   \map; 
    \addplot[pamed]      table[x=n,  y=par]   \map;

\end{axis}
\end{tikzpicture}
        \caption{Impact of the number of iterations}
        \label{fig:ablation_T}
    \end{subfigure}
\caption{\textbf{Varying the number of \mname $N$ and of iterations $T$} in the \fimlong.}
\label{fig:ablation_NT}
\end{center}
\end{figure}

\subsection{Impact of hard negatives}
\label{app:nnegative}
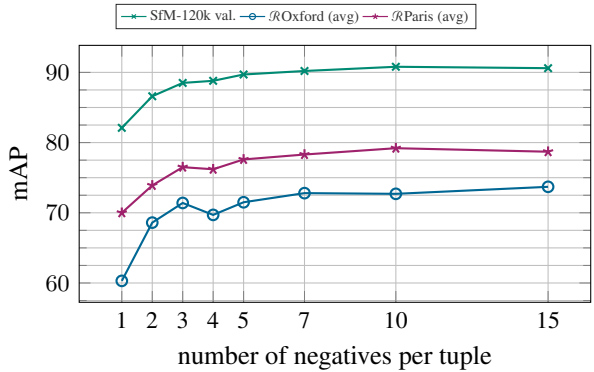
\begin{figure}[htb]
\begin{center}
        \begin{center}
\resizebox{.35\linewidth}{!}
{
    \begin{tikzpicture}
    \begin{axis}[%
        hide axis,
        xmin=10, xmax=50,
        ymin=0,ymax=0.4,
        legend columns=-1,
        legend style={draw=white!15!black,legend cell align=left}
        ]
        \addlegendimage{val} \addlegendentry{\val}
        \addlegendimage{oxmed} \addlegendentry{\ROxford (avg)}
        \addlegendimage{pamed} \addlegendentry{\RParis (avg)}
    \end{axis}
    \end{tikzpicture}
    }
    \end{center}
        \vspace{5pt}
        \begin{tikzpicture}
\begin{axis}[%
  width=.6\linewidth,
  height=5cm, 
  xlabel={number of negatives per tuple},
  xtick = {1,2,3,4,5,7,10,15},
  xticklabels = {1,2,3,4,5,7,10,15},
  ylabel={mAP},   minor tick num=3,
  ]

\pgfplotstableread{
n val oxmed oxhard pamed pahard ox par
1						82.1	73.4	47.2	80.2	59.7	60.3	70.0
2						86.6	80.0	57.3	82.6	65.1	68.6	73.9
3						88.5	81.9	60.8	84.4	68.7	71.4	76.5
4						88.8	80.7	58.7	84.6	67.9	69.7	76.2
5						89.7	81.8	61.2	85.3	70.0	71.5	77.6
7						90.2	83.1	62.6	85.9	70.6	72.8	78.3
10						90.8	82.8	62.7	86.3	72.0	72.7	79.2
15                      90.6    83.1    64.3    86.1    71.2    73.7    78.7
}{\map}

    \addplot[val, thick]      table[x=n,  y=val]   \map; 
    \addplot[oxmed, thick]      table[x=n,  y=ox]   \map; 
    \addplot[pamed, thick]      table[x=n,  y=par]   \map;

\end{axis}
\end{tikzpicture}
\end{center}
\caption{\textbf{Varying the number of hard negatives per training tuple.}}
\label{fig:ablation_nneg}
\end{figure}

Each training tuple is composed of one image pair depicting the same landmark, and $5$ negative images (\ie from different landmarks). We plot the performance when varying this number in Figure~\ref{fig:ablation_nneg}.
We observe that adding more negatives improves overall the performance on all datasets, \ie by more than $1$\% when increasing it to $10$ or $15$ negatives, but at the cost of a longer training time as more images need to be processed at each iteration. This is why we use $5$ negatives in the rest of the paper as it offers a good compromise between performance and cost.

\subsection{Impact of the update function}
\label{app:update}

The formula for $\attfunc$ in Equation~(\ref{eq:phi}) sums the previous $\gQ$ with the output of the attention component $\Vfunc(\gU) \cdot \mbm{\alpha}$, \ie, with a residual connection. This is a different choice than the one made in the object-centric slot attention of~\cite{slotattention}, which proposes to use a Gated Recurrent Unit: $\attfunc(\gU ; \gQ) = \texttt{GRU} ( \Vfunc(\gU) \cdot \mbm{\alpha}, \gQ$ ). We thus compare the residual connection we use to a $\texttt{GRU}$ function and report results in Table~\ref{tab:update} with $T=3$.
We observe that the residual connection reaches a better performance in all datasets while having the interest of not adding extra parameters.

\begin{table}[htb]
\centering
    \begin{tabular}{l  c cc cc}
\toprule
update & SfM-120k & \multicolumn{2}{c}{\ROxford} & \multicolumn{2}{c}{\RParis} \\
function &  val & med & hard & med & hard \\
\cmidrule(l){1-1} \cmidrule(l){2-2} \cmidrule(l){3-4} \cmidrule(l){5-6}
residual & \bf{79.9} & \bf{71.3} & \bf{43.9} & \bf{78.7} & \bf{59.4} \\ 
\texttt{GRU}      & 74.7 & 66.7 & 41.4 & 77.2 & 57.8 \\
\bottomrule
    \end{tabular}
\caption{\textbf{Impact of the update function} in the \lit module. We compare the performance of the residual combination of the previous template value with the cross-attention tensor compared to a GRU as used in slot attention~\citep{slotattention}. In this experiment, we use $T=3$.}
\label{tab:update}
\end{table}

In summary, we propose the \lit module to obtain a few hundred features for image retrieval while slot attention handles a handful of object attentions. Technical differences to the slot attention include: a) we use a learned initialization of the templates instead of i.i.d. sampling, which allows to obtain an ordered set of features, and thus to apply the constraint on the ID for the matching, leading to a clear gain, see Table~\ref{tab:match} and Figure~\ref{fig:trainingmatch}, b) to handle a larger number of templates, we also add a decorrelation loss on the attention maps, which has clear benefit, see Table~\ref{tab:loss_ablation} and Figure~\ref{fig:correlation}, c) we use a residual connection with 6 iterations instead of a \texttt{GRU}, leading to improved performance for our task (see Table~\ref{tab:update}).

\subsection{Extended ablation on matching constraints}
\label{app:matching_extended}

\begin{table}[h]
\centering
\begin{tabular}{lcc c cc cc}
\toprule
reci- & ratio & same & SfM-120k & \multicolumn{2}{c}{\ROxford} & \multicolumn{2}{c}{\RParis} \\
 proc.    & test  & ID  & val & med & hard & med & hard \\

\cmidrule(l){1-3} \cmidrule(l){4-4} \cmidrule(l){5-6} \cmidrule(l){7-8}

 &             &              & 68.3 & 64.3 & 39.8 & 74.1 & 52.4 \\
$\checkmark$    &     &              & 79.6 & 69.2 & 44.3 & 79.2 & 60.9 \\
$\checkmark$    &     &    $\checkmark$           & \bf{89.9} & \bf{81.9} & 61.1 & 85.1 & 69.6 \\
    & $\checkmark$     &              & 73.5 & 64.6 & 39.0 & 75.5 & 55.0 \\
        & $\checkmark$     &  $\checkmark$  & 84.2 & 75.0 & 49.8 & 79.4 & 61.1 \\
$\checkmark$ &   $\checkmark$ &           & 80.8 & 70.7 & 45.1 & 80.3 & 61.9 \\
            & &  $\checkmark$ & 75.9 & 63.8 & 35.1 & 77.3 & 56.5 \\
$\checkmark$ &   $\checkmark$  & $\checkmark$ & 89.7 & 81.8 & \bf{61.2}	& \bf{85.3} & \bf{70.0} \\
\bottomrule
\end{tabular}
\caption{\textbf{Extended ablation on matching constraints.} We study the impact of removing constraints on reciprocity,  Lowe’s ratio test and the Super-feature ID.}%
 \label{tab:matching_extended}
\end{table}

We show in Table~\ref{tab:matching_extended} an extended version of Table~\ref{tab:match} of the main paper, where we evaluate all possible combinations of constraints among reciprocity, Lowe's ratio test and the Super-feature ID. We observe that the reciprocity constraint and the Super-feature ID constraint are the two most important ones, while the Lowe's ratio only only brings a small improvement.

\subsection{Impact of the template initialization}
\label{app:template_init}

\begin{table}[h]
\centering
\begin{tabular}{l c cc cc}
\toprule
Template & SfM-120k & \multicolumn{2}{c}{\ROxford} & \multicolumn{2}{c}{\RParis} \\

Initialization  & val & med & hard & med & hard \\

\cmidrule(l){1-1} \cmidrule(l){2-2} \cmidrule(l){3-4} \cmidrule(l){5-6}

Frozen from ImageNet pretraining & 89.5 & 81.3 & 59.6 &	\bf{85.3} & \bf{70.2}  \\
Fine-tuned for landmark retrieval & \bf{89.7} & \bf{81.8} & \bf{61.2}	& \bf{85.3} & 70.0 \\
\bottomrule
\end{tabular}

\caption{\textbf{Fine-tuning the initial templates.}  Comparison where we either fine-tune the initial templates $\gQ^0$ of LIT (bottom row, as in the main paper) or keep them frozen after ImageNet pretraining (top row).}
\label{tab:initialization}
\end{table}

In the \lit module, initial templates $\gQ^0 \in \mathbb{R}^{\nsfeat \times \attdim}$ are learned together with the LIT module. We can therefore assume that they are adapted to the task at hand.
To explore the sensitivity to the initial templates, we run a variant of FIRe where the initializations are not fine-tuned, but instead frozen to the values after pretraining on ImageNet. We report performance in Table~\ref{tab:initialization}. We observe that the two variants perform overall similarly. This ablation suggests that the initial templates are up to some point transferable to other tasks.

\section{Analysis and discussions}
\label{app:analysis}

\subsection{Single-scale results}
\label{app:ablation_nfeat_onescale}

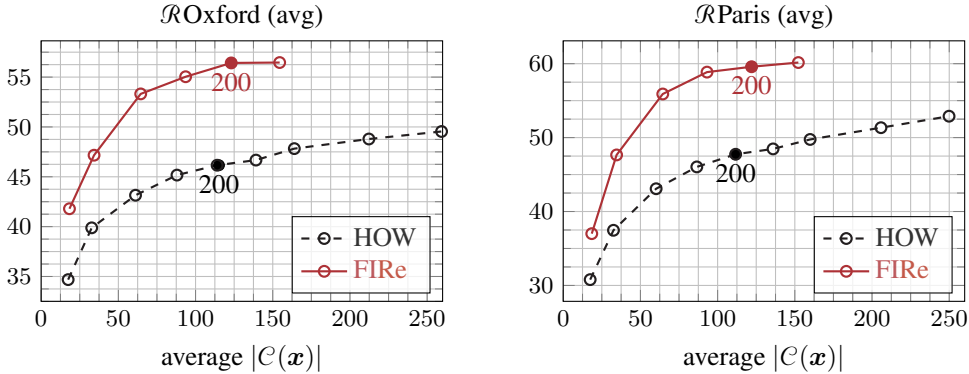
\begin{figure}
\begin{center}
    \vspace{2pt}
    \resizebox{\linewidth}{!}{
\begin{subfigure}{.49\linewidth}
        \centering
        \begin{tikzpicture}
\begin{axis}[%
  height=5cm, 
xlabel={average $|\mathcal{C}(\vx)|$},
    xmin=0, xmax=260,
    minor tick num=3,
   legend pos=south east,
    title={\ROxford (avg)},
            title style={yshift=-5pt},
  ]

\pgfplotstableread{
nfeats xbaseval xbaseox xbasepar ybaseval ybaseoxavg ybaseparavg  xoursval xoursox xourspar yoursval  yoursoxavg yoursparavg  
25 17.69744992	17.4820749	17.45697564 30.18	34.685	30.785 18.52167967	18.33246545	18.49082569 44.31	41.795	37.005
50 32.99058805	32.74324054	32.54792787 39.28	39.88	37.46 34.29933231	34.27117965	34.58130339 53.96	47.165	47.655
100 61.81055426	61.12036852	60.17257197 48.34	43.15	43.08 63.81160003	64.48027238	64.47342613 63.36	53.315	55.9
150 89.14584506	87.9601442	86.52657387 53.29	45.17	46.035 91.74298126	93.57280192	93.13982917 66.56	55.035	58.86
200 115.1859866	113.9631484	111.6732047 57.2	46.16	47.745 119.59963	123.1025436	122.0022145 70.86	56.42	59.585
300 140.3409219	139.1265772	135.9120531 58.7	46.675	48.47 149.3854074	154.474264	152.2896235 72.47	56.46	60.155
400 165.2404473	163.9697577	159.9188548 60.99	47.825	49.75 	nan nan nan nan nan nan
500 212.4074491	212.2603645	205.8035432 63.54	48.79	51.34 	nan nan nan nan nan nan
600 258.205615	259.3302624	249.8850047 63.61	49.545	52.895 	nan nan nan nan nan nan
}{\map}

    \addplot[howoxf]      table[x=xbaseox,  y=ybaseoxavg]   \map; 
    \leg{HOW} 
    \addplot[oursoxf]      table[x=xoursox,  y=yoursoxavg]   \map; 
    \leg{\fire}

    \draw[fill] (axis cs: 115.1859866,46.16) circle (2pt) node [below] {200};
    \draw[fill,\ourscolor] (axis cs: 123.1025436,56.42) circle (2pt)  node [below] {200};

\end{axis}
\end{tikzpicture}
  
        \label{fig:ablation_nfeat_one_scale_par}
    \end{subfigure}%
    \begin{subfigure}{.49\linewidth}
        \centering
        \begin{tikzpicture}
\begin{axis}[%
  height=5cm, 
xlabel={average $|\mathcal{C}(\vx)|$},
    xmin=0, xmax=260,
    minor tick num=3,
    legend pos=south east,
        title={\RParis (avg)},
                title style={yshift=-5pt},
  ]

\pgfplotstableread{
nfeats xbaseval xbaseox xbasepar ybaseval ybaseoxavg ybaseparavg  xoursval xoursox xourspar yoursval  yoursoxavg yoursparavg  
25 17.69744992	17.4820749	17.45697564 30.18	34.685	30.785 18.52167967	18.33246545	18.49082569 44.31	41.795	37.005
50 32.99058805	32.74324054	32.54792787 39.28	39.88	37.46 34.29933231	34.27117965	34.58130339 53.96	47.165	47.655
100 61.81055426	61.12036852	60.17257197 48.34	43.15	43.08 63.81160003	64.48027238	64.47342613 63.36	53.315	55.9
150 89.14584506	87.9601442	86.52657387 53.29	45.17	46.035 91.74298126	93.57280192	93.13982917 66.56	55.035	58.86
200 115.1859866	113.9631484	111.6732047 57.2	46.16	47.745 119.59963	123.1025436	122.0022145 70.86	56.42	59.585
300 140.3409219	139.1265772	135.9120531 58.7	46.675	48.47 149.3854074	154.474264	152.2896235 72.47	56.46	60.155
400 165.2404473	163.9697577	159.9188548 60.99	47.825	49.75 	nan nan nan nan nan nan
500 212.4074491	212.2603645	205.8035432 63.54	48.79	51.34 	nan nan nan nan nan nan
600 258.205615	259.3302624	249.8850047 63.61	49.545	52.895 	nan nan nan nan nan nan
}{\map}
    \addplot[howoxf]      table[x=xbasepar,  y=ybaseparavg]   \map; 
    \leg{HOW}
    \addplot[oursoxf]      table[x=xourspar,  y=yoursparavg]   \map; 
    \leg{\fire}

    \draw[fill] (axis cs: 111.6732047,47.745) circle (2pt) node [below] {200};
    \draw[fill,\ourscolor] (axis cs: 122.0022145,59.585) circle (2pt)  node [below] {200};

\end{axis}
\end{tikzpicture}
  
        \label{fig:ablation_nfeat_one_scale_ox}
    \end{subfigure}%
    }
\end{center}
\vspace{-0.3cm}
\caption{\textbf{Performance versus memory when varying the number of selected features at a single scale} for HOW and \fire. The x-axis represents the average number of vectors per image in ASMK, which is proportional to the memory, when varying the number of selected features in $(25,50,100,150,200,300,400,500,600)$, \fire is limited to 256 features.}
\label{fig:ablation_nfeat_onescale}
\end{figure}

Similar to Figure~\ref{fig:ablation_nfeat} of the main paper, we perform the same ablation in Figure~\ref{fig:ablation_nfeat_onescale} when extracting features at a single scale ($1.0$). We observe that \fire significantly improves the mAP on the two datasets compared to HOW. The average number of clusters, \ie the memory footprint of images, remains similar for both methods at a same number of selected features. This stands in contrast to the multi-scale case where our approach allows to save memory (about 20\%), which we hypothesize is due to the correlation of our features across scales that we discuss below.

\subsection{Consistency of attention maps across scales} 
\label{app:consistency}

\begin{figure}[htb]
    \centering
    \resizebox{\linewidth}{!}{
    \begin{tabular}{c@{ }c@{ }c@{ }c@{ }c@{ }c@{ }c@{ }c}
    image & $0.25$ & $0.353$ & $0.5$ & $0.707$ & $1.0$ & $1.414$ & $2.0$ \\
        \includegraphics[width=0.115\linewidth]{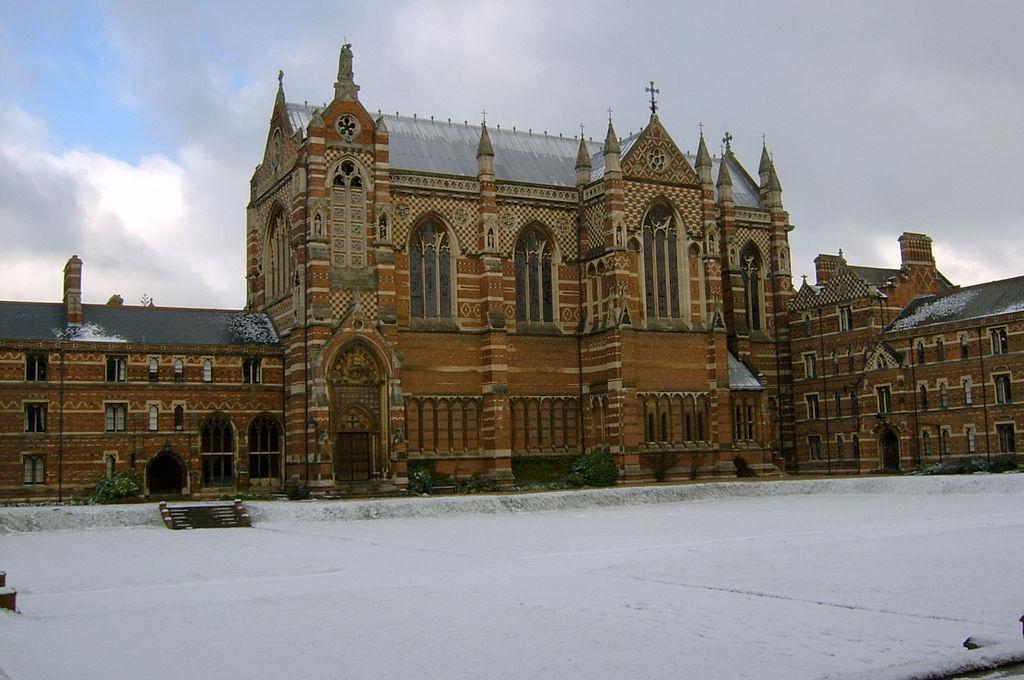} &
    \includegraphics[width=0.115\linewidth]{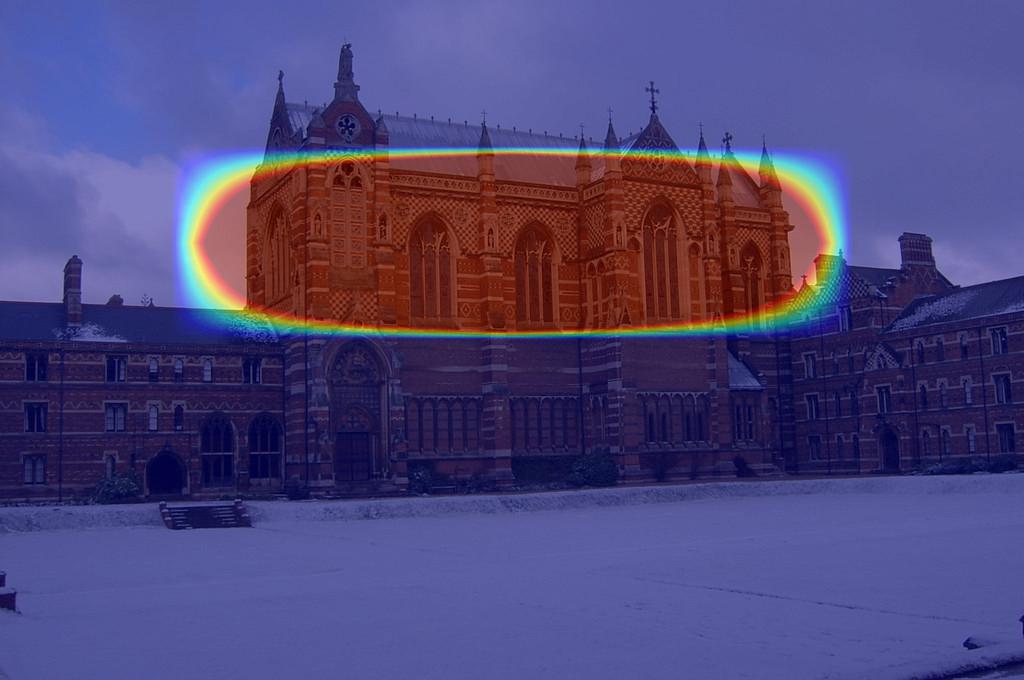} &
    \includegraphics[width=0.115\linewidth]{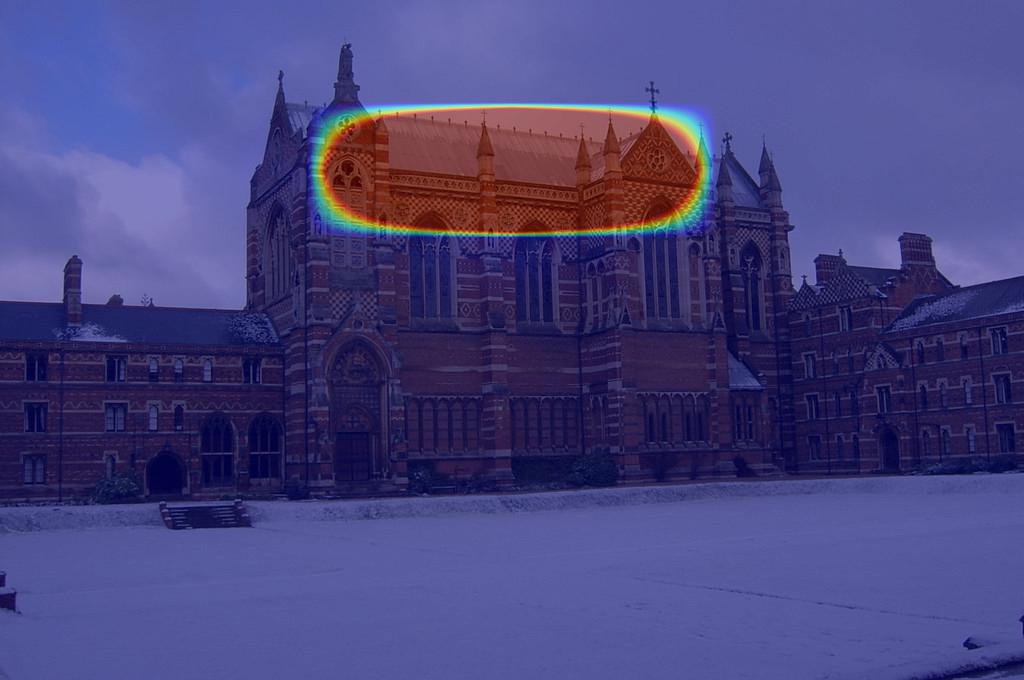} &
    \includegraphics[width=0.115\linewidth]{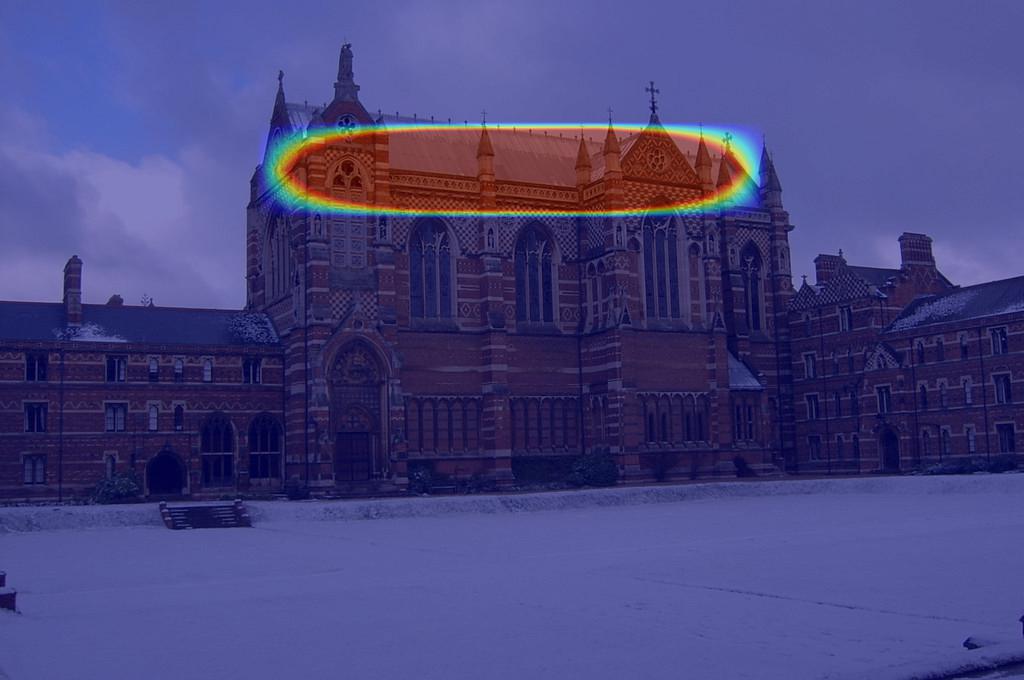} &
    \includegraphics[width=0.115\linewidth]{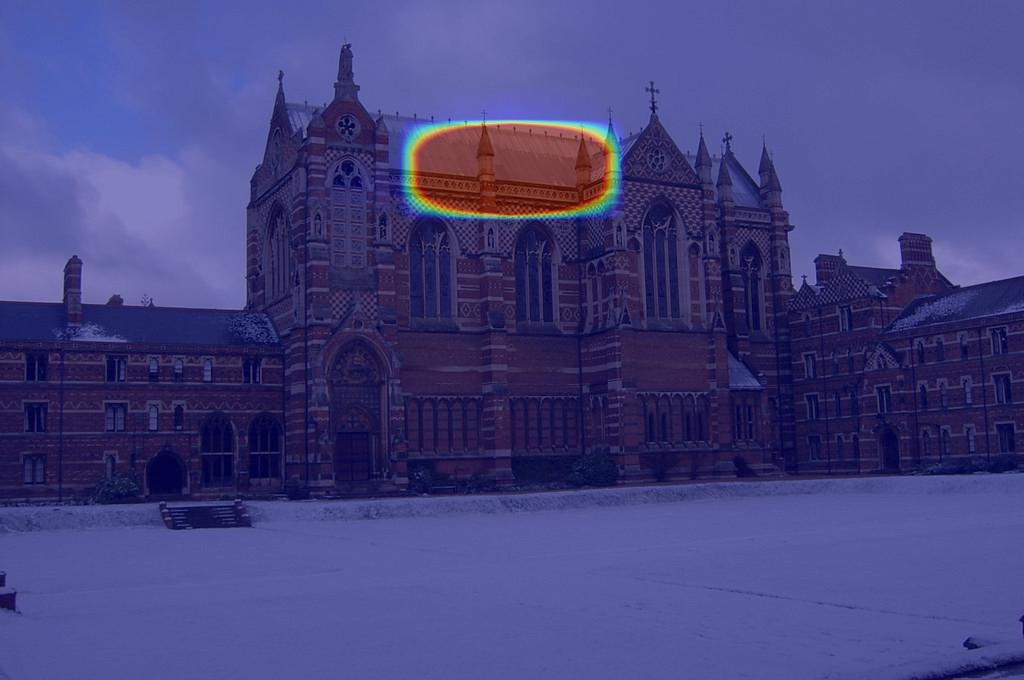} &
    \includegraphics[width=0.115\linewidth]{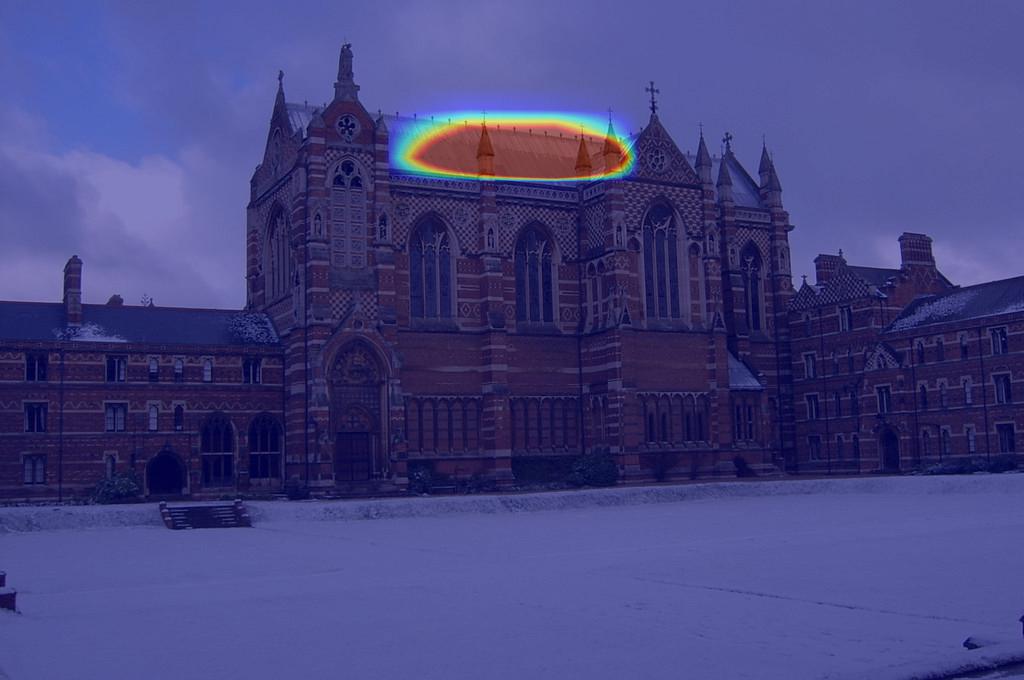} &
    \includegraphics[width=0.115\linewidth]{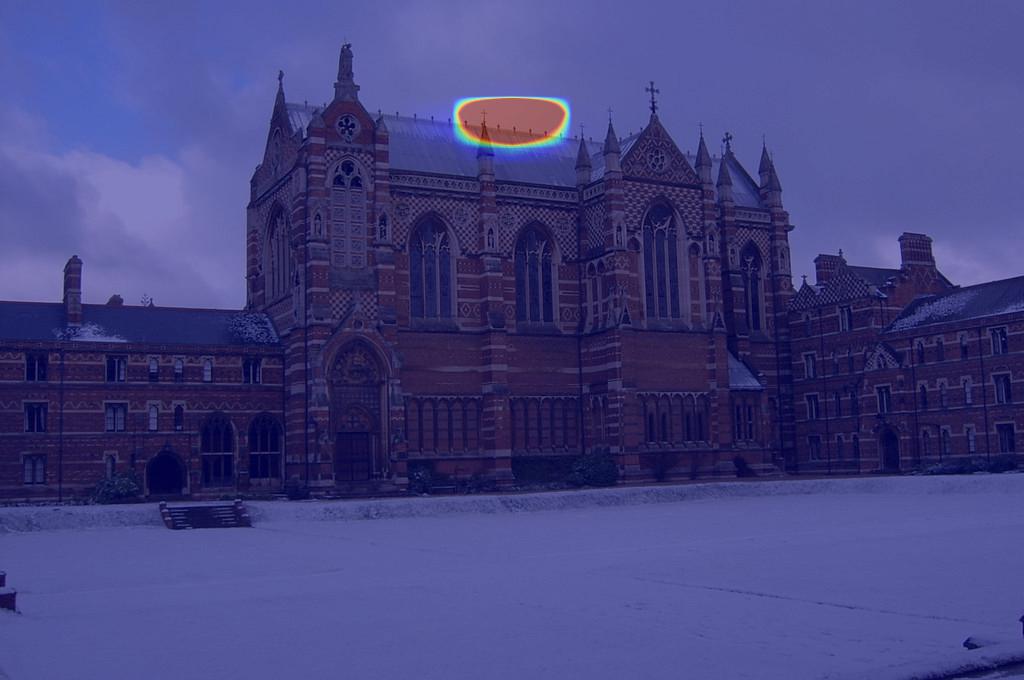} &
    \includegraphics[width=0.115\linewidth]{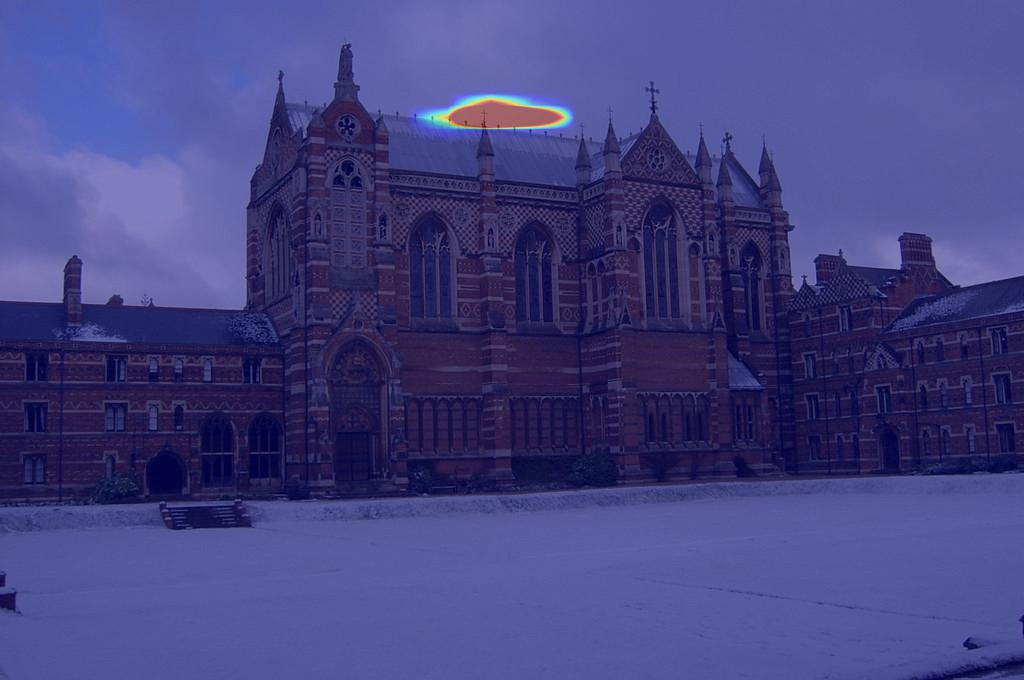} \\
    
    \includegraphics[width=0.115\linewidth]{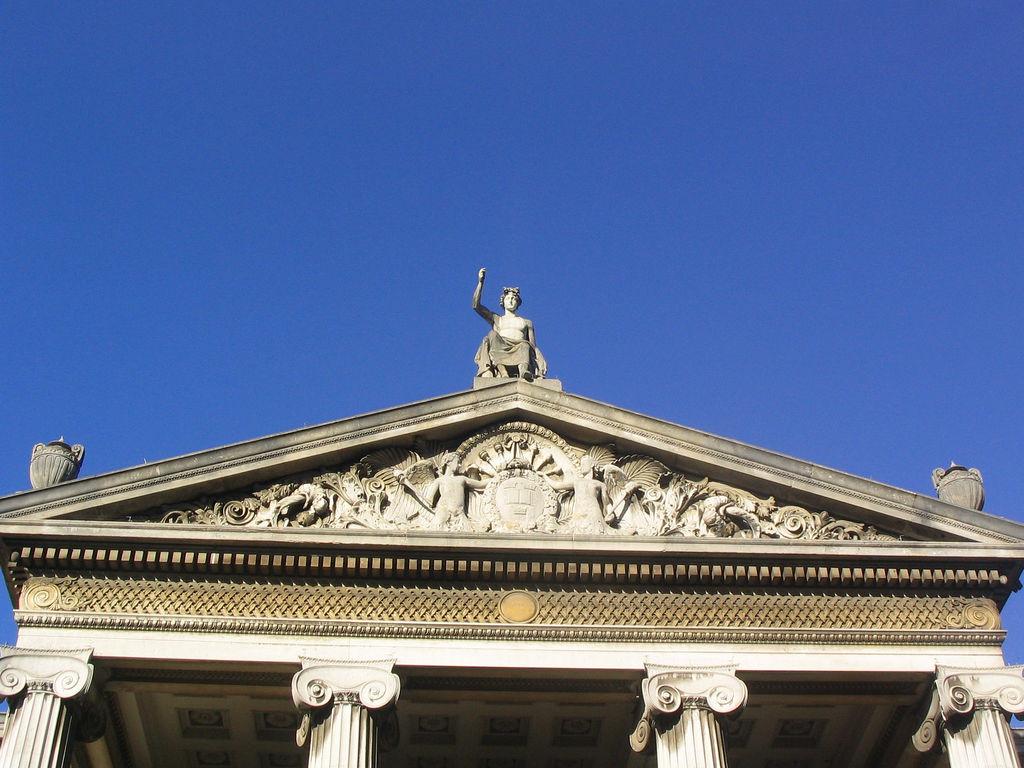} &
    \includegraphics[width=0.115\linewidth]{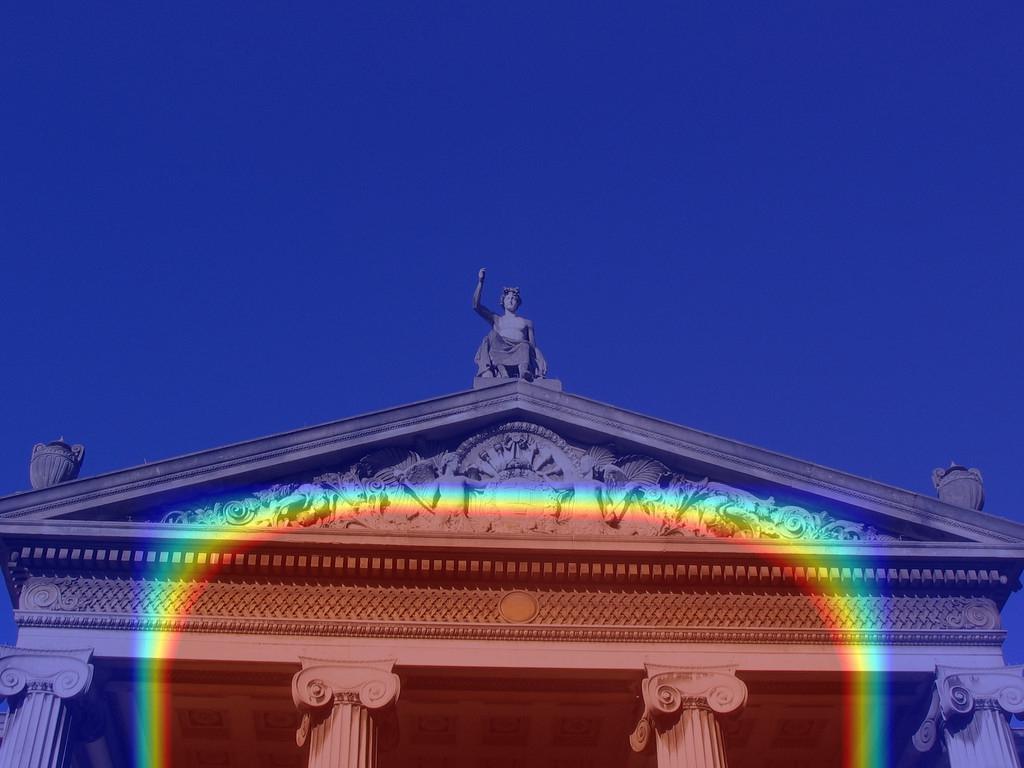} &
    \includegraphics[width=0.115\linewidth]{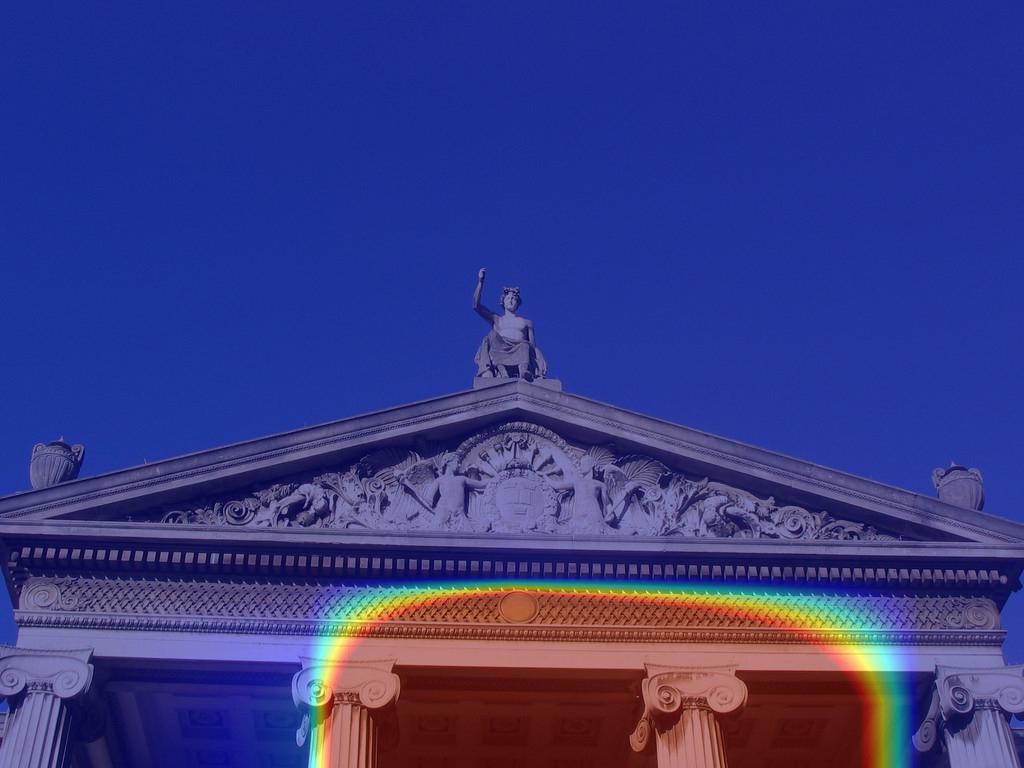} &
    \includegraphics[width=0.115\linewidth]{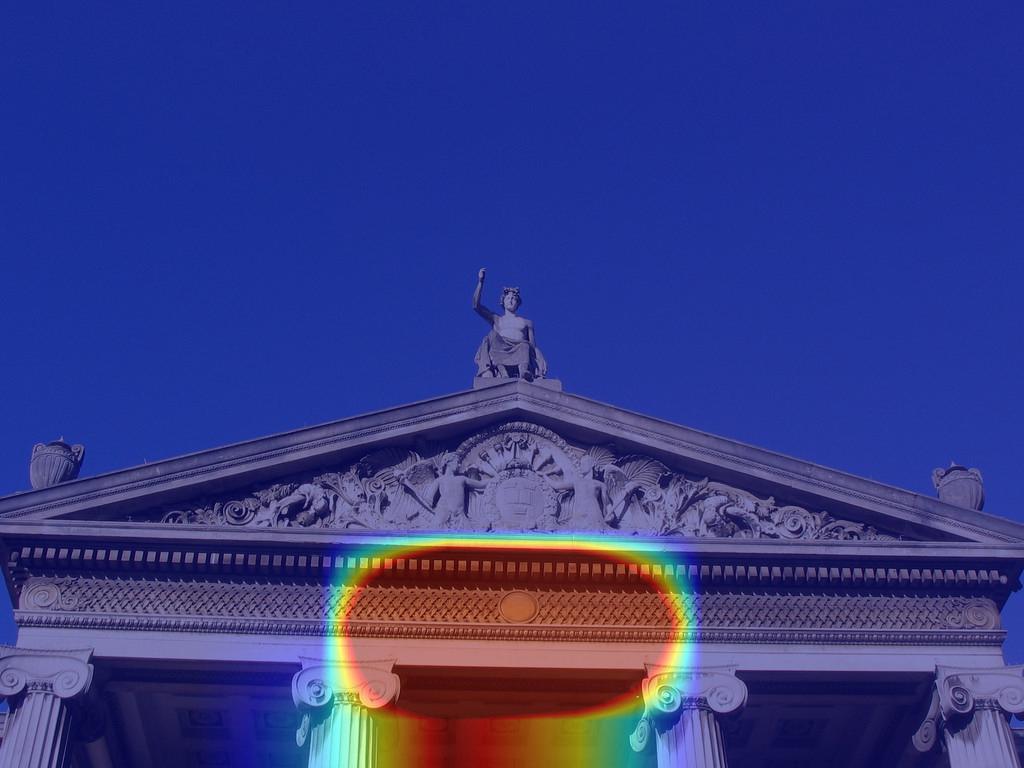} &
    \includegraphics[width=0.115\linewidth]{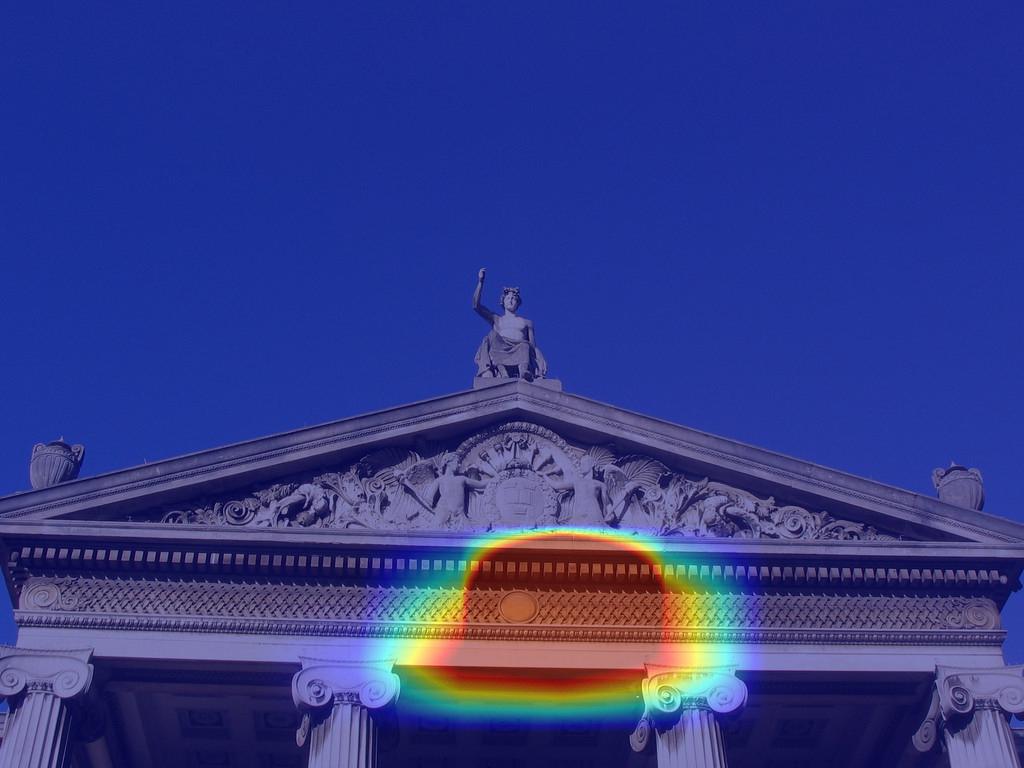} &
    \includegraphics[width=0.115\linewidth]{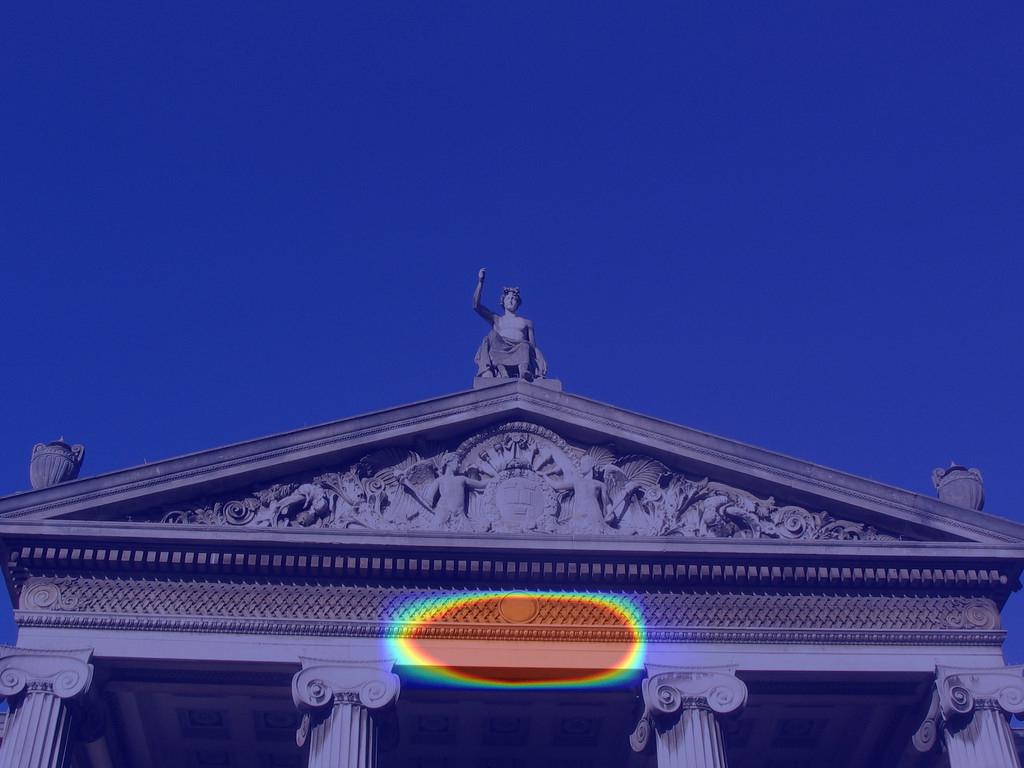} &
    \includegraphics[width=0.115\linewidth]{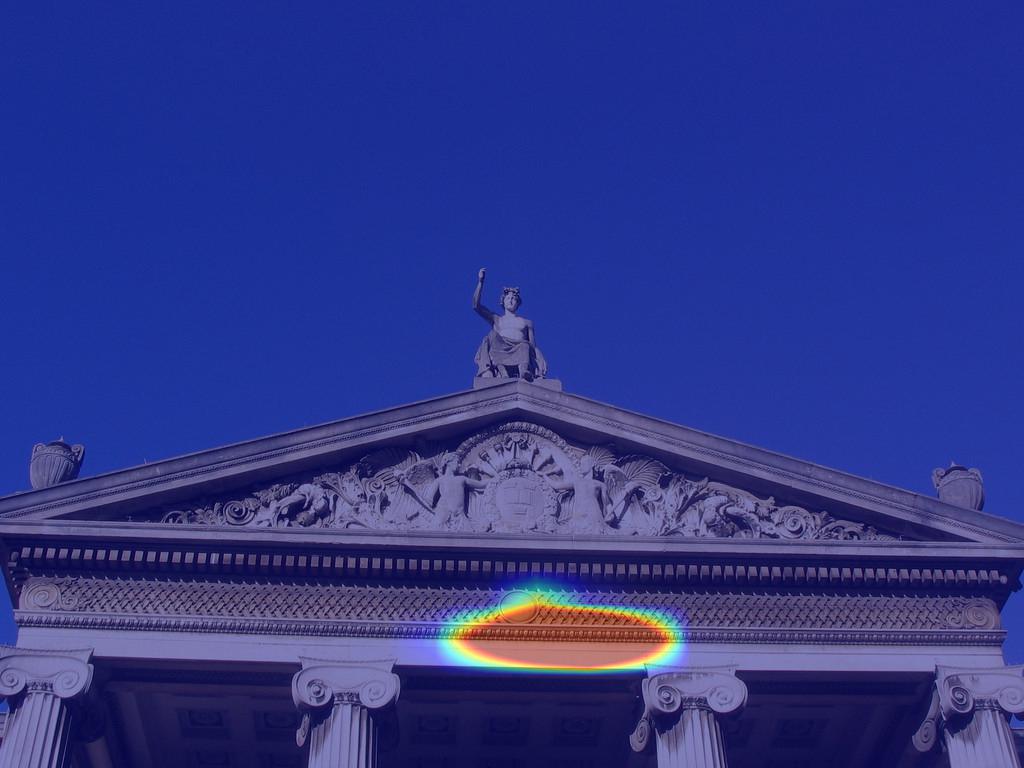} &
    \includegraphics[width=0.115\linewidth]{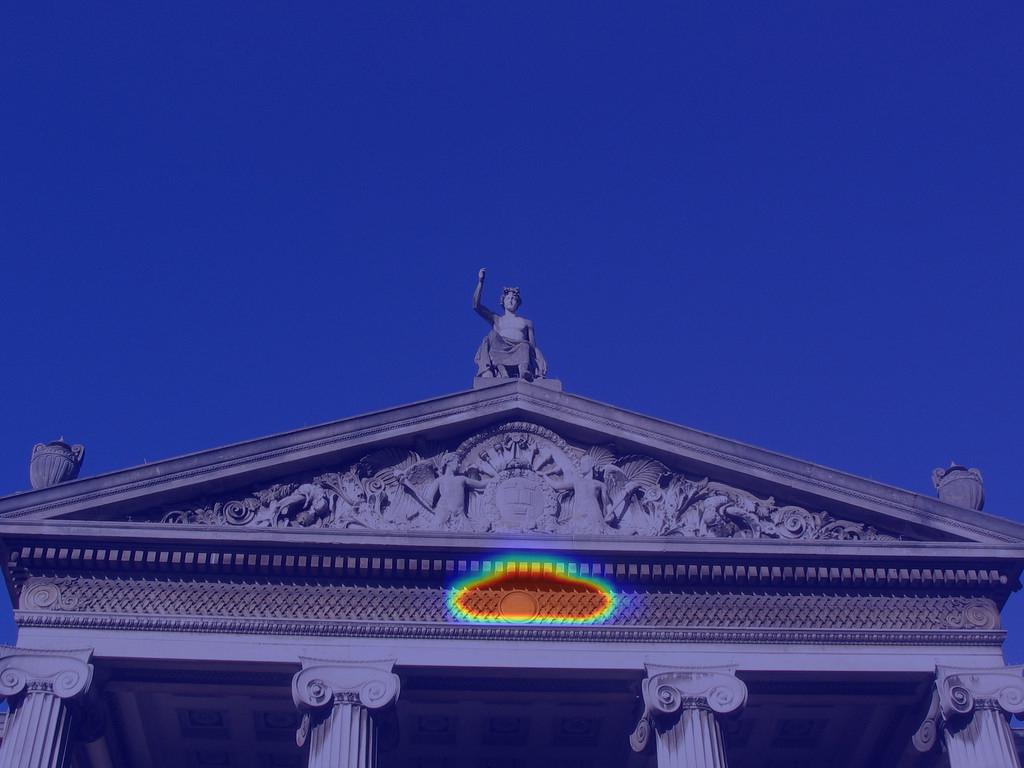} \\
    
    \includegraphics[width=0.115\linewidth]{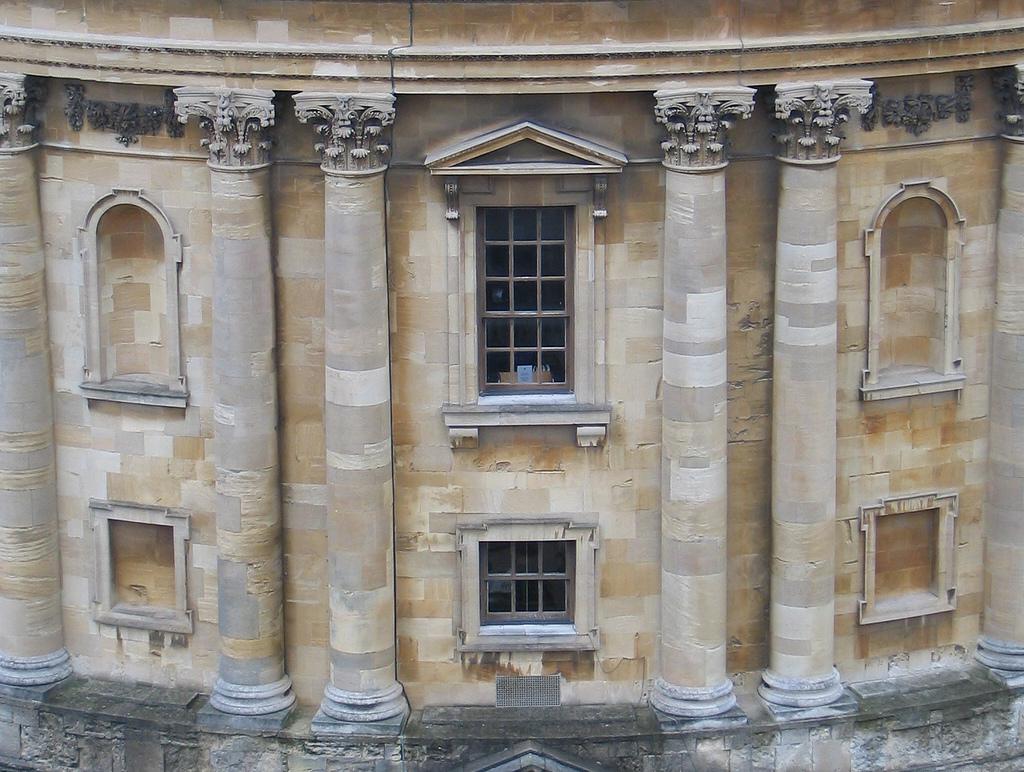} &
    \includegraphics[width=0.115\linewidth]{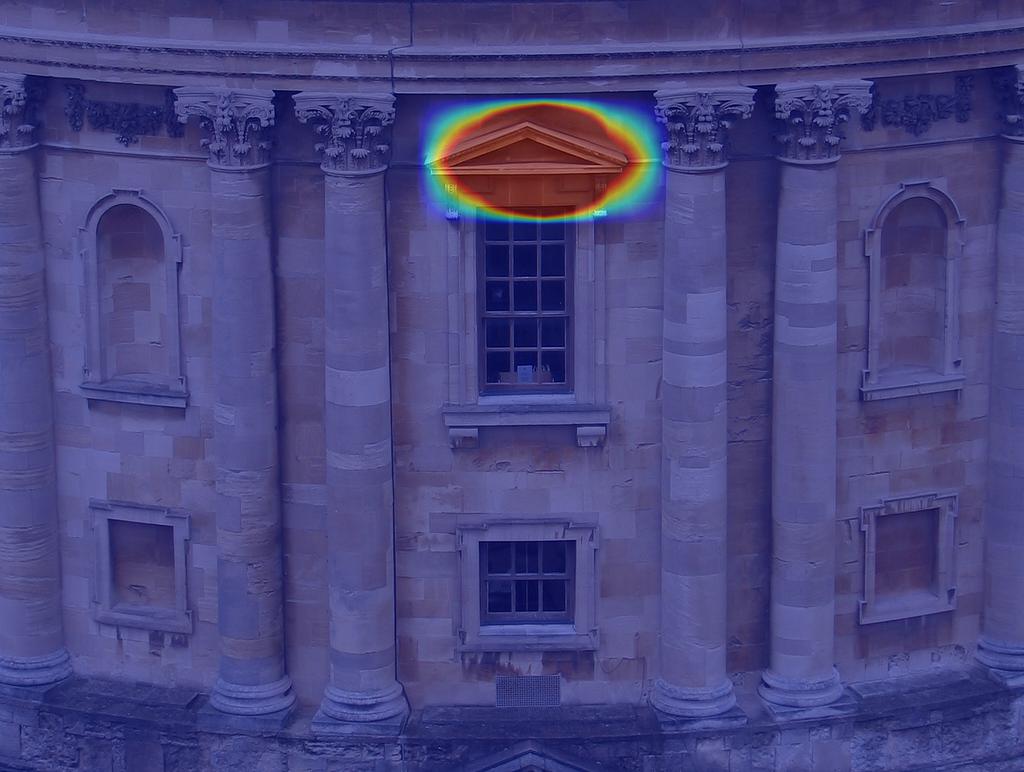} &
    \includegraphics[width=0.115\linewidth]{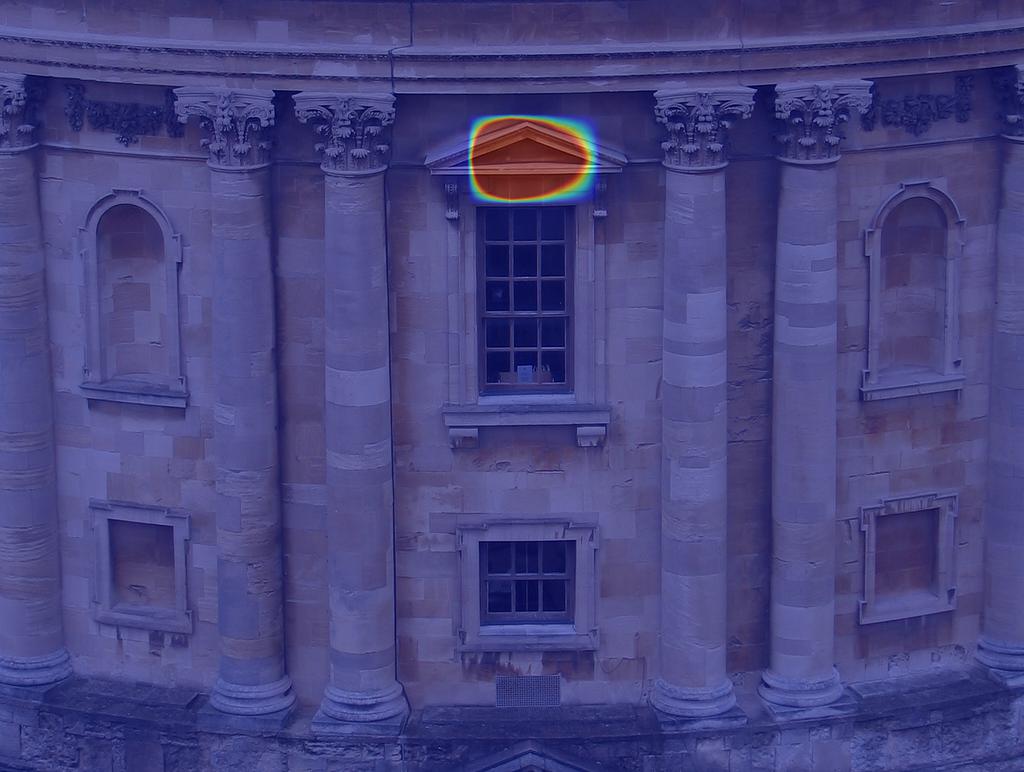} &
    \includegraphics[width=0.115\linewidth]{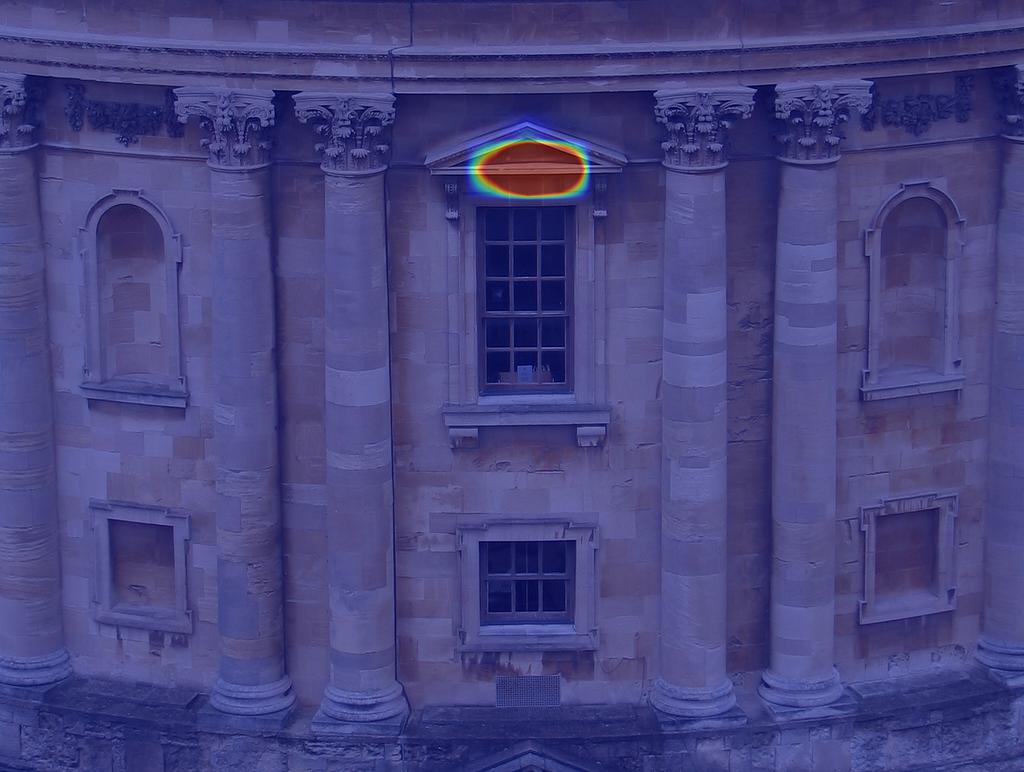} &
    \includegraphics[width=0.115\linewidth]{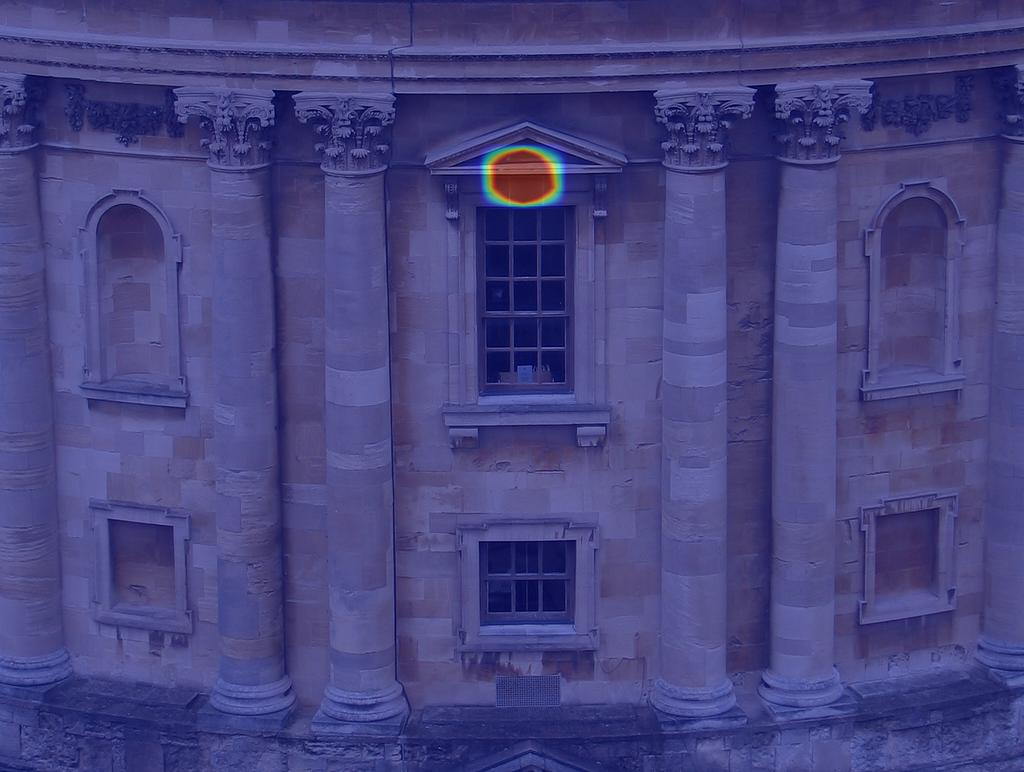} &
    \includegraphics[width=0.115\linewidth]{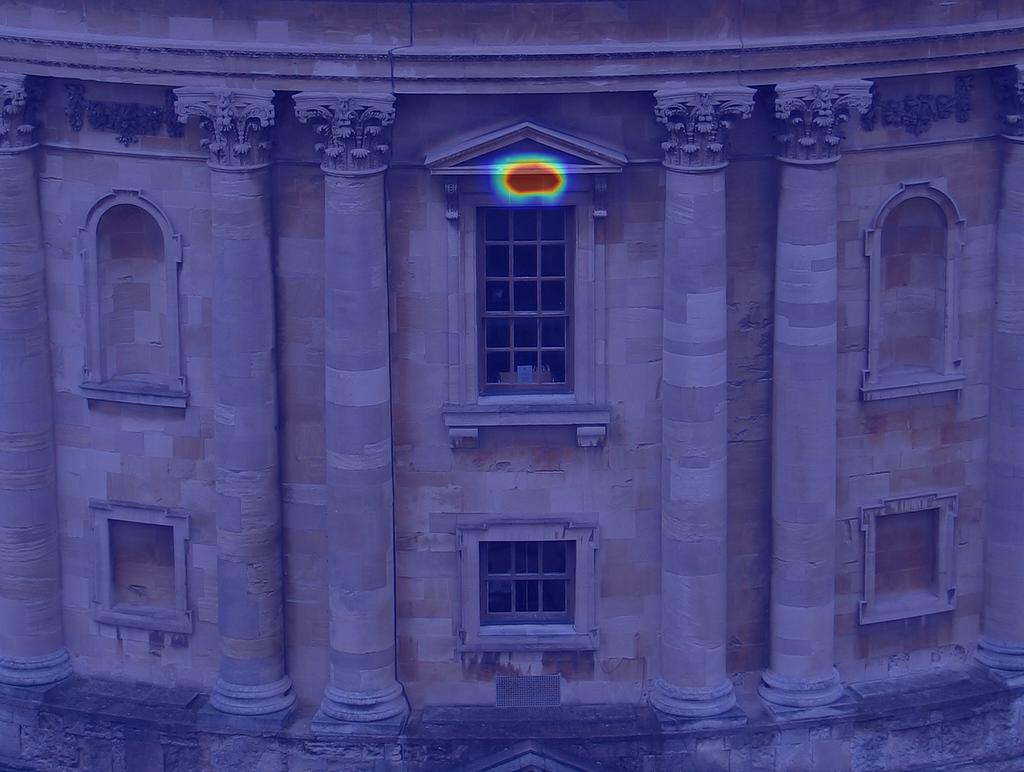} &
    \includegraphics[width=0.115\linewidth]{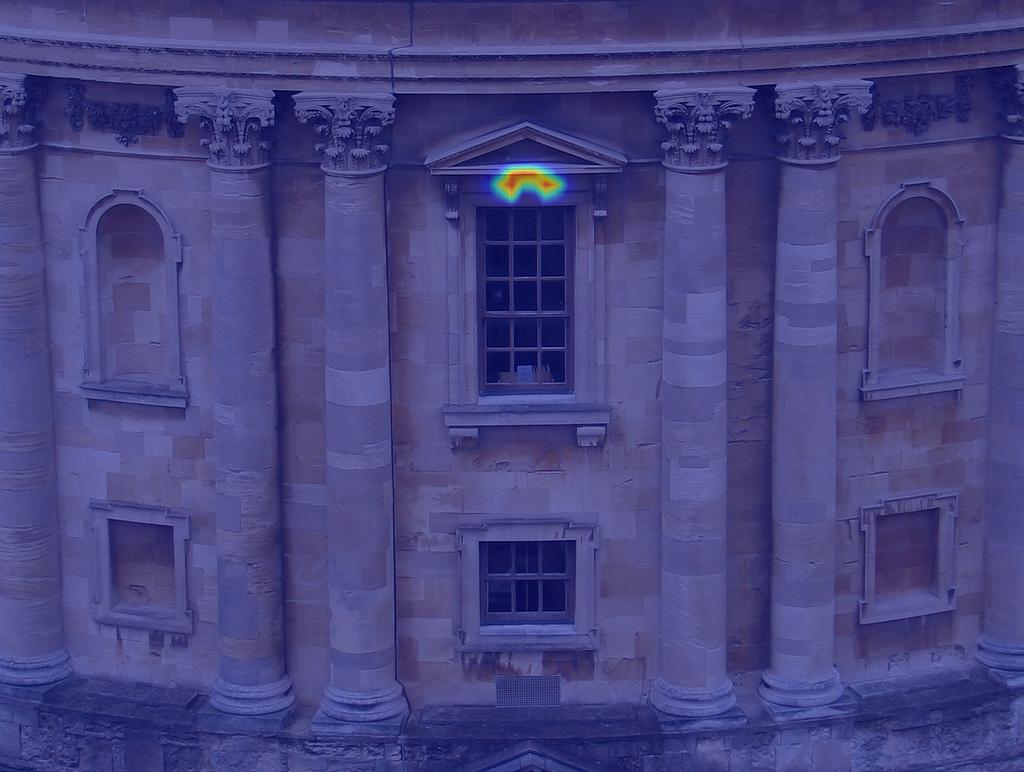} &
    \includegraphics[width=0.115\linewidth]{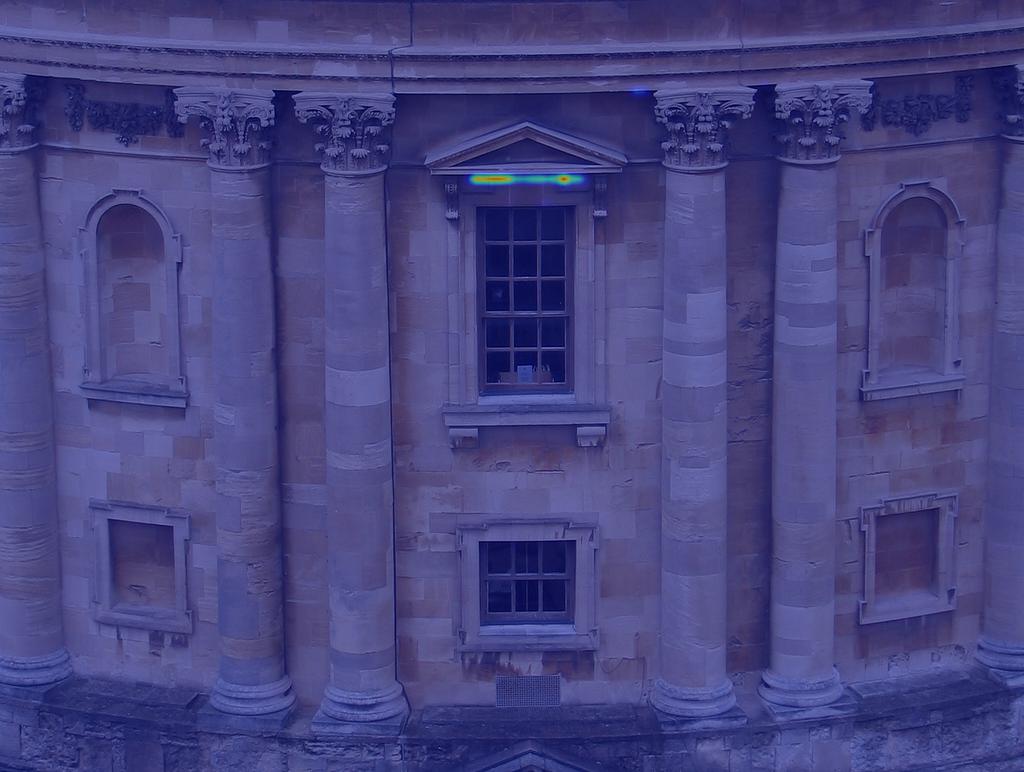} \\
    \end{tabular}
    }
    
    \caption{\textbf{Attention consistency across scales.} For three images, we select one \sfeature per image and show the attention maps of the latest iteration of \lit for different image scales, with from left to right $0.25, 0.353, 0.5, 0.707, 1.0, 1.414$ and $2.0$. We clearly observe that attention maps are correlated. They show larger regions at small scales as for visualization, we resize the lower resolution attention to the original image size.}
    \label{fig:attn_scale}
\end{figure}

At test time, we extract \mname at different image resolutions. We propose here to study if the attention maps of the \mname across different image scales are correlated. We show in Figure~\ref{fig:attn_scale} the attention maps at the last iteration of \fim for the different image scales. We observe that they fire at the same image location at the different scales. Note that the attention maps are larger/smoother at small scales (right columns), as for visualization, we resize lower resolution attention maps to the original image size.

\subsection{Redundancy in \sfeatures}
\label{app:redundancy}

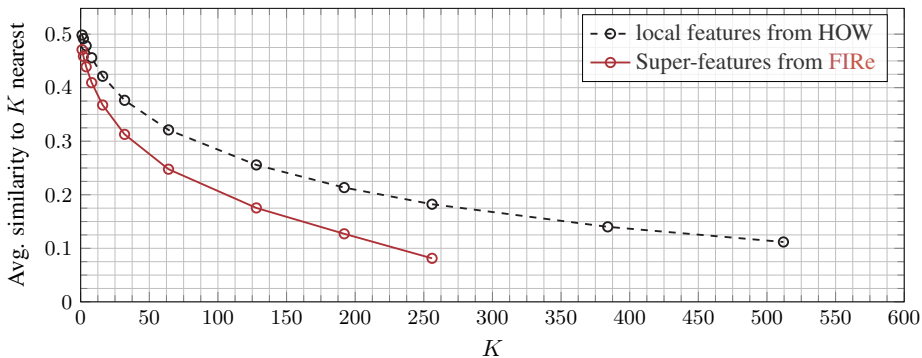
\begin{figure}[ht]
\begin{center}
\resizebox{0.9\linewidth}{!}{
        \begin{tikzpicture}
\begin{axis}[%
  height=6cm, 
  xlabel={$K$},
  ylabel={Avg. similarity to $K$ nearest}, 
  minor tick num=3,
  ymin=0,
 xmin=0, xmax=600,
  ]

\pgfplotstableread{
how ours xxx
0.4983 0.4710 1
0.4916 0.4586 2
0.4782 0.4390 4
0.4559 0.4094 8
0.4213 0.3676 16
0.3764 0.3128 32
0.3211 0.2477 64
0.2557 0.1754 128
0.2134 0.1271 192
0.1823 0.0813 256
0.1401 nan    384 
0.1118 nan    512
}{\map}

    \addplot[howoxf, dashed]      table[x=xxx,  y=how]   \map; 
        \addlegendentry{local features from HOW}
    \addplot[oursoxf]      table[x=xxx,  y=ours]   \map; 
        \addlegendentry{\sfeatures from \fire}
\end{axis}
\end{tikzpicture}
  
}
\end{center}
\vspace{-0.3cm}
\caption{\textbf{Measuring \sfeature redundancy.}
We compute the average cosine similarity between every feature (local feature from HOW or \sfeature from \fire) and its K most similar features from the same image, for varying K. 
Results are averaged over the 70 query images of the \ROxford dataset. Only 256 \mname can be extracted per image, explaining its maximum x value.}
\label{fig:kmean_similarity}
\end{figure}

To evaluate the redundancy of \sfeatures versus local features,  Figure~\ref{fig:kmean_similarity} displays the average cosine similarity between every local feature / \sfeature and its $K$ nearest local features / \sfeatures from the same image, for different values of $K$. We observe that \sfeatures are significantly less correlated to the most similar other ones, compared to local features used in HOW.

\subsection{Are all \mname trained?}
\label{app:alltrained}

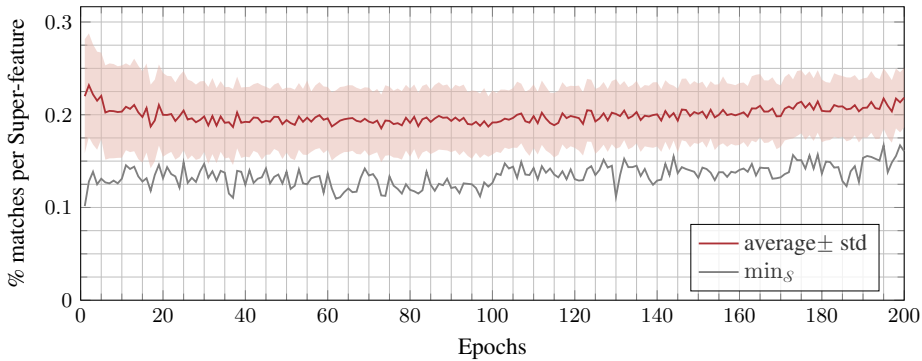
\begin{figure}[ht]
\begin{center}
\vspace{5pt}
\resizebox{0.9\linewidth}{!}{
        \begin{tikzpicture}
\begin{axis}[%
  height=6cm, 
  xlabel={Epochs},
  ylabel={\% matches per \sfeature}, 
  minor tick num=3,
  ymin=0, 
  xmin=0, xmax=200,
  legend pos=south east,
  ]

\pgfplotstableread{
xxx y_mean y_std y_min 
1 440.0234375 123.02667170448689 203
2 463.62890625 111.75796507711972 256
3 443.76171875 107.93213407412965 277
4 430.3203125 105.74811121198498 250
5 440.6953125 93.3109769294446 262
6 404.828125 101.64430239312175 254
7 408.296875 102.3919292607302 252
8 407.73828125 101.63306135085128 258
9 405.82421875 98.82281907384625 252
10 406.71484375 99.48225704698847 262
11 417.12890625 93.68294989980146 291
12 412.61328125 93.24972174145293 283
13 421.11328125 99.80249470132698 288
14 405.0625 100.85542328625665 267
15 395.140625 98.21804779219232 256
16 414.39453125 96.09441874449719 263
17 374.69140625 90.19403175181527 236
18 387.5625 87.99842416912931 270
19 421.015625 87.47334275857631 292
20 399.75 90.40585572848697 263
21 399.578125 84.30600043285398 287
22 401.5 81.61643331768425 272
23 386.546875 83.22159419426171 267
24 396.83984375 87.80697121371165 243
25 409.07421875 85.35444458449219 297
26 389.68359375 86.26003952198818 253
27 392.2578125 80.01491522031904 264
28 383.84765625 81.15622132733779 255
29 388.75390625 83.06020726928276 278
30 396.328125 76.46114304981567 294
31 376.9453125 82.97465808472695 264
32 397.15234375 82.09020045128308 278
33 376.546875 80.85957729443294 258
34 396.21875 76.13920613545626 279
35 382.6015625 76.91552182465249 275
36 383.87109375 76.13271579569923 230
37 373.51171875 81.95878262987377 221
38 404.87109375 77.28843972049579 279
39 382.73046875 75.95954932827915 277
40 384.71875 74.9116080686932 249
41 384.02734375 76.02540715819505 280
42 394.03125 71.90493262591586 253
43 393.25 75.3759846536017 262
44 383.24609375 80.34611397955854 265
45 386.8515625 73.33794739975066 284
46 385.05078125 70.85053424297271 252
47 395.60546875 69.45550375127067 266
48 395.2734375 73.76898218040964 271
49 379.63671875 75.14982989324326 252
50 388.453125 69.84876065639514 277
51 381.984375 76.01783141710486 240
52 378.07421875 75.97593137189665 247
53 396.3203125 71.90895215411183 282
54 384.80078125 70.82667112846437 252
55 393.49609375 80.26732669487137 250
56 385.09375 65.12121360154079 263
57 394.37109375 68.89370623053104 266
58 399.10546875 71.15443570215966 268
59 389.14453125 71.89460744358908 232
60 392.1875 67.65482821831714 273
61 374.76171875 73.52743070307774 241
62 380.8125 71.77675350522618 219
63 387.33984375 72.34124411409847 222
64 390.21484375 70.79672660803666 233
65 391.78515625 71.97767621570654 241
66 392.84765625 70.5562458088002 233
67 386.81640625 71.38031862554945 234
68 384.76953125 70.3107889331735 264
69 391.28515625 67.8516356999821 284
70 385.03515625 73.31265141698319 269
71 382.3046875 66.75954503310626 272
72 387.3046875 68.27495635683222 259
73 370.6171875 72.8312069417351 226
74 387.51953125 75.97784171079272 225
75 386.6484375 73.87738890424183 267
76 379.828125 72.15089541360089 247
77 381.7734375 74.75603182642584 240
78 377.12109375 73.90310500955769 230
79 393.45703125 70.89482547186729 250
80 382.6328125 73.2747794151565 238
81 395.53515625 69.63985713324006 235
82 375.48046875 72.05797097497454 230
83 390.25 71.93747284969079 237
84 394.11328125 71.71432352472132 277
85 384.48828125 69.29674541542985 249
86 391.9453125 72.40584495934941 236
87 397.4375 72.6093195206373 251
88 393.44921875 69.52897720565612 248
89 394.1328125 71.15853461349414 262
90 386.6953125 67.47031275329427 242
91 393.8515625 70.67128901688857 244
92 386.7734375 74.0869648246815 231
93 393.23828125 71.76362198771672 252
94 380.0625 68.86192593697913 251
95 378.75 73.57537589098679 250
96 384.63671875 66.6398674442214 242
97 376.125 69.4685180495453 224
98 386.0078125 71.42401951700033 254
99 374.46484375 65.12325104398649 245
100 383.2890625 69.06435544382568 253
101 383.4453125 63.24800004171945 273
102 385.8828125 68.36130131214475 270
103 391.546875 62.57087973038556 295
104 389.30859375 63.23371273416943 284
105 393.1015625 64.08006952289139 288
106 399.921875 71.36472226516666 264
107 402.7109375 64.24369525775346 298
108 385.89453125 63.591842795226285 271
109 382.92578125 71.540241894525 270
110 396.92578125 68.5921935997614 285
111 381.83203125 67.60731565628843 261
112 404.32421875 69.35991079832894 289
113 391.6171875 68.69686822985342 275
114 382.1484375 65.29429648758453 284
115 377.4765625 69.48622444904885 271
116 382.3984375 68.92787669048418 265
117 403.0234375 67.0982796961561 278
118 392.70703125 67.1248553112893 261
119 397.25390625 65.25582158218691 282
120 395.12109375 66.63007854980896 261
121 392.71875 70.12900361788623 258
122 380.2890625 68.15855278592038 259
123 409.71875 67.59691393427292 283
124 408.5703125 66.93387403738367 289
125 385.55078125 70.0739938120031 270
126 391.85546875 67.71071428856568 281
127 402.984375 67.56489754938858 303
128 387.72265625 67.81129920001781 286
129 404.49609375 68.13457490321056 290
130 401.6640625 75.01783926837732 223
131 397.12109375 66.81209793183949 270
132 408.953125 63.59550634859648 305
133 392.36328125 67.05516662035372 287
134 397.0078125 67.10346443340197 287
135 405.79296875 67.36340424749571 290
136 388.56640625 68.57070003077088 268
137 397.94140625 66.09593291589476 283
138 396.25390625 66.26876196871503 262
139 399.26953125 70.60277311590015 249
140 400.89453125 69.35428047419116 258
141 387.51953125 67.54141965142777 260
142 407.97265625 66.3346395827198 294
143 394.7578125 71.23829356613508 270
144 403.4921875 59.15861836169641 310
145 387.265625 68.90130899235061 274
146 411.24609375 66.30933357466512 287
147 410.26953125 67.781888319486 283
148 397.32421875 67.49085846210691 280
149 405.71875 66.68617471738426 277
150 402.125 65.08582074845488 264
151 406.7890625 64.52423870431245 286
152 393.2421875 64.36022234435524 283
153 410.61328125 61.64105758225112 275
154 394.328125 70.48421913084073 259
155 399.91796875 65.78953413822006 276
156 410.4921875 72.10111130533872 276
157 399.29296875 68.49609441830624 266
158 401.55078125 64.55845642721523 277
159 398.734375 65.00186107612132 275
160 401.58203125 66.7879186932339 281
161 405.203125 64.23551142268873 284
162 396.0390625 66.04211515480931 277
163 413.6484375 69.35411733493977 294
164 413.40234375 72.88860182330868 269
165 419.2421875 68.30781046274902 286
166 406.71484375 67.54073376239768 285
167 406.91015625 66.9115243015027 285
168 413.28125 67.42889049537668 262
169 396.54296875 70.69674252528559 263
170 396.9765625 66.26939160112754 265
171 413.0625 68.44052504729927 272
172 417.6953125 70.72510147060478 276
173 412.5859375 63.635132707853245 313
174 424.1484375 70.01635856218599 302
175 428.0390625 68.05161211625403 302
176 415.2109375 69.33345065962817 284
177 424.73828125 68.80518991722862 312
178 411.0859375 67.54992683005729 284
179 428.66796875 67.41826003761165 314
180 408.41796875 67.53188706732564 273
181 407.3671875 62.99775183163161 299
182 407.171875 61.733184625324284 298
183 420.8828125 68.50337823487133 286
184 417.5625 68.05532560902196 287
185 418.0390625 65.34624156461558 257
186 411.5703125 68.91539010737982 246
187 413.015625 67.26318230175684 278
188 420.6015625 64.83474664528731 289
189 407.33203125 67.13939491087943 255
190 414.76171875 63.5485316769467 307
191 415.33203125 64.75448757807465 311
192 427.53515625 66.41564444118634 309
193 414.109375 66.65750079405449 305
194 412.40234375 64.4804827118783 301
195 434.98046875 65.29610746269852 334
196 419.47265625 68.64895439713074 276
197 407.19921875 67.40081949160299 297
198 434.98046875 64.3840680004011 311
199 426.77734375 64.80699435974743 334
200 436.89453125 64.143581382651 321
}{\map}

    \addplot[numean]      table[x=xxx,  y expr=\thisrow{y_mean}/2000]   \map; 
        \addlegendentry{average$\pm$ std}
    \addplot[numin]      table[x=xxx,  y expr=\thisrow{y_min}/2000]   \map; 
        \addlegendentry{$\min_\gS$}

    \addplot[name path=upper,draw=none] table[x=xxx,y expr=(\thisrow{y_mean}+\thisrow{y_std})/2000] {\map};
    \addplot[name path=lower,draw=none, forget plot] table[x=xxx,y expr=(\thisrow{y_mean}-\thisrow{y_std})/2000] {\map};
    \addplot[fill=\ourscolor!30, forget plot, opacity=0.5] fill between[of=upper and lower];
\end{axis}
\end{tikzpicture}
}
\end{center}
\vspace{-0.2cm}
 \caption{\textbf{Are all \mname trained?} As a sanity check, this plot shows the average number (in percentage) of times (with standard deviation) that the \sfeatures receive training signal, across training epochs. We also plot the minimum value, which is always significantly positive, showing that each \sfeature ID receives training signal. Thus our loss $\mathcal{L}_{\text{super}}$ proposed in Equation~(\ref{eq:loss_superfeatures}) does not lead to degenerate solutions where some IDs are never selected and never trained.}
\label{fig:are_all_sf_trained}
\end{figure}

The loss in Equation~(\ref{eq:loss_superfeatures}) only operates on a subset of the \sfeature pairs that pass the criteria of Equation~(\ref{eq:pair_matching_criteria}); if a \sfeatureid is not matched, it receives no training signal. To investigate if all of the IDs contribute to the loss, we monitored how many times each ID is matched at each epoch and report the percentage over all 2000 training tuples per epoch. In Figure~\ref{fig:are_all_sf_trained} we report mean, standard deviation and minimum number of matches for the $\nsfeat$ \sfeatureid. We clearly observe that all \mname receive training signals regularly and each \sfeature is matched about one quarter of the time on average. 

\subsection{Using the \sfeature loss with local features} 
\label{app:local}

It is worth noting that the loss in Equation~(\ref{eq:loss_superfeatures}) could also theoretically be used over local feature activations. It could be computed on pairs of local features that pass the conditions described in Equation~(\ref{eq:pair_matching_criteria}),  after removing the constraint on having the same \sfeatureid. 
Unfortunately, we were unable to get any gains over HOW~\citep{how} when appending such a loss side-by-side with the global loss over local features. Empirically, our ablations (see Section~\ref{sub:ablations}) show that adding the constraint on the \sfeatures ID, which is only possible with 
ordered feature sets, is key to the success of our approach, and significantly improves the quality of the matching.

\subsection{Pretraining the backbone together with \lit}
\label{app:pretrain}

Before training on landmark retrieval, we pretrain the backbone network, with the proposed \lit module appended after the last layer, for image classification on ImageNet-1K. Given that we remove the last convolutional block similar to HOW, to avoid overfitting we further append a classification head composed of one fully-connected layer, batch norm, leaky ReLU, dropout and another fully-connected layer at the end of the network and train it for 80 epochs using a standard cross-entropy loss and in addition the decorrelation loss  $\mathcal{L}_{\text{attn}}$ weighted by $0.1$, the same weight as during fine-tuning. We start with a learning rate of $0.01$ and divide it by $10$ every 20 epochs. This new architecture reaches a top-1 (resp. top-5) accuracy of 73.48\% (resp. 91.54\%) on the ImageNet validation set. Note that this performance is a bit lower than a standard ResNet-50: this is because we have removed the last convolutional block which contains most parameters in the ResNet architecture.

\subsection{Computational cost of \sfeatures extraction}
\label{app:speed}
We report the time required for extracting multi-scale features for 5000 images, for HOW and \fire. On our server, it took 157 seconds for HOW and 172 for FIRe, \ie extraction for \sfeatures only requires 10\% more wall-clock time. 

\subsection{Is  Google Landmarks v2 clean an appropriate training dataset for testing on \ROxford and \RParis?}
\label{app:gldv2}

In the comparison to the state of the art (Table~\ref{tab:sota}), all reported methods are trained on the SfM-120k dataset.
Several recent works have also released a model trained on the Google Landmarks v2 dataset~\citep{gldv2} or its clean version~\citep{gldv2_clean} with excellent performance on \RParis.
However, we find out that several of the \emph{query landmarks} from \RParis and \ROxford were present in the training set of the cleaned version of Google Landmarks v2, such as `La Defense', `Eiffel Tower', `Sacr\'e Coeur' or `Hotel les Invalides' for \RParis or such as `Mary Madgalen', `Bodleian Library', `All Souls College' or `Radcliffe' for \ROxford.

\section{Application to visual localization}
\label{app:localize}

In this section, we evaluate \fire for the task of visual localization, where retrieval is used as a first-stage filtering and before more precise, local feature-based geometric matching.
To this end, we follow the pipeline proposed by Kapture\footnote{\url{https://github.com/naver/kapture-localization}}~\citep{kapture} on the Aachen Day-Night v1.1 dataset~\citep{aachen}. In this scenario, a global Structure-from-Motion map is built from the training images using R2D2 local descriptors~\citep{R2D2}. At test time, given a query image to localize, image retrieval is used to retrieve the top-50 nearest images. On these retrieved images, R2D2 local features are extracted and matched with the ones on the query image, and this is used to estimate the position of the camera. The percentage of successfully localized images within three levels of thresholds is then reported on the day or night images, following the visual localization benchmark protocol\footnote{\url{https://www.visuallocalization.net/}}, the latter ones being more challenging as training images are taken during daytime.

We compare our retrieval method to AP-GeM~\citep{apgem} and HOW~\citep{how} and report results in Table~\ref{tab:localization}. AP-GeM is the default method used in Kapture~\citep{kapture}. We observe that using \fire leads to better visual localization, specially in the most challenging scenario of night image localization and the strictest localization threshold: Performance improves by 2\% compared to AP-GeM and by 1.5\% compared to HOW on night images at a threshold of 0.25m and 2\textdegree. Overall, for either day or night images, \fire either improves or performs on par to both methods compared. 

\begin{table}
\centering
 \resizebox{\linewidth}{!}{
 \begin{tabular}{l ccc ccc}
 \toprule
 \multirow{2}{*}{Retrieval method} & \multicolumn{3}{c}{Day images} & \multicolumn{3}{c}{Night images} \\
  & 0.25m, 2\textdegree & 0.5m, 5\textdegree & 5m, 10\textdegree & 0.25m, 2\textdegree & 0.5m, 5\textdegree & 5m, 10\textdegree \\
 \cmidrule(l){1-1} \cmidrule(l){2-4} \cmidrule(l){5-7}
 AP-GeM~\citep{apgem} & 88.8 & \bf{96.6} & \bf{99.6} & 72.3 & 86.9 & 97.9 \\
 HOW~\citep{how}      & \bf{90.8}	& 96.2 & \bf{99.6} & 72.8 & \bf{90.1} & 97.9 \\
\textbf{\fire (ours)}         & \underline{90.7}	& \underline{96.5} & \underline{99.5} &	\bf{74.3} & \bf{90.1} & \bf{98.4} \\
 \bottomrule
 \end{tabular}
 }
 \caption{\textbf{Visual localization results.} Percentage of successfully localized images on the Aachen Day-Night v1.1 dataset when changing the retrieval method in the Kapture pipeline from ~\citet{kapture}.
 In the most challenging scenario, \ie night images at strictest localization threshold, using \fire yields a $2\%$ improvement compared to AP-GeM and a 1.5\% improvement compared to HOW. \textbf{Bold} number denotes the best performance, \underline{underlined} indicates performance within a $0.1$ margin to the best one.}
 \label{tab:localization}
\end{table}

\end{document}